\documentclass{article}

\usepackage{PRIMEarxiv}

\usepackage[utf8]{inputenc} 
\usepackage[T1]{fontenc}    
\usepackage{hyperref}       
\usepackage{url}            
\usepackage{booktabs}       
\usepackage{amsfonts}       
\usepackage{nicefrac}       
\usepackage{microtype}      
\usepackage{lipsum}
\usepackage{fancyhdr}       
\usepackage{graphicx}       
\graphicspath{{media/}}     
\usepackage{comment}
\usepackage{cite}
\usepackage{amsmath,amssymb,amsfonts}
\usepackage{algorithmic}
\usepackage{graphicx}
\usepackage{textcomp}
\usepackage{subcaption}

\usepackage[table,xcdraw]{xcolor} 

\definecolor{Gray_B}{gray}{0.9} 

\usepackage{multirow}
\usepackage{pifont}
\usepackage{siunitx}


\title{Benchmarking 3D Human Pose Estimation Models under Occlusions}

\author{
  Filipa Lino, Carlos Santiago, Manuel Marques \\
  Institute for Systems and Robotics, LARSyS\\
  Instituto Superior Técnico \\
  Portugal\\
  \texttt{\{filipa.lino, carlos.santiago\}@tecnico.ulisboa.pt, manuel@isr.tecnico.ulisboa.pt} \\
}

\begin{document}
\maketitle

\begin{abstract}
Human Pose Estimation (HPE) involves detecting and localizing keypoints on the human body from visual data. In 3D HPE, occlusions, where parts of the body are not visible in the image, pose a significant challenge for accurate pose reconstruction. This paper presents a benchmark on the robustness of 3D HPE models under realistic occlusion conditions, involving combinations of occluded keypoints commonly observed in real-world scenarios. We evaluate nine state-of-the-art 2D-to-3D HPE models, spanning convolutional, transformer-based, graph-based, and diffusion-based architectures, using the BlendMimic3D dataset, a synthetic dataset with ground-truth 2D/3D annotations and occlusion labels. All models were originally trained on Human3.6M and tested here without retraining to assess their generalization. We introduce a protocol that simulates occlusion by adding noise into 2D keypoints based on real detector behavior, and conduct both global and per-joint sensitivity analyses. Our findings reveal that all models exhibit notable performance degradation under occlusion, with diffusion-based models underperforming despite their stochastic nature. Additionally, a per-joint occlusion analysis identifies consistent vulnerability in distal joints (e.g., wrists, feet) across models. Overall, this work highlights critical limitations of current 3D HPE models in handling occlusions, and provides insights for improving real-world robustness.
\end{abstract}


\section{Introduction}\label{sec:introduction}
Human Pose Estimation (HPE) rom RGB images is a longstanding problem in computer vision field and plays a critical role in numerous real-world applications, including sports\cite{wang2019ai,baumgartner2023monocular,yeung2024autosoccerpose}, healthcare\cite{chen2018patient,mroz2021comparing,ngo2024toward,avogaro2023markerless}, and retail\cite{liu2018customer}. Specifically, 3D HPE involves estimating the three-dimensional coordinates of key human body points — such as elbows, knees, and wrists — from single images or videos. This capability allows systems to accurately interpret and analyze human movement, which is essential for tasks like motion capture in animation\cite{kumarapu2020animepose,xia20243d,tous2023pictonaut} and posture assessment in healthcare\cite{mroz2021comparing} or sports \cite{yeung2024autosoccerpose}.

\begin{figure}[!t]
    \centering
    \includegraphics[width=0.5\linewidth]{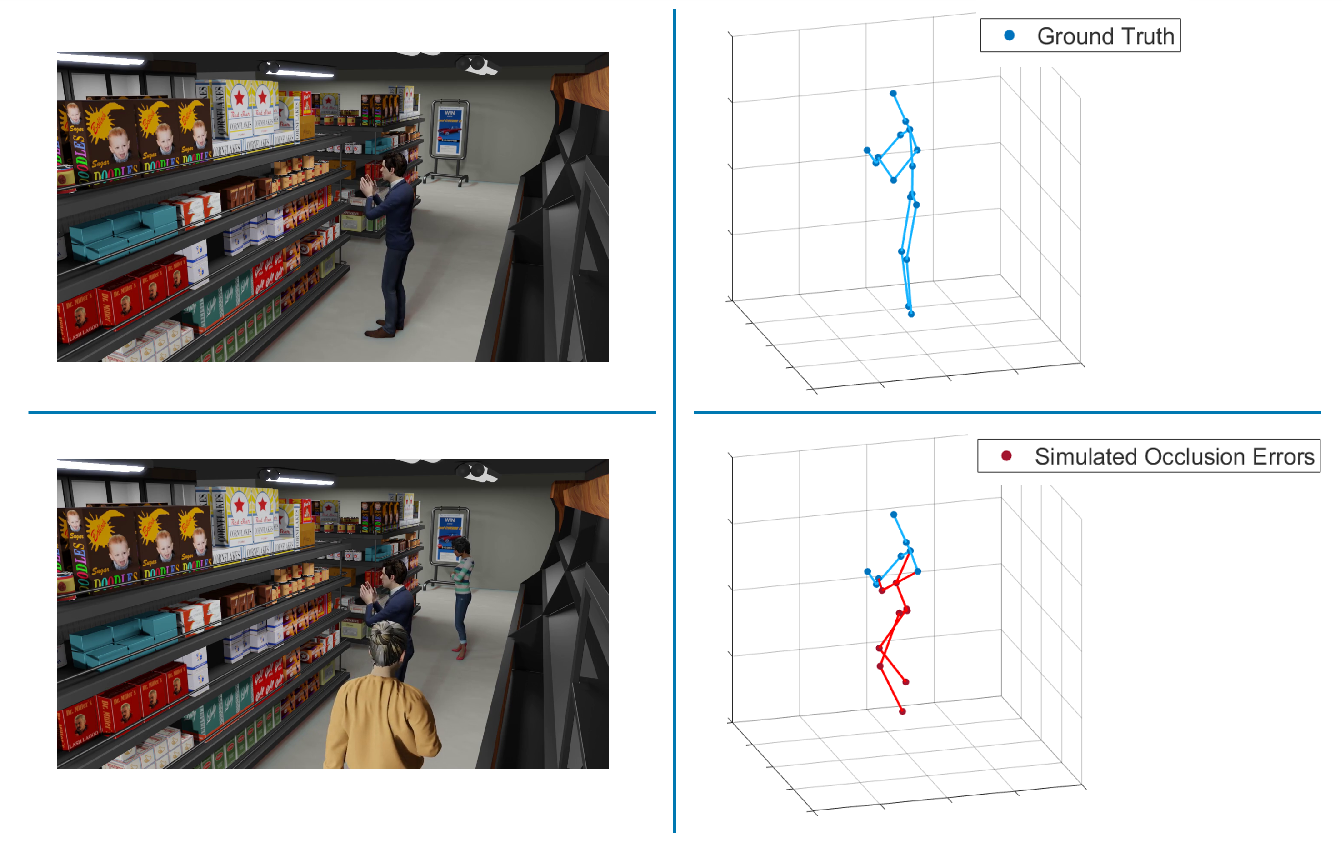}
    \caption{ BlendMimic3D frames show a subject without (top left) and with occlusion (bottom left).  Corresponding 3D ground-truth (top right) and noisy prediction (bottom right) illustrate occlusion impact.   
    }
    \label{fig:simulated_occl_scenario}
\end{figure}

A central challenge in HPE is handling occlusions, where parts of the body are hidden from view—not only due to the subject’s own pose (self-occlusion), but also because of external objects or other people. For instance, in sports such as soccer or basketball, players often block one another, making it difficult for algorithms to reliably estimate keypoint locations\cite{yeung2024autosoccerpose}. In retail settings, limbs may be occluded by shelves or merchandise, preventing full-body pose detection. Similarly, in clinical gait analysis, assistive devices or medical equipment can obstruct key body parts\cite{chen2018patient,ngo2024toward}, requiring pose estimation methods to be robust under partial visibility.

Over the years, various models~\cite{vp3d,pstmo,mixste,pfv2,d3dp,diffupose,finepose,t3d-cnn,dtf,sarandi2023learning,jiang2024back,qiu2020dgcn,hu2021conditional,cai2019exploiting,bm3d} have been developed to address 3D HPE, with a significant focus on the method of lifting from 2D to 3D poses. Despite recent advances, model performance continues to decline in the presence of occlusions, highlighting the critical need for improved robustness in 3D HPE systems.

Given this, our focus is on understanding how well current 3D HPE models handle occlusions, which are inherently present in most real-world settings. To this end, we evaluate several state-of-the-art (SOTA) models for which code and pre-trained weights are publicly available~\cite{vp3d,pstmo,mixste,pfv2,d3dp,diffupose,finepose,t3d-cnn,dtf}.

In 3D HPE, two-stage models first capture a person on camera, detect 2D keypoints, and then lift them to a 3D pose. However, occlusions can lead to inaccuracies in detected 2D poses. As shown in Figure~\ref{fig:simulated_occl_scenario}, these occlusion scenarios can lead to a decline in the performance of 3D reconstruction. This work focuses on 2D-to-3D lifting models and studies their ability to handle such inaccuracies: Can they effectively predict 3D poses from noisy 2D inputs?

Most existing studies use the Human3.6M~\cite{h36m} dataset, which is a controlled environment with indoor settings and minimal obstructions, except for occasional objects in scenarios such as sitting poses. This setup does not fully represent the challenges of real-world occlusions. We opt for the BlendMimic3D~\cite{bm3d} dataset, which includes occlusion labels and synthetic images with occlusions, To the best of our knowledge, BlendMimic3D is the only 3D HPE dataset that provides explicit, frame-level and keypoint-level occlusion labels. These occlusion annotations are central to this work, as they allow us to simulate occlusion noise and analyze how SOTA models respond to it. Thus, we propose a benchmarking protocol for systematic comparison of model performance under controlled occlusion conditions. Additionally, we perform per-keypoint sensitivity analyses to identify vulnerabilities in specific body regions.

To summarize, our work makes the following contributions:
\begin{itemize}
    \item We introduce a comprehensive benchmark of SOTA 3D HPE models under realistic occlusion conditions, moving beyond clean-input and randomly masked evaluation settings.

    \item Through a controlled occlusion simulation based on 2D detector behavior, we reveal differences in robustness across convolutional, transformer-based, graph-based, and diffusion-based lifting architectures.
    
    \item We propose two evaluation protocols, supporting both global and per-keypoint performance analyses, to assess model robustness to occluded 2D keypoints.

    \item Our results show consistent vulnerability patterns across models, with distal keypoints (e.g., wrists and feet) among the most error-prone. 

    \item Finally, we show that recent diffusion-based and occlusion-aware approaches, while effective in controlled settings, do not inherently confer robustness to realistic occlusion patterns, highlighting open challenges for deploying 3D HPE systems in unconstrained environments.
    
    
\end{itemize}

This paper raises critical questions about the applicability of 3D HPE advances in real-world settings and the adequacy of current benchmarks. By analyzing model behavior under realistic occlusion and distribution shifts, this study aims to guide future research toward more robust pose estimation systems.

This paper is structured as follows. Section~\ref{sec:related_work} reviews related work on 3D HPE, with a focus on occlusion handling. Section~\ref{sec:experimental_setup} describes the BlendMimic3D dataset, the selected models, the occlusion simulation protocols, and the evaluation metrics. Section~\ref{sec:results} presents benchmarking results under varying occlusion conditions, including a per-keypoint occlusion analysis. Section~\ref{sec:beyond_occl} extends the analysis to the effects of dataset distribution shifts and camera placement. Finally, Section~\ref{sec:conclusion} concludes the paper and outlines directions for future research.

\section{Related Work}
\label{sec:related_work}
Deep Learning has significantly advanced 2D Human Pose Estimation (HPE)~\cite{zheng2023deep,yan2024efficient}. Although 3D HPE has also benefited from these advantages, it remains more challenging, particularly in monocular settings, due depth ambiguities, limited training data, and occlusions~\cite{bm3d,yan2024efficient,udayan2025deep}. 

Recent surveys on 3D HPE have reviewed deep learning architectural trends~\cite{zheng2023deep}, benchmark performances~\cite{wang2021deep}, and progress in monocular and multi-view setups~\cite{zheng2023deep,sarafianos20163d,wang2021deep}. However, these works typically evaluate performance using global accuracy metrics on controlled benchmarks like Human3.6M~\cite{h36m} or MPI-INF-3DHP~\cite{mehta2017monocular}, which do not reflect the complexities of real-world. Udayan et. al~\cite{udayan2025deep} recently conducted a systematic review of monocular 3D HPE, highlighting the gap between academic performance and practical applicability, and emphasizing the need to tackle challenges such as occlusions and depth ambiguity. However, their work focuses on synthesizing existing literature and does not include new experimental validation of the highlighted challenges.

Sarafinos et al. ~\cite{sarafianos20163d} tackled the benchmark limitations by introducing a synthetic dataset that allows full control over scene variables and human pose configurations. Their goal was to test model robustness to anthropometric variability, viewpoint changes and action complexity. While this work offered valuable insights into classical methods, it primarily covered pre-deep learning techniques from 2008 to 2015 and included only a limited discussion of modern deep learning-based approaches. Building on this line of research, our work extends the evaluation of occlusion robustness to current state-of-the-art (SOTA) deep learning models, leveraging a dataset specifically designed for fine-grained occlusion analysis.

3D HPE models are typically divided into direct estimation and 2D-to-3D lifting. Some works have explored direct 3D HPE, which estimates 3D poses directly from images without relying on intermediate 2D detections~\cite{li20143d,pavlakos2018ordinal,sarandi2023learning,wang2024text}. However, direct methods face challenges in resolving depth ambiguities and occlusions due to the absence of intermediate spatial cues and limited contextual reasoning.

In this work, we focus on lifting approaches, which remain widely used due to their ability to leverage existing 2D pose detectors as input. In the following sections, we review the main approaches to monocular 2D-to-3D HPE models.

\subsection{2D-to-3D HPE}
Many studies have adopted 2D-to-3D lifting approaches, benefiting from the advances in 2D HPE. This two-stage approach has demonstrated superior performance on several benchmark datasets and is highly efficient, especially in settings where temporal context is available. Temporal information was incorporated by Pavllo et al.~\cite{vp3d}, they proposed the VideoPose3D model to address video sequences with a temporal dilated convolutional network. VideoPose3D is noted for its simplicity and efficiency. However, it assumes independence between consecutive errors, making it less robust to continuous occlusions or persistent noise.

Given that a human pose can be represented as a graph, where keypoints are nodes and bones are edges, graph-based approaches~\cite{qiu2020dgcn,hu2021conditional,cai2019exploiting,bm3d} have been applied to model relationships between keypoints over time. Qiu et al.~\cite{qiu2020dgcn} introduced a Dynamic Graph Convolutional Network to capture temporal dependencies between 2D keypoints. Hu et al.~\cite{hu2021conditional} represented the 3D human skeleton as a directed graph to encode anatomical hierarchies, modeling bones as directed edges from parent to child keypoints, following the natural articulation from the hip (root) to the extremities. Cai et al.~\cite{cai2019exploiting} proposed leveraging spatial-temporal relationships within a graph based framework. Building on these ideas, our previous work~\cite{bm3d} introduced a 3D pose refinement block designed to correct occluded poses using graph-based reasoning.

While graph-based models have effectively exploited the structured nature of the human skeleton, capturing local and hierarchical dependencies among keypoints, transformer-based models have recently gained prominence due to their ability to capture long-range dependencies of keypoints. Transformer architectures \cite{zheng20213d,pstmo,zhang2022mixste,zhao2023poseformerv2} process spatial-temporal features with global attention mechanisms, enabling keypoint reasoning across multiple pose sequences and leading to more robust 3D HPE. Zheng et al.~\cite{zheng20213d} presented the first purely transformer-based 3D HPE approach, named PoseFormer, without convolutional architectures involved. This model processes both spatial and temporal aspects of human poses. Inspired  by the progress of Masked Image Modeling (MIM)~\cite{xie2022simmim} in image classification, Shan et al.~\cite{pstmo} proposed P-STMO, that applies masked keypoint modeling to 3D HPE with self-supervised learning. P-STMO improves accuracy in 3D pose estimation while reducing the need for large labeled datasets by randomly masking keypoints and frames using learnable embeddings, enabling the encoder to infer missing information. Zhang et al.~\cite{mixste} pointed out that PoseFormer overlooks motion differences between body keypoints and proposed MixSTE, a model that captures inter-frame correspondences using a mixed spatio-temporal encoder to extract fine-grained, keypoint-specific features. Building on PoseFormer, Zhao et al.~\cite{zhao2023poseformerv2} proposed PoseFormerV2, which leverages a compact frequency-domain representation of long skeleton sequences to efficiently scale up the receptive field and enhance robustness to noisy 2D keypoint detections, making it well-suited for complex and occluded scenarios. 

Recently, diffusion-based models have emerged as a promising direction for 3D HPE~\cite{d3dp,mixste, diffupose,finepose,lee2024hdpose}, generating and refining multiple 3D hypotheses from 2D observations through a gradual denoising process. These models show potential in addressing challenges such as occlusions and depth ambiguity~\cite{d3dp,diffupose,finepose}. Shan et al.~\cite{d3dp} introduced D3DP, which generates multiple 3D pose candidates from a single 2D input, using a denoiser conditioned on 2D keypoints and adapted from MixSTE~\cite{mixste}. Choi et al.~\cite{diffupose} proposed DiffuPose, which integrates a graph convolutional architecture to capture keypoint topology and guide the denoising process across the graph structure. Xu et al.~\cite{finepose} presented FinePose, a prompt-driven diffusion model that incorporates semantic information, such as body parts, action classes, and motion dynamics, to enable controllable pose generation, which shows promising results in handling complex poses.

Despite improved accuracy, diffusion models face challenges in computational efficiency due to their iterative nature, requiring multiple forward passes for each prediction. Their performance is also highly dependent on the quality of 2D inputs, making them vulnerable to occlusions. Furthermore, even in the absence of occlusions, not all keypoints are equally difficult to estimate. In 2D detection, Yu et al.~\cite{yu2022scale} showed that some keypoints are more easily detectable than others due to their scale. However, 3D HPE models typically allocate equal attention to all keypoints. 

\subsection{Occlusion Handling in 3D HPE}

Our previous work \cite{bm3d} employed graph-based 3D pose refinement block trained on occluded scenarios. While effective in these conditions, this approach does not consistently perform well in non-occluded settings, highlighting the need for occlusion-aware networks. Addressing this, Cheng et al.~\cite{cheng2019occlusion} included a cylinder man model to generate self-occlusion labels. Ghafoor and Mahmood~\cite{t3d-cnn} proposed T3D-CNN, an occlusion-guided framework based on temporal dilated convolutions, explicitly modeling missing keypoints via an occlusion guidance matrix. Later, Ghafoor et al.~\cite{dtf} extended this idea with a temporal interpolation-based occlusion guidance mechanism and a transformer-based approach that generates intermediate views, subsequently fused for the final 3D pose estimation, a model named DTF. However, both T3D-CNN and DTF~\cite{t3d-cnn,dtf} rely on the assumption that occlusion labels are known in advance. Their occlusion-guided frameworks depend on an external model to indicate which keypoints are occluded, treating these as missing inputs during prediction. Wang et al.~\cite{wang2022ocr} separate features into occlusion-aware and view-dependent components, completing occluded poses in 2D space, potentially leading to inaccurate 3D coordinates. Hardy and Kim~\cite{hardy2024links} reconstruct occluded parts in 3D but do not address all occlusion types. Thus, there is still room for improvement in occlusion-aware models.

Despite recent advancements in the field~\cite{yan2024efficient}, robust handling of occlusions under complex, real-world conditions remains a key open problem. To address this, our study evaluates the performance of several SOTA models under controlled occlusion scenarios using the BlendMimic3D dataset~\cite{bm3d}. In this work, we leverage its existing occlusion labels to simulate 2D occlusion noise for comprehensive benchmarking.

\subsection{3D HPE Generalization}
Learning-based methods have achieved significant progress in 3D HPE, but their performance is highly dependent on the training datasets~\cite{sarandi2023learning,jiang2024back}. In practice, models trained on one dataset often struggle when evaluated on others due to domain shifts and discrepancies in skeleton annotations. Jiang et al.~\cite{jiang2024back} approached this challenge by integrating optimization strategies into a diffusion-based model that iteratively refines 3D pose predictions. However, because skeleton annotations differ across datasets, their diffusion model must be retrained for each dataset, limiting its generalization.

A common strategy to improve generalization is to train on multiple datasets. However, this introduces challenges related to differences in skeleton definitions~\cite{rapczynski2021baseline,sarandi2023learning}. While many datasets use similarly named keypoints, their spatial definitions often differ—for instance, hip keypoints may be located at different heights, or some datasets label surface markers while others use anatomical interior points. Furthermore, certain keypoints may be missing entirely in some datasets. Mahmood et al.~\cite{mahmood2019amass} introduced the AMASS dataset and addressed the problem of discrepancy in MoCap data representations by mapping them to the SMPL~\cite{loper2023smpl} representation, a parametric body mesh widely used in graphics and vision. However, since our focus lies in 3D skeleton-based pose prediction this solution is not directly applicable.

In 2D-to-3D pose lifting, Rapczyński et al.~\cite{rapczynski2021baseline} proposed combining multiple datasets and manually aligning skeletons through hand-crafted rules (e.g., scaling the hip-to-pelvis distance). However, this approach scales poorly with large keypoint sets or many datasets. As an alternative, Sárándi et al.~\cite{sarandi2023learning} proposed projecting poses from different datasets into a shared latent space via autoencoders. While effective as an intermediate format, this approach does not generate a definitive or universal keypoint schema. 

\subsection{2D and 3D HPE Datasets}

Popular 2D HPE datasets include MPII~\cite{andriluka20142d}, LSP~\cite{johnson2010clustered}, LSP-extended~\cite{johnson2011learning} and COCO ~\cite{coco}. Among these, COCO has become the most widely used benchmark for image-based pose estimation for its scale and diversity. However, being human-annotated, COCO is also prone to label noise and inconsistencies.

On the other hand, 3D HPE datasets typically require sophisticated equipment like motion capture (MoCap) systems for accurate pose recording~\cite{h36m,mehta2017monocular,mahmood2019amass,varol17_surreal,sigal2010humaneva,joo2015panoptic}. The most commonly used 3D HPE datasets include Human3.6M~\cite{h36m}, one of the largest MoCap dataset with RGB video, 3D keypoint locations, and camera parameters. It is standard for benchmarking 3D pose estimation under controlled conditions. MPI-INF-3DHP~\cite{mehta2017monocular} was captured with a commercial markerless MoCap system in both indoor and outdoor scenes, with multi-view RGB, green screen and studio backgrounds. HumanEva-I~\cite{sigal2010humaneva} is an early MoCap indoor dataset for monocular/multi-view 3D HPE. CMU Panoptic~\cite{joo2015panoptic} is a multi-camera indoor dataset capturing social interactions between multiple people. While these datasets provide accurate 3D ground-truth, they often lack diversity, realism, and occlusion scenarios. To address these issues, both synthetic and in-the-wild datasets have emerged. 

3DPW~\cite{von2018recovering} is the first dataset with 3D ground-truth obtained via wearable IMUs and cameras in real-world environments, including multi-person settings. SURREAL~\cite{varol17_surreal} is a synthetic dataset with random combinations of 3D poses, body shapes, textures, and backgrounds. BEDLAM~\cite{black2023bedlam}, a recent synthetic dataset designed for 3D human pose and shape (HPS) estimation, showed that models trained purely on synthetic data can generalize to real images. Other than these, recent efforts are moving toward more diverse (e.g., occlusion-heavy or multi-person\cite{fan2023human}) and sensor-fused datasets.

Despite recent advances, Human3.6M~\cite{h36m} remains the main benchmark for single-person 3D pose, while COCO~\cite{coco} dominates in 2D HPE. However, each has limitations, COCO’s human curated nature is prone to errors, and Human3.6M lacks occluded and real-world conditions. 

Sárándi et al.\cite{sarandi2018robust} showed that SOTA 3D HPE models suffer significant performance drops when evaluated under synthetic occlusions. To address this, in our previous work we introduced BlendMimic3D~\cite{bm3d}, a synthetic dataset designed to include realistic occlusion scenarios. It includes action-specific sequences with keypoint visibility labels per frame, allowing precise evaluation under occlusion. Building on this resource, the present study leverages BlendMimic3D to analyze the behavior of SOTA 3D HPE models under different types and severities of occlusion. To the best of our knowledge, this is the first benchmark to perform a structured, occlusion-aware evaluation across a diverse set of models.

\section{Experimental Setup}
\label{sec:experimental_setup}

\subsection{Selected Models}
For this benchmark, we selected a diverse set of SOTA 2D-to-3D human pose estimation models covering different architectural types: convolutional, graph-based, transformer-based, and diffusion-based. Selection was based on the availability of open-source implementations, as well as their relevance and performance in the literature. All models were pretrained using an off-the-shelf 2D keypoint detector, CPN~\cite{cpn}, on the Human3.6M~\cite{h36m} dataset, with a receptive field of 243 frames, except for DTF~\cite{dtf} and DiffuPose~\cite{diffupose}, which used 351 and 384 frames, respectively, as in their original proposals. Where pretrained weights were not available, we trained the models from scratch using authors' official configurations, specifically for DiffuPose~\cite{diffupose}, DTF~\cite{dtf} and T3D-CNN~\cite{t3d-cnn}. All the selected models are summarized in Table~\ref{tab:selected_models}. This selection enables a comprehensive comparison across methodologies, particularly under occlusion conditions explored in this study.

\begin{table}[t]
    \centering
    \caption{Overview of the selected 2D-to-3D human pose estimation models benchmarked in this study. For each model, we summarize its main architectural focus, if it involves an occlusion aware training (OAT), and the availability of pretrained weights.
    }
    \vspace{0.2cm}
    \resizebox{\textwidth}{!}{
    \begin{tabular}{l|l|c|c}
    \textbf{Model} & \multicolumn{1}{c|}{\textbf{Highlight}} & \textbf{OAT} &\textbf{Pretrained Model} \\
    \hline
    \rowcolor{Gray_B} VideoPose3D~\cite{vp3d}& Temporal convolutional network known for its simplicity and competitive performance. & \ding{53} & \checkmark \\
     P-STMO~\cite{pstmo} & Transformer-based model trained with masked keypoint modeling for self-supervised learning. & \ding{53}  & \checkmark  \\
     \rowcolor{Gray_B} MixSTE~\cite{mixste} & Mixed spatio-temporal transformer that captures fine-grained inter-keypoint motion dynamics.& \ding{53}  & \checkmark \\
     PoseFormerV2~\cite{pfv2} & Frequency-domain transformer designed for long-sequence efficiency and robustness.& \ding{53}  & \checkmark \\
     \rowcolor{Gray_B} D3DP~\cite{d3dp} & Diffusion-based model using a denoiser adapted from MixSTE.& \ding{53}  & \checkmark \\
     DiffuPose~\cite{diffupose} & Diffusion model with a graph convolutional backbone for topology-aware denoising.& \ding{53}  & \ding{53} \\
     \rowcolor{Gray_B} FinePose~\cite{finepose} & Prompt-driven diffusion model incorporating action class, body part, and kinematic context. & \ding{53} & \checkmark \\
     T3D-CNN~\cite{t3d-cnn}& Occlusion-aware model using temporal convolutions guided by a keypoint-level occlusion mask.& \checkmark & \ding{53} \\
     \rowcolor{Gray_B} DTF~\cite{dtf}& Occlusion-aware transformer-based model with view-fusion and occlusion interpolation. & \checkmark & \ding{53} \\
     \hline
    \end{tabular}
    }
    \label{tab:selected_models}
\end{table}

Before analyzing model robustness under occlusions, we first evaluate the baseline performance of all selected models on the Human3.6M test set using 2D keypoints predicted by CPN~\cite{cpn}. This evaluation establishes an upper-bound reference for each model’s accuracy in an in-distribution setting. Table~\ref{tab:h36m_results} summarizes the results.

\begin{table}[t]
    \centering
    \caption{Baseline MPJPE of each model on Human3.6M using CPN detections. OAT denotes Occlusion-Aware Training. The best results are highlighted in \textbf{bold} and the second-best results are \underline{underlined}. }
    \vspace{0.2cm}
    \resizebox{0.5\textwidth}{!}{
    \begin{tabular}{l c | c| c |}
    \cline{3-4}
        \textbf{Models} & \textbf{Year} & \textbf{OAT} & \textbf{MPJPE - Avg} [mm]\\ 
        \hline
        \rowcolor{Gray_B} VideoPose3D~\cite{vp3d} & 2019 & \ding{53} & 46.8\\
        P-STMO~\cite{pstmo} & 2022 & \ding{53} & 42.8 \\
        \rowcolor{Gray_B} MixSTE~\cite{mixste} & 2022 & \ding{53} & 40.9 \\
        PoseFormerV2~\cite{pfv2} & 2023 & \ding{53} & 45.2 \\
        \rowcolor{Gray_B} D3DP~\cite{d3dp} (P-Agg) & 2023 & \ding{53} & \textbf{39.9}\\
        DiffuPose~\cite{diffupose} (P-Agg)& 2023 & \ding{53} & 53.8\(^\dagger\) \\
        \rowcolor{Gray_B} FinePose~\cite{finepose} (P-Agg)& 2024 & \ding{53} & \underline{40.6}\\
        T3D-CNN~\cite{t3d-cnn} (None) & 2022 & \checkmark &  56.6\(^\ddagger\) \\ 
         \rowcolor{Gray_B} DTF-GT~\cite{dtf} (Rand 16) & 2024 & \checkmark & 53.5 \\ 
        \hline
        \multicolumn{4}{l}{\small  \(^\dagger\) This value is higher than in their paper because they use HR-Net~\cite{sun2019deep} as}\\
        \multicolumn{4}{l}{\small 2D pose detector. We retrained it with CPN detections.}\\
        \multicolumn{4}{l}{\small  \(^\ddagger\) This value is higher than in their paper because they use 2D ground-truth }\\
        \multicolumn{4}{l}{\small input. We retrained it with CPN detections.}
    \end{tabular}
    }
    \label{tab:h36m_results}
\end{table}

For diffusion-based models, we report results using the Pose-level Aggregation (P-Agg) strategy, where multiple 3D pose hypotheses are averaged to produce a final output. We chose this approach because alternative aggregation strategies, such as Joint-wise reProjection-based Multi-hypothesis Aggregation (JPMA)~\cite{d3dp}, which relies on reprojection accuracy, can become unreliable in the presence of noisy or occluded 2D inputs. Additionally, methods like P-Best and J-Best, which require access to ground-truth 3D poses, are not feasible in real-world applications, where ground-truth 3D poses are unavailable at inference time. 


For occlusion-aware models, namely T3D-CNN~\cite{t3d-cnn} and DTF \cite{dtf}, we replicate the training protocols proposed by the original authors. For DTF, we report results using the \textit{Rand 16} setting, where 16 keypoints are randomly masked as occluded. In contrast, for T3D-CNN we select the \textit{None} configuration, in which the occlusion-guidance channels are removed, as it yields the most stable performance among the evaluated T3D-CNN variants. A more detailed analysis of T3D-CNN under different masking strategies is provided later in the paper.

As shown in Table~\ref{tab:h36m_results}, D3DP achieves the best overall performance, followed closely by FinePose and MixSTE. These results highlight the effectiveness of diffusion-based models in occlusion-free settings. Among transformer-based models, MixSTE, P-STMO and PoseFormerV2 all perform competitively, while VideoPose3D, despite its simplicity and earlier design, still offers a solid baseline. However, these numbers represent upper-bound performance, and do not reflect robustness to occlusion, addressed in the following sections.

On the other hand, occlusion-aware models T3D-CNN and DTF show higher MPJPE. This can be attributed to their training protocol, which simulates occlusion by randomly masking 16 keypoints in each frame (Rand 16), allowing the model to observe only one keypoint at a time. In this work, we evaluate how well these models generalize to real occlusion scenarios and whether this training scheme translates into improved robustness in practice.

\subsection{Benchmark Dataset}

To assess generalization to in-the-wild scenarios, we benchmark models on the BlendMimic3D~\cite{bm3d} dataset. Since all selected models were trained on Human3.6M~\cite{h36m} dataset, evaluating them on that same dataset would not assess their robustness to distribution shifts or unseen occlusions. BlendMimic3D provides a more realistic test set, with occlusions caused by object interactions and different viewpoints.

BlendMimic3D offers synchronized RGB videos from four camera views, along with ground-truth 2D and 3D pose annotations and occlusion labels, per keypoint and per frame. These occlusion annotations are crucial for this benchmark, as they allow us to simulate and manipulate different levels and types of realistic occlusion noise in a controlled manner.

\begin{figure}[t]
    \centering
    \includegraphics[width=0.6\linewidth, trim={80pt 50pt 40pt 0pt}, clip]{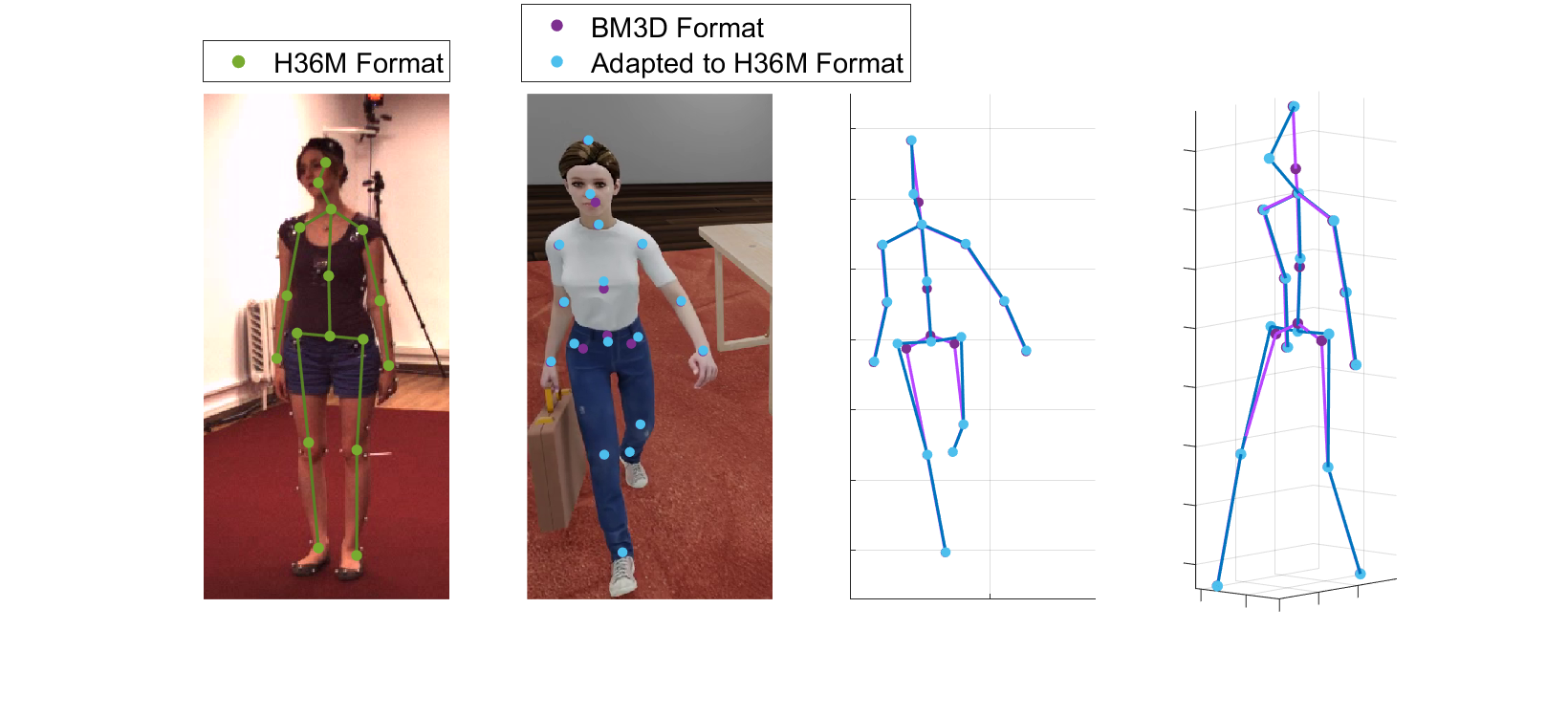}
    \caption{ Comparison of 2D human pose representations across different formats. (Left) Human3.6M (H36M) format over the original image. (Middle-Left) BlendMimic3D (BM3D) format and its version adapted to H36M. (Middle-Right) The corresponding 2D poses. (Right) The corresponding 3D poses.}
    \label{fig:formats}
\end{figure}

We use subject S2 from BlendMimic3D, who performs 14 actions with various objects placed in the scene. To ensure compatibility across models and avoid retraining, we manually adapted the BlendMimic3D skeleton format in Blender to closely match the keypoint structure and conventions of Human3.6M, as can be seen in Figure~\ref{fig:formats}. The converted 2D and 3D pose annotations, along with occlusion labels, will be made publicly available.

In the Human3.6M dataset, keypoints such as \textit{Nose} and \textit{Head} are captured based on the orientation of the head using surrounding markers. Consequently, the position of the \textit{Nose} or \textit{Head} is not anatomically grounded and may vary across subjects and actions. In contrast, our Blender environment enforces anatomical consistency by fixing all keypoints, including \textit{Nose} and \textit{Head}, to precise, predefined locations.

\begin{figure}[t]
    \centering
    \begin{minipage}{0.49\linewidth}
        \centering
        \includegraphics[width=\linewidth]{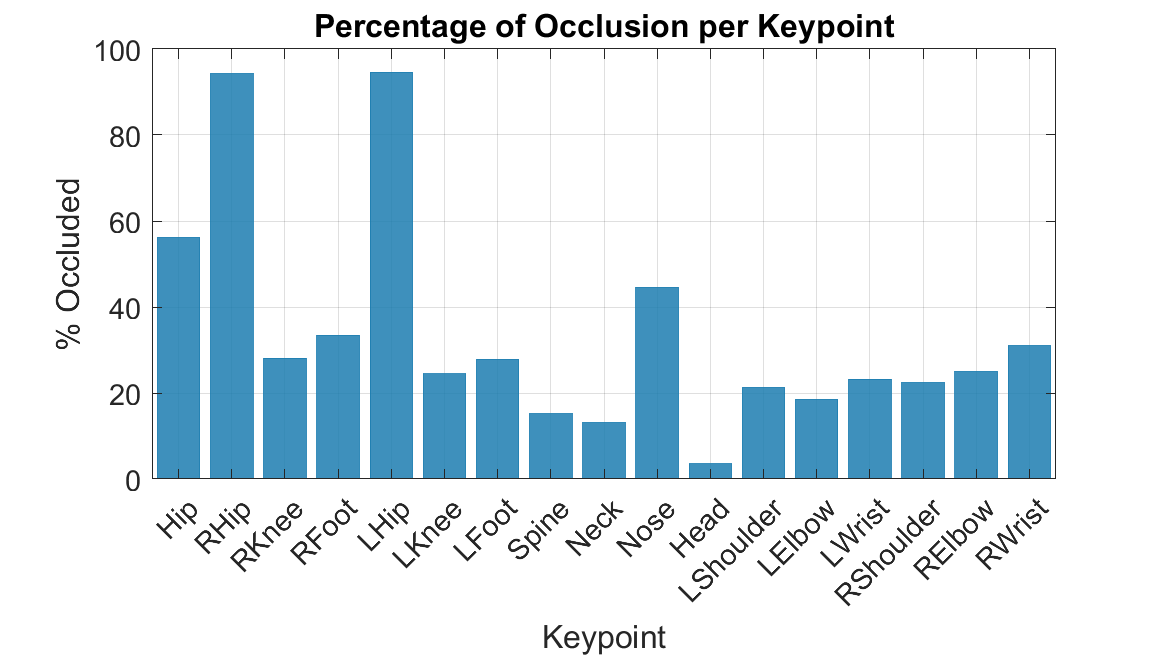}
        \subcaption{}
        \label{fig:bm3d_stats_a}
    \end{minipage}
    \hfill
    \begin{minipage}{0.49\linewidth}
        \centering
        \includegraphics[width=\linewidth]{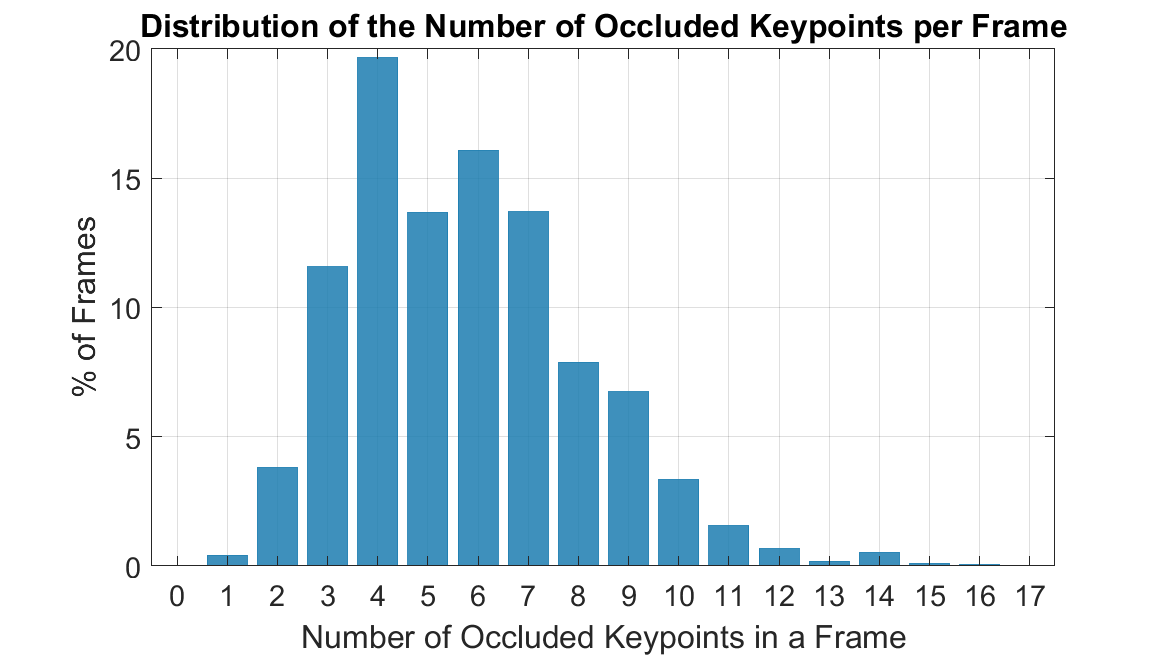}
        \subcaption{}
        \label{fig:bm3d_stats_b}
    \end{minipage}
    \caption{Occlusion statistics of subject S2 in the BlendMimic3D dataset. (a) Percentage of occlusion per keypoint across all actions and frames.  (b) Distribution of the number of occluded keypoints per frame.}
    \label{fig:bm3d_stats}
\end{figure}

Figure~\ref{fig:bm3d_stats_a} shows the percentage of occlusion per keypoint across all frames, with an average of $33.93\%$ of keypoints occluded per frame. Notably, \textit{RHip} and \textit{LHip} are the most frequently occluded keypoints. In the Human3.6M format, \textit{Hip}, \textit{RHip}, and \textit{LHip} are collinear (as illustrated in Figure~\ref{fig:formats}), which makes it difficult in Blender to accurately segment the body volume of each keypoint individually. As a result, these keypoints are often labeled as occluded. However, our results show that, despite frequent occlusion, hip keypoints are still well predicted by 3D HPE models. Figure~\ref{fig:bm3d_stats_b} presents the distribution of the number of occluded keypoints per frame. Notably, there are no frames where all keypoints are visible or completely occluded, the majority of frames have 3 to 7 occluded keypoints.

\subsection{Occlusion Simulation Protocol}
In 2D-to-3D human pose estimation, the accuracy of the predicted 3D pose depends heavily on the quality of the 2D keypoints used as input. When body parts are occluded in the image, 2D keypoint detectors such as CPN~\cite{cpn} and Detectron2~\cite{detectron2}, often produce noisy or inaccurate predictions. This makes occlusion in the 2D input space one of the primary sources of error for lifting-based models, and a critical challenge in real-world predictions. 

To evaluate the robustness of 3D HPE models under occlusions, we designed two noise injection experiments using the BlendMimic3D dataset. In both cases, we simulate the effect of occlusion by corrupting 2D input poses with Gaussian noise, following the behavior of real 2D detectors.

\subsubsection{Protocol 1: Adding Noise to Occluded Keypoints} \label{sec:protocol1}
In order to understand how well each model can infer 3D poses from corrupted 2D input data, and thus assess the models' practical usability in different occlusion levels, we tracked their performances across different noise levels. We applied zero-mean Gaussian noise with varying levels of standard deviation to occluded 2D keypoints, using the per-keypoint occlusion annotations available in BlendMimic3D. This approach allows us to systematically simulate errors in 2D detections without introducing bias.

Let \( \mathbf{x}_{ij} \in \mathbb{R}^2 \) denote the 2D position of keypoint \( j \) in frame \( i \), and let \( \mathcal{O}_{ij} \in \{0, 1\} \) be the corresponding occlusion indicator, where \( \mathcal{O}_{ij} = 1 \) denotes that the keypoint is occluded. We define the noise level parameter \( \sigma \in \{0.001, 0.005, 0.01, 0.03, 0.05\} \), which is expressed as a fraction of the average image resolution. The corresponding standard deviation in pixel units is given by \[ \sigma_{\text{px}} = \sigma \cdot \frac{w + h}{2}, \] where \( w \) and \( h \) are the image width and height, respectively.

We add noise into each occluded keypoint by sampling a 2D perturbation vector from an isotropic Gaussian distribution:
\[
\tilde{\mathbf{x}}_{ij} =
\begin{cases}
\mathbf{x}_{ij} + \boldsymbol{\varepsilon}_{ij}, & \text{if } \mathcal{O}_{ij} = 1 \\
\mathbf{x}_{ij}, & \text{otherwise}
\end{cases}
\quad \text{with} \quad \boldsymbol{\varepsilon}_{ij} \sim \mathcal{N}(\mathbf{0}, \sigma_{\text{px}}^2 \mathbf{I}_2)
\]

\begin{figure}[t]
    \centering 
    \includegraphics[width=\linewidth, trim={120pt 150pt 90pt 130pt}, clip]{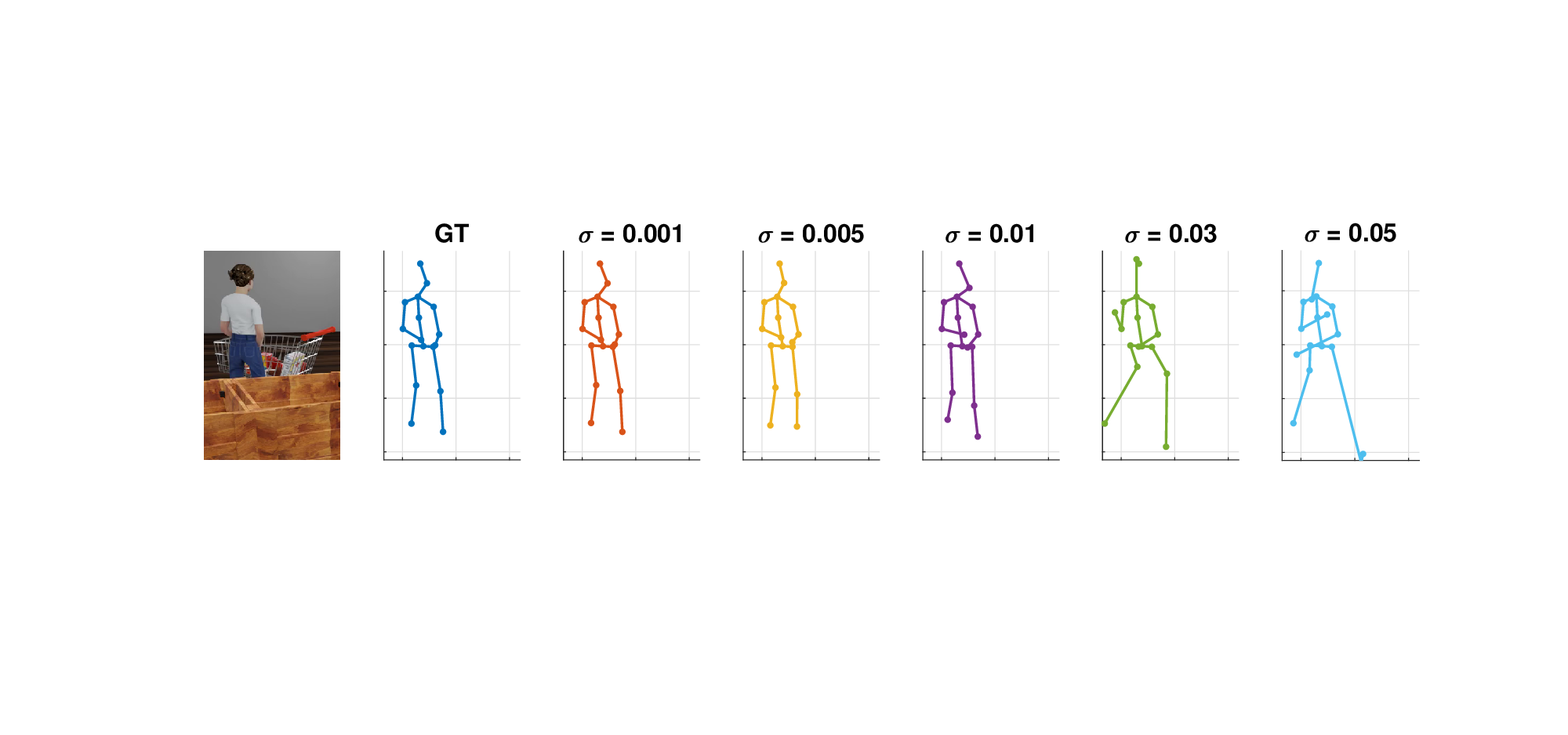}
    \caption{2D human poses derived from the ground-truth skeleton, with varying levels of Gaussian noise applied to the occluded keypoints: the nose, hands, knees, and feet. }
    \label{fig:2d_noises}
\end{figure}

This process perturbs only the occluded keypoints in the 2D pose sequence, injecting Gaussian noise independently to the \(x\) and \(y\) coordinates of each occluded keypoint. We tested five noise levels, \(\sigma=\{0.001, 0.005, 0.01, 0.03, 0.05\}\), applied over 10  independent runs per level for statistical robustness. For comparison, we also included the case of 2D ground-truth pose input. Figure~\ref{fig:2d_noises} illustrates qualitative examples across noise levels. 

\subsubsection{Protocol 2: Per-Keypoint Occlusion Analysis} \label{sec:protocol2}
 
In the second evaluation, we analyze the sensitivity of each model to the occlusion of individual keypoints. The goal is to quantify the relative importance of each keypoint in accurate 3D pose estimation. For this, we selected the first 243 frames of each sequence and, over a 100-frame window (frame 50--149), for each of the 17 keypoints, injected zero-mean Gaussian noise into that keypoint alone, keeping all the others unchanged. This was repeated over five independent runs per keypoint to account for variability. By isolating occlusion to a single keypoint at a time, this test allows us to assess which keypoints impact 3D pose accuracy the most. The results help identify which body parts are easier or more difficult for each model to predict. 

The standard deviation used for the Gaussian noise was chosen based on findings from the first evaluation and informed by real detector statistics, which are discussed later in the results section. This ensures that the simulated noise realistically reflects the typical errors caused by occlusions in real-world scenarios. 

\subsection{Evaluation Metrics}
Our evaluation is based on the Mean Per-keypoint Positional Error (MPJPE), which measures the average Euclidean distance (in millimeters) between the predicted and ground-truth 3D keypoint positions. MPJPE is represented by the equation
\begin{equation}
    \text{MPJPE} = \frac{1}{N}\sum^N_{i=1}\lVert J_i-J_i^*\rVert_2, 
\end{equation}
where \(N\) is the number of keypoints, and \(J_i\), \(J^*_i\), are the ground-truth and predicted positions of the \(i_{th}\) keypoint, respectively. 

To analyze model performance under occlusions, we compute and report the following variants:

\begin{itemize}
    \item Overall MPJPE: The average error over all keypoints in all frames, regardless of visibility.
    
    \item Visible MPJPE: The average error computed over all keypoints labeled as \textit{visible} across all frames. This isolates the model’s performance on visible keypoints.
    
    \item Occluded MPJPE: The average error computed over all keypoints labeled as \textit{occluded} across all frames. This directly reflects the model's ability to estimate occluded keypoints. 
   
\end{itemize}

All errors are reported in millimeters and averaged over all actions and test runs for statistical robustness. When applicable, we report standard deviation as a confidence measure across runs.

\noindent\textbf{Hip keypoint evaluation:} We observed discrepancies in how different models handle the hip keypoint. Specifically, P-STMO, D3DP, FinePose and DTF, explicitly reset the predicted hip coordinates to zero before evaluation, regardless of their actual prediction. This post-processing step artificially reduces the MPJPE for the keypoint to zero, introducing an evaluation bias. To ensure consistency and fairness across models, we excluded the hip keypoint from all poses during evaluation.

\section{Results and Discussion}
\label{sec:results}

\subsection{2D Detector Error Analysis} \label{sec:gaussian}

To better understand the behavior of 2D pose detectors under occlusion, we conducted an analysis focusing on the types and magnitudes of detection errors in realistic scenarios. To this end, we analyzed two commonly used 2D detectors, CPN~\cite{cpn} and Detectron2~\cite{detectron2}, on 1000x1002 images from subject S2 in BlendMimic3D. The detectors output 2D poses in COCO format~\cite{coco}, which we aligned semantically with the Human3.6M-converted ground-truth, excluding non-aligned keypoints such as head, neck, spine, eyes and ears. Table~\ref{tab:detectors_err} presents the MPJPE and standard deviation for both visible and occluded keypoints. Occlusions were further categorized into two groups: self-occlusions, where a subject occludes themselves, and external occlusions, caused by objects or out-of-frame positioning.

\begin{table}[t]
    \centering
    \caption{Mean and standard deviation of 2D pose errors (MPJPE in pixels) for visible and occluded keypoints (Kpts) using CPN and Detectron2 on subject S2 of BlendMimic3D.}
    \vspace{0.2cm}
    \resizebox{0.5\textwidth}{!}{
    \begin{tabular}{|l|c c | c c |}
        \cline{2-5}
        \multicolumn{1}{c|}{} & \multicolumn{2}{c|}{\textbf{CPN}~\cite{cpn}} & \multicolumn{2}{c|}{\textbf{Detectron2}~\cite{detectron2}}  \\
         \multicolumn{1}{c|}{}& \textbf{Mean} & \textbf{Std} & \textbf{Mean} & \textbf{Std}\\ 
         \hline
         \rowcolor{Gray_B}\textbf{ Visible Kpts} & 6.11 & 3.53 & 6.84 & 4.21 \\
         \textbf{Self-Occluded Kpts} & 11.47 & 4.72 & 14.27 & 6.28 \\
         \rowcolor{Gray_B} \textbf{Externally-Occluded Kpts} &  19.77 & 16.65 & 28.07 & 23.74 \\ 
         \hline
    \end{tabular}
    }
    \label{tab:detectors_err}
\end{table}

Based on the results in Table~\ref{tab:detectors_err}, externally-occluded keypoints consistently exhibit greater standard deviation in 2D detection, with standard deviations reaching \(\SI{16.65}{px}\) (CPN) and \(\SI{23.74}{px}\) (Detectron2). For a 1000-pixel frame, this corresponds to approximately 1.66--\SI{2.37}{\%} of the image resolution. Visible keypoints showed lower standard deviations, 3.53 to \SI{4.21}{px}, corresponding to normalized errors of 0.35--\SI{0.42}{\%}. Interestingly, self-occluded keypoints exhibit intermediate error levels. This pattern suggests that 2D detectors are relatively robust to self-occlusions. This can be attributed to the prevalence of such occlusion patterns in standard human pose datasets used during training, such as COCO~\cite{coco}. Both detectors exhibit more than twice the mean error for self-occluded keypoints compared to visible ones, and over three times the error for externally-occluded keypoints, highlighting occlusion as key factor in error amplification of 2D HPE. 

Additionally, as shown in Table~\ref{tab:detectors_err}, the standard deviation of 2D detection errors for visible keypoints is generally below \(\SI{1}{\%}\) of the image resolution, while occluded keypoints show a higher spread, reaching up to \(\SI{2.37}{\%}\). Consequently, we interpret values of \(\sigma=\{0.001, 0.005, 0.01\}\)
as representative of noise from visible keypoints, and values \(\sigma=\{0.03, 0.05\}\) as simulating the broader uncertainty caused by occlusions. 

\subsection{Protocol 1: Adding Noise to Occluded Keypoints} \label{sec:occl_severity}
To assess how 3D HPE models handle increasing occlusion errors, we evaluated their performance using Protocol 1 (Section~\ref{sec:protocol1}), where zero-mean Gaussian noise is applied to keypoints labeled as \textit{occluded} in the BlendMimic3D annotations. The selected models span four categories: convolutional-based~\cite{vp3d}, transformer-based~\cite{pstmo,mixste,pfv2}, diffusion-based~\cite{d3dp,diffupose,finepose}, and occlusion-aware~\cite{dtf,t3d-cnn} methods. 

To better structure our analysis, we divide the discussion into two parts. First, we analyze models without explicit occlusion handling mechanisms (convolutional, transformer-based, and diffusion-based). Then, we study occlusion-aware models, designed to handle occluded keypoints through architectural or training strategies.

\subsubsection{Models Without Occlusion Awareness}
\begin{figure}[t]
    \centering
    \begin{minipage}{0.245\linewidth}
        \centering
        \includegraphics[width=\linewidth]{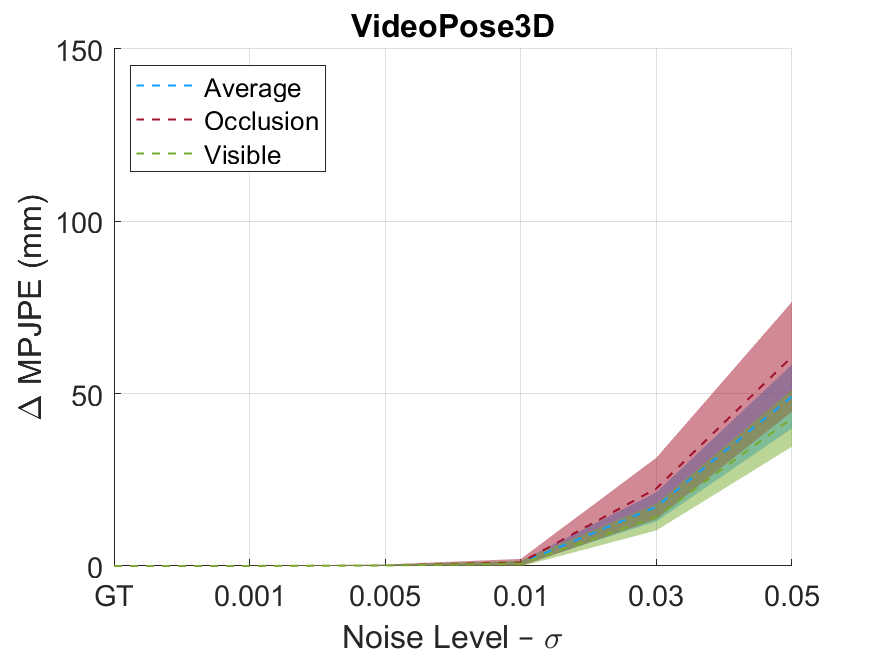}
        \subcaption{}
    \end{minipage}
    \hfill
    \begin{minipage}{0.245\linewidth}
        \centering
        \includegraphics[width=\linewidth]{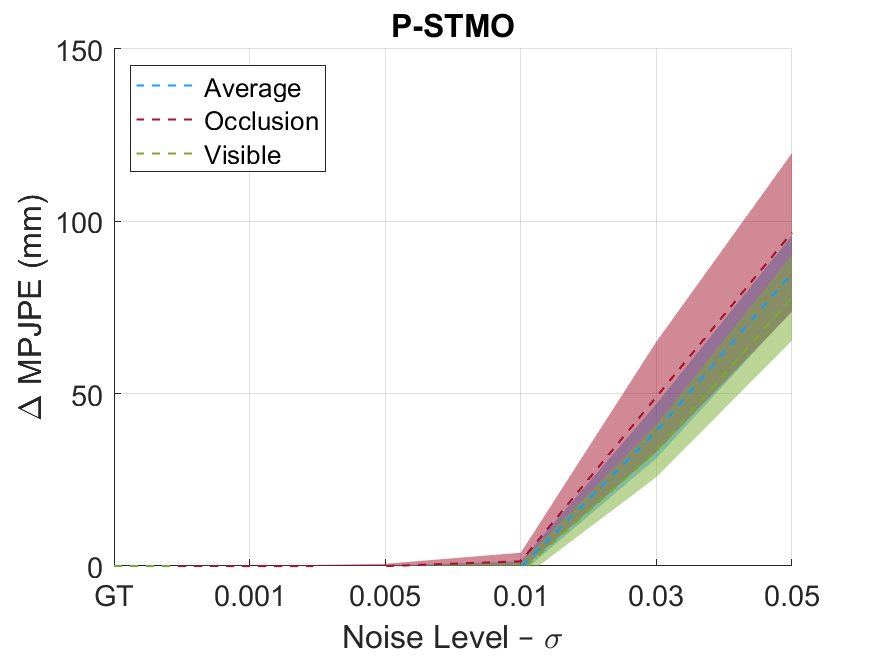}
        \subcaption{}
    \end{minipage}
    \hfill
    \begin{minipage}{0.245\linewidth}
        \centering
        \includegraphics[width=\linewidth]{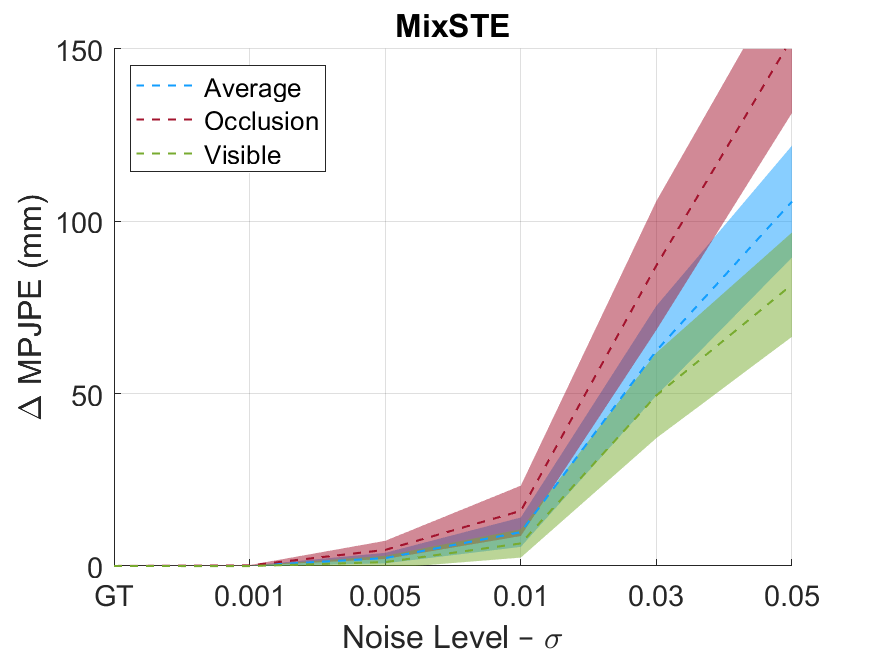}
        \subcaption{}
    \end{minipage}
     \hfill
    \begin{minipage}{0.245\linewidth}
        \centering
        \includegraphics[width=\linewidth]{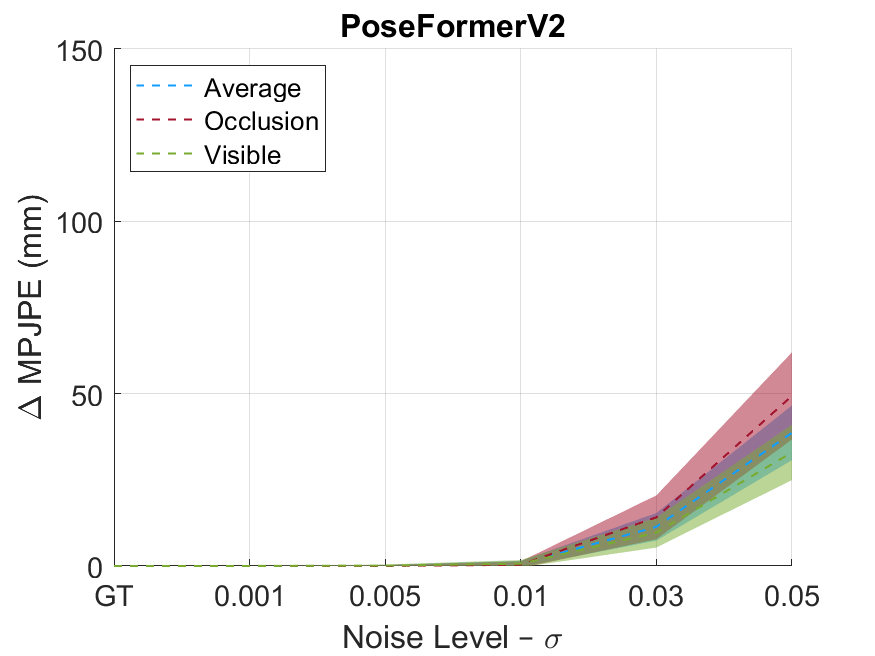}
        \subcaption{}
    \end{minipage}
     \hfill
    \begin{minipage}{0.245\linewidth}
        \centering
        \includegraphics[width=\linewidth]{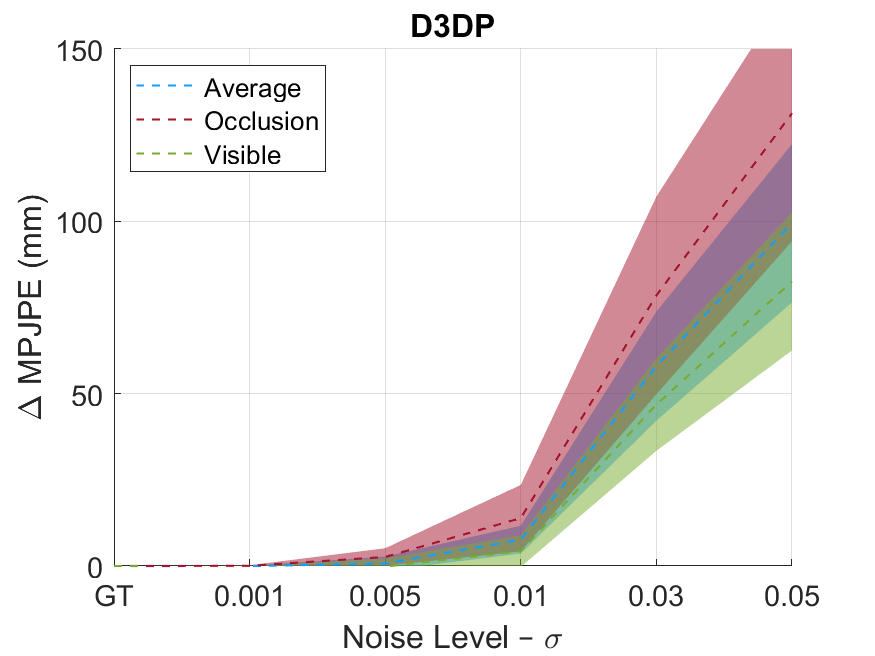}
        \subcaption{}
    \end{minipage}
     \hfill
    \begin{minipage}{0.245\linewidth}
        \centering
        \includegraphics[width=\linewidth]{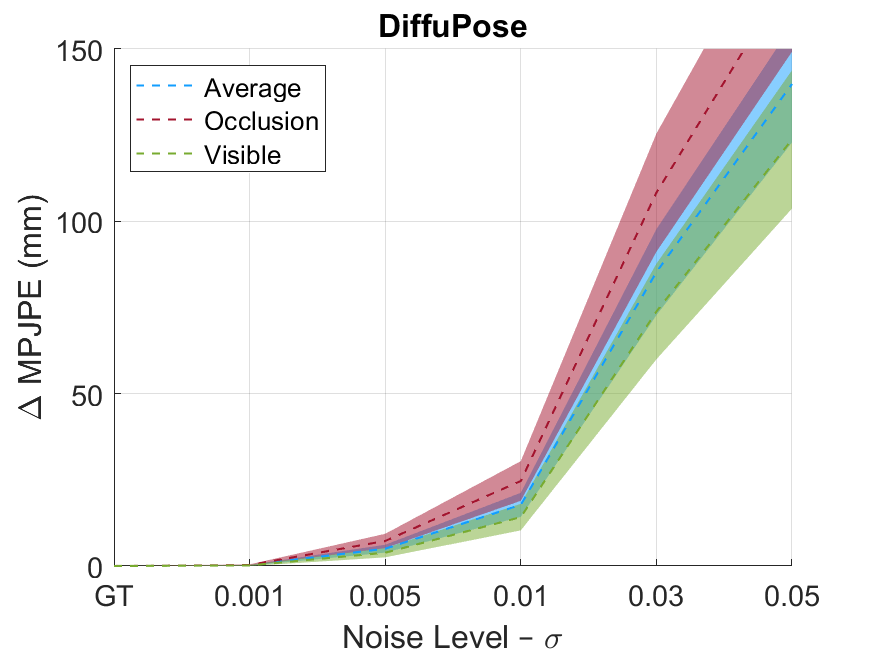}
        \subcaption{}
    \end{minipage}
     \hfill
    \begin{minipage}{0.245\linewidth}
        \centering
        \includegraphics[width=\linewidth]{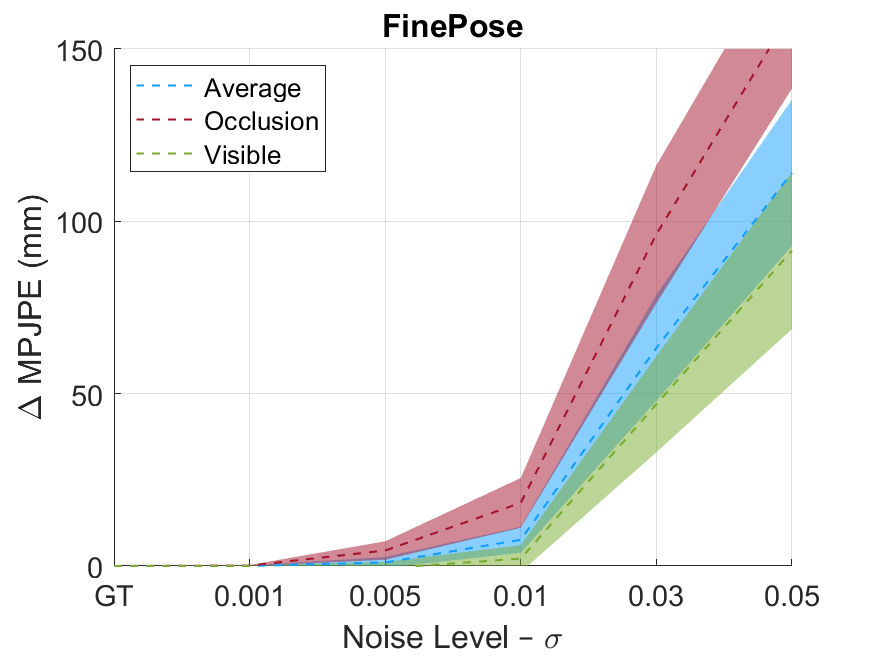}
        \subcaption{}
    \end{minipage}
     \hfill
    \begin{minipage}{0.245\linewidth}
        \centering
        \includegraphics[width=\linewidth]{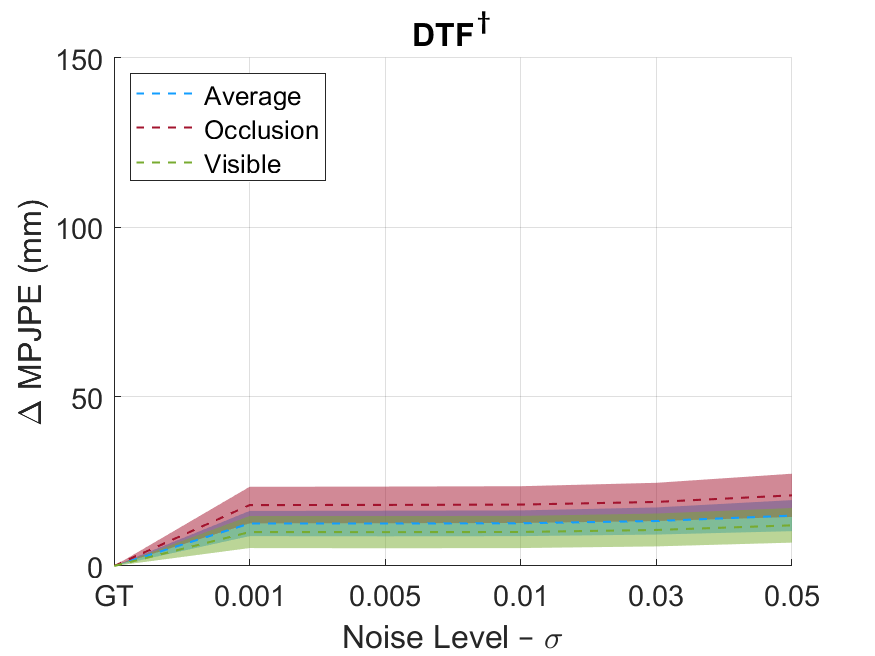}
        \subcaption{}
    \end{minipage}
    \caption{Normalized MPJPE across increasing noise levels. Each curve represents the MPJPE for overall, occluded, and visible keypoints, normalized by the respective model's baseline performance using 2D ground-truth poses. (a) VideoPose3D, (b) P-STMO, (c) MixSTE, (d) PoseFormerv2, (e) D3DP, (f) DiffuPose, (g) FinePose, and (h) DTF\(^\dagger\). These plots align with the quantitative results presented in Table~\ref{tab:noise_level}.}
    \label{fig:normalized_mpjpe_sigmas}
\end{figure}

Table~\ref{tab:noise_level} reports the MPJPE results for all evaluated models except T3D-CNN~\cite{t3d-cnn}. We exclude T3D-CNN from this comparison because the model was originally designed to treat occluded keypoints as missing, setting their 2D coordinates to zero. This behavior is incompatible with our benchmark setting, where occluded keypoints retain noisy values, and leads to invalid or misleading results. For this reason, we evaluate T3D-CNN separately in the following subsection, where we analyze the model with its masking-based occlusion mechanism.

\begin{table*}[t]
    \centering
    \caption{Evaluation of 3D HPE performance for state-of-the-art models under varying levels of simulated occlusion noise on the BlendMimic3D dataset. Results are reported as MPJPE~(mm). The best results for each noise level are shown in \textbf{bold}, and the second-best are \underline{underlined}. DTF-GT corresponds to the original configuration proposed in~\cite{dtf}, where occluded keypoints are treated as missing (i.e., zero 2D coordinates) prior to interpolation. \(^\dagger\) No missing keypoints, with simulated noise for the occluded keypoints. \(^\ddagger\)No interpolation, considering the ground-truth occlusion labels. 
    }
    \vspace{0.2cm}
    \resizebox{\textwidth}{!}{
    \begin{tabular}{l c| c | c c c c c |}
    \cline{3-8}
        \multicolumn{2}{c|}{} &  &\multicolumn{5}{c|}{\textbf{Noise Level}}  \\
        \textbf{Models} & \textbf{Year} & \textbf{GT} & \boldmath{$\sigma=0.001$} & \boldmath{$\sigma=0.005$} & \boldmath{$\sigma=0.01$} & \boldmath{$\sigma=0.03$} & \boldmath{$\sigma=0.05$} \\ 
        \hline
        
        \rowcolor{Gray_B} VideoPose3D~\cite{vp3d} & 2019 & \(73.94\pm4.85\) & \(73.93\pm4.69\) & \(74.09\pm4.70\) & \underline{\(74.77\)}\(\pm4.74\) & \underline{\(91.11\)}\(\pm4.90\) & \(123.12\pm7.99\)\\
        
        P-STMO~\cite{pstmo} & 2022 & \(75.44\pm4.67\) & \(75.41\pm4.52\) & \(74.95\pm4.67\) & \(75.02\pm4.93\) & \(114.58\pm7.36\) & \(160.41\pm9.87\) \\
        
        \rowcolor{Gray_B} MixSTE~\cite{mixste} & 2022 &  \(\mathbf{69.46} \pm 4.61\)& \(\mathbf{69.51}\pm4.50\) & $71.76\pm5.05$ & $79.27\pm6.68$ & $131.94\pm14.20$ & $175.21\pm17.03$\\
        
        PoseFormerV2~\cite{pfv2} & 2023& $73.20\pm4.23$& $73.20\pm4.10$& $73.28\pm4.24$& \(\mathbf{73.79}\pm4.67\)& \(\mathbf{84.57}\pm5.98\)&  $111.85\pm8.28$\\
        
        \rowcolor{Gray_B} D3DP~\cite{d3dp}  & 2023 &$69.84\pm3.74$ & $69.81\pm3.61$& \(\mathbf{70.57}\pm4.24\)&$77.48\pm5.97$ &$127.82\pm16.80$ & $169.29\pm23.58$\\
        
        DiffuPose~\cite{diffupose} & 2023 & $83.45\pm 6.99$ & $83.66\pm6.72$ & $88.48\pm6.69$ & $101.20\pm6.99$ & $168.71\pm12.68$ & $223.32\pm16.59$\\
        
        \rowcolor{Gray_B} FinePose~\cite{finepose} & 2024 & \underline{$69.75$}\(\pm8.87\) & \underline{$69.78$}\(\pm8.65\) & \underline{$70.78$}\(\pm9.67\) & $77.34\pm11.12$ & $133.05\pm18.84$ & $183.86\pm23.95$\\

         DTF-GT~\cite{dtf} & 2024 &$ 80.83\pm 3.69$ & --- & --- & --- & --- & --- \\ 
        \hline
        \hline
         \rowcolor{Gray_B} DTF\(^\dagger\)~\cite{dtf} & 2024 &$ 80.83\pm 3.69$ & $93.37 \pm 3.92$ & $93.37 \pm 3.95$ &$93.44 \pm 3.97$ & $94.11 \pm 4.24$ &   \(\mathbf{95.70}\pm 5.02\) \\
         
         DTF\(^\dagger\)\(^\ddagger\)~\cite{dtf} & 2024 &$ 80.83\pm 3.69$ & $93.39 \pm 3.92$ & $93.37 \pm 3.96$ &$93.42 \pm 4.01$ & $94.14 \pm 4.29$ & \underline{$95.79$}\(\pm 4.89\) \\
    \end{tabular}
    }
    \label{tab:noise_level}
\end{table*}

Despite Gaussian noise being zero-mean, the models included in Table~\ref{tab:noise_level} exhibit clear sensitivity to increasing standard deviation. This indicates that the evaluated models are vulnerable to noise provided by 2D detectors when a keypoint is occluded.

Focusing on models without occlusion-awareness~\cite{vp3d,mixste,pstmo,pfv2,d3dp,diffupose,finepose}, the performance trends observed in Table~\ref{tab:noise_level} can be largely explained by the architectural designs of each model. VideoPose3D, a convolution-based baseline, maintains stable accuracy at low-to-moderate noise levels but lacks explicit mechanisms to handle severe occlusions, leading to a sharper decline under high noise. 

P-STMO, despite its transformer-based design, shows limited robustness, likely due to its reliance on masked modeling strategies that may not generalize to real occlusion patterns. MixSTE, which alternates spatial and temporal transformer blocks, performs best in clean settings but degrades quickly under noise, suggesting its structured modeling may be sensitive to keypoint corruption. PoseFormerV2, in contrast, demonstrates strong resilience, especially at moderate noise levels, likely due to its frequency-domain temporal encoding, which helps mitigate the impact of occluded keypoint noise and supports more balanced performance.

D3DP, a diffusion-based method, maintains strong performance under moderate noise, highlighting the advantage of probabilistic modeling. It uses MixSTE as a denoiser backbone, conditioned only on the 2D input pose, which guides every step. As a result, noise in occluded keypoints significantly degrades their predictions. DiffuPose, while also diffusion-based, consistently underperforms. It uses a Graph Convolutional Network (GCN) denoiser, likewise conditioned solely on the 2D input pose, which may limit its robustness to noisy inputs. FinePose achieves competitive performance through fine-grained prompt conditioning, incorporating action and kinematic-aware guidance. Nonetheless, its denoiser is also conditioned on 2D poses. Thus, the lack of explicit mechanisms for handling occlusions in these diffusion-based models results in degraded performance at higher noise levels (\(\sigma\)). Moreover, these models are computationally expensive, requiring iterative sampling during inference, which further limits their practicality in real-time.

Overall, the results in Table~\ref{tab:noise_level} suggest that PoseFormerV2 is particularly resilient to moderate occlusion noise, outperforming both older models like VideoPose3D and newer diffusion-based approaches such as FinePose. 

As a complementary analysis to the results presented in Table~\ref{tab:noise_level}, Figure~\ref{fig:normalized_mpjpe_sigmas} shows the normalized MPJPE, computed as the difference between each noise level and the baseline (i.e., using ground truth 2D poses), for three error types: overall pose error (average MPJPE), occluded keypoint error, and visible keypoint error. This visualization highlights how model performance degrades under increasing occlusion noise and reveals differences in sensitivity to visible versus occluded keypoints. 

Across all models without occlusion-awareness, errors on occluded keypoints increase more steeply than on visible ones, revealing the common lack of explicit occlusion modeling. MixSTE and FinePose show the greatest sensitivity to noise, particularly in occluded keypoints, where errors rise sharply relative to visible keypoints. 

DiffuPose suffers the most degradation across all error types, indicating poor inference robustness. D3DP maintains relative stability for visible keypoints but is notably affected by noise in occluded keypoints. 

VideoPose3D, P-STMO and PoseFormerV2 remain relatively stable under mild noise (\(\sigma<0.01\)). Among them, P-STMO degrades faster, while VideoPose3D shows a more abrupt error increase as noise intensifies, and PoseFormerV2 demonstrates the highest resilience.

Collectively, these models are sensitive to occluded keypoints. This sensitivity is especially evident in diffusion-based architectures since their denoisers are conditioned directly on the 2D input at every iteration, any corruption introduced by occlusions propagates through the entire sampling process. Future diffusion-based methods should explore alternative conditioning mechanisms, uncertainty-aware denoisers, or hybrid formulations. In contrast, PoseFormerV2’s resilience highlights the potential of frequency-domain representations, which suppress high-frequency jitter produced by occlusions. Nonetheless, all evaluated models show that existing training strategies are insufficient to prepare models for real-world scenarios. Occlusion-aware learning, explicit visibility modeling, or architectures capable of reasoning over uncertain inputs could improve robustness.

\subsubsection{Occlusion-Aware Models}

Unlike the models discussed previously, DTF~\cite{dtf} and T3D-CNN~\cite{t3d-cnn} incorporate explicit mechanisms to address occlusions. These models are evaluated separately to better understand how occlusion-aware components.

\textbf{DTF Performance}: 
DTF handles occlusion through preprocessing and architecture design. During preprocessing, occluded keypoints have their 2D coordinates set to zero (i.e., treated as missing) and are temporally interpolated using the nearest visible instance of the same keypoint within a window spanning the previous 40 frames and the next 40 frames (81-frame window). If no such instance exists, the keypoint remains zero. Each keypoint receives also a confidence score reflecting its visibility or interpolation uncertainty. These (x, y, confidence) inputs are then fed to the model. During inference, the confidence values inform the transformer about the reliability of each joint. The architecture itself is a dual-view transformer that generates two intermediate 3D hypotheses, refines them with self-attention, and fuses them with cross-view attention to produce the final 3D pose.

\begin{figure}[t]
    \centering
    \includegraphics[width=0.6\linewidth]{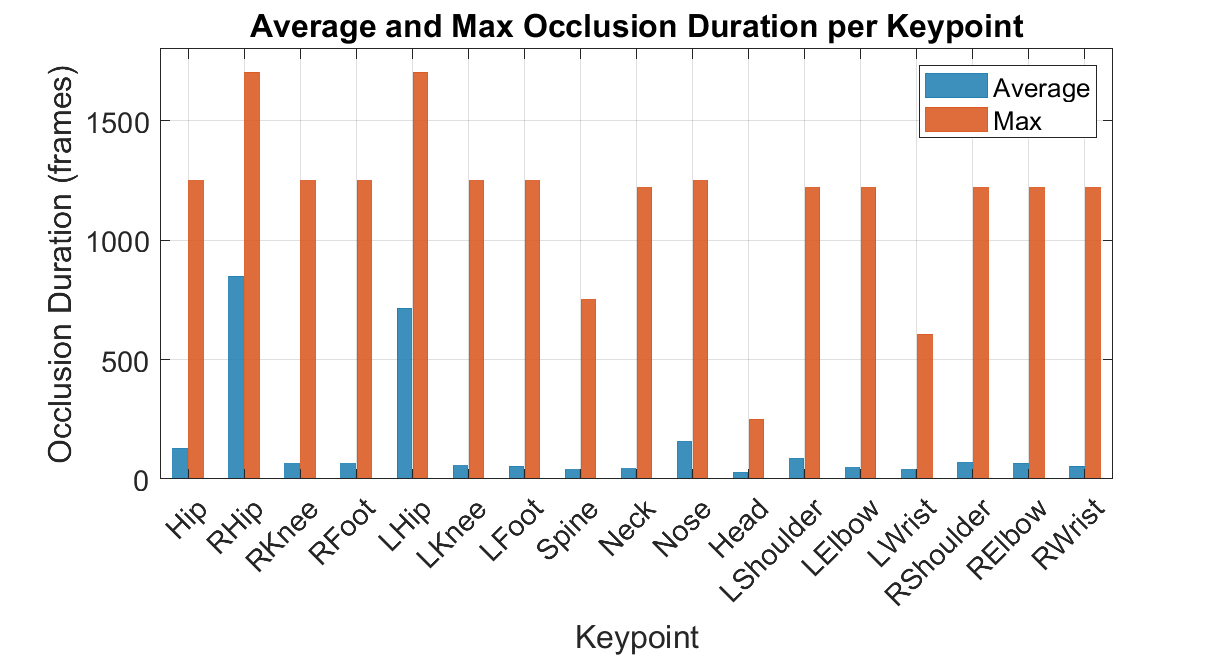}
    \caption{Analysis of occlusion duration per keypoint across actions. Based on ground-truth occlusion labels from BlendMimic3D (subject S2), this figure reports the average and maximum occlusion durations for each keypoint.}
    \label{fig:bm3d_occl_duration}
\end{figure}

During training the authors randomly occlude 16 keypoints per frame, and their positions are then estimated through interpolation. However, their assumption of having only one visible keypoint at a time and random occlusion patterns is unrealistic. In our evaluation using BlendMimic3D, which provides naturally occurring occlusion annotations, keypoints can remain occluded for longer durations than DTF expects, as can be seen in Figure~\ref{fig:bm3d_occl_duration}. Such prolonged occlusions render DTF's interpolation strategy less effective, indicating that its training approach is inadequate for real-world occlusions.

To bridge this gap between simulated and realistic occlusion, we evaluated two modified variants of the original DTF model (DTF-GT): (1) DTF\(^\dagger\), which uses full 2D poses (including noisy occluded keypoints) and applies interpolation using the ground-truth occlusion labels, and (2) DTF\(^{\dagger
\ddagger}\), which omits interpolation entirely and instead takes the noisy poses with occlusion labels as input. Table~\ref{tab:DTF_legends} summarizes the differences among these variants in terms of input preparation and processing. These correspond directly to the DTF results reported in Table~\ref{tab:noise_level}.

\begin{table}[t]
    \centering
    \caption{DTF model configurations used in the occlusion robustness study, indicating use of occlusion labels, zeroed 2D keypoints, interpolation, and lifting. Variants help assess the role of each component.}
    \vspace{0.2cm}
    \resizebox{0.5\textwidth}{!}{
    \begin{tabular}{|c|c|c|c|c|}
    \cline{2-5}
         \multicolumn{1}{c|}{}& \textbf{Occl. Labels} & \textbf{Zero 2D Coord.} & \textbf{Interp.} &  \textbf{Lifting} \\
         \hline
          \rowcolor{Gray_B}\textbf{DTF-GT}& \checkmark & \checkmark & \checkmark & \checkmark  \\
         \textbf{DTF}\(^\dagger\)& \checkmark & \ding{53} & \checkmark& \checkmark  \\
          \rowcolor{Gray_B}\textbf{DTF}\(^\dagger\)\(^\ddagger\)& \checkmark & \ding{53}&\ding{53} &\checkmark  \\
         \hline
    \end{tabular}
    }
    \label{tab:DTF_legends}
\end{table}

In Table~\ref{tab:noise_level}, the DTF-GT\ configuration fails when occlusion is present (\(\sigma=\{0.001, 0.005, 0.01, 0.03, 0.05\}\)). This occurs because the model treats occluded keypoints as missing (i.e., zeroed 2D coordinates) and attempts to recover it through interpolation within a \(\pm40\)-frame window. However, as shown in Figure~\ref{fig:bm3d_occl_duration}, BlendMimic3D contains realistic occlusion that often last longer than 81 frames, meaning many keypoints have no visible instance within the search window. Therefore these joints remain zeroed, a scenario that almost never occurs in the original DTF training setup. Since the model is not trained to handle prolonged missing joints, it collapses and outputs the same high error  (\(301.91 \pm \SI{34.60}{mm}\)) across all noise levels. These results show that while the original interpolation strategy may be effective for synthetic occlusion patterns, it struggles with the duration and structure of realistic occlusions in BlendMimic3D, suggesting the need to extend the temporal window or replacing interpolation with a learned occlusion-aware reconstruction of the 2D pose. 

To evaluate how well DTF handles occlusions and to address the failure of the DTF-GT configuration caused by zeroed 2D inputs, we introduced two alternative configurations using the same architecture. In these variants, occluded keypoints are never zeroed, the model receives instead the raw noisy 2D poses together with occlusion labels, either with interpolation (DTF\(^\dagger\)) and without interpolation (DTF\(^{\dagger\ddagger}\)). As  shown in Table~\ref{tab:noise_level}, both configurations yield more stable predictions, even as noise increases. This resilience stems from its dual-view generation and fusion architecture, along with a confidence-aware mechanism that enables the model to down-weight unreliable keypoints and leverage temporal consistency effectively. DTF\(^\dagger\) outperforms all models without occlusion awareness  at the highest noise level (\(\text{MPJPE}=\SI{95.70}{mm}\)), with both visible and occluded keypoint errors remaining stable across noise levels, as shown in Figure~\ref{fig:normalized_mpjpe_sigmas}. While DTF performs worse at lower noise levels, its performance degrades significantly less as noise increases.

We adopt the DTF\(^\dagger\) configuration for all subsequent evaluations. This variant preserves the core components of the original model while using complete 2D poses, including occluded keypoints, rather than treating them as missing. 

\textbf{T3D-CNN Performance}: Unlike DTF, T3D-CNN is explicitly designed to operate on 2D keypoints where occluded joints are masked to zero, using a binary occlusion-guidance input to indicate which joints are missing. As a temporal dilated CNN, it relies on this mask to inpaint the missing joints and lift the sequence to 3D under a specific occlusion setting. To evaluate its robustness, we trained T3D-CNN under three different masking strategies: no masking (\textit{None}), 8 randomly occluded keypoints (\textit{Rand 8}) and 16 randomly occluded keypoints per frame (\textit{Rand 16}). Our goal was to assess the model's ability to generalize across varying numbers of missing keypoints. 

Table~\ref{tab:t3d-cnn} presents the performance of each model under multiple test conditions: 2D ground-truth input from Human3.6M  with no missing keypoints, with 4, 8, 12 or 16 randomly masked keypoints, and 2D input from BlendMimic3D dataset, where keypoints are zeroed, providing realistic occlusions, though not in the fixed random-mask format expected by T3D-CNN. 

\begin{table*}[t]
    \centering
    \caption{Evaluation of T3D-CNN under varying conditions. All inputs use 2D ground-truth keypoints in both Human3.6M (H36M) and BlendMimic3D (BM3D). \textit{>1000} indicates that the error exceeds \(\SI{1000}{mm}\). \textit{Rand 4}, \textit{Rand 8}, \textit{Rand 12} and \textit{Rand~16} refer to random occlusion of 4, 8, 12 or 16 keypoints, respectively, applied during training or testing as indicated. The best results for each test configuration are shown in \textbf{bold}.} 
    \vspace{0.2cm}
    \resizebox{0.65\textwidth}{!}{
    \begin{tabular}{| c l| c c c c c || c |}
    \cline{3-8}
        \multicolumn{2}{c|}{} & \multicolumn{6}{c|}{\textbf{Test Configuration}}  \\
        \multicolumn{2}{c|}{} & \multicolumn{5}{c||}{\textbf{H36M}} &  \\
        \multicolumn{2}{c|}{} & \textbf{No miss} & \textbf{Rand 4} & \textbf{Rand 8} & \textbf{Rand 12} & \textbf{Rand 16} & \textbf{BM3D} \\
        \hline
         \multirow{3}{*}{\rotatebox[origin=c]{90}{\textbf{Model}}} & \textbf{None}\(^\dagger\)&  \textbf{41.20} & 274.23 & 352.97&  403.38 & 457.04& 529.61\\
         &  \cellcolor{Gray_B}\textbf{Rand 8} &\cellcolor{Gray_B} 153.50  & \cellcolor{Gray_B}  \textbf{100.78} &\cellcolor{Gray_B} \textbf{56.79} &\cellcolor{Gray_B} \textbf{256.89} &\cellcolor{Gray_B}419.15 & \cellcolor{Gray_B}\textbf{314.65} \\
        &\textbf{Rand 16} & \textit{>1000} & \textit{>1000} & \textit{>1000} & 907.93 & \textbf{74.42} & \textit{>1000}\\
        \hline
        \multicolumn{8}{l}{\small  \(^\dagger\) For the \textbf{None} model, the two occlusion-guidance input channels of T3D-CNN were removed,}\\
        \multicolumn{8}{l}{\small  as the network was trained using only fully visible keypoints.} 
    \end{tabular}
    }
    \label{tab:t3d-cnn}
\end{table*}

The results indicate that T3D-CNN struggles to generalize beyond the specific masking condition it was trained on. Because the model is always trained with a fixed number of missing keypoints (e.g., exactly 16 per frame), its filters and batch-norm statistics become specialized to that sparsity level. When evaluated on unseen masking configurations, such as a different number of occlusions or the naturally varying patterns in BM3D, the occlusion-mask distribution shifts, placing the model in a regime it was never trained for, leading to large performance drops. A more robust design would require training with a mixture of occlusion patterns so the model learns to handle a broader range of scenarios. 

Notably, as shown in Table~\ref{tab:t3d-cnn}, the model trained with 8 randomly occluded keypoints (\textit{Rand 8}) maintains more stable performance across a wider range of missing keypoints when evaluated on the H36M dataset. This behavior is consistent with the fact that it performs better on BM3D than the versions trained with no masking or 16 occlusions. This also suggests that the configuration of \textit{Rand 8} is more reasonable and better aligned with the range of occlusion patterns found in BM3D, as shown in Figure~\ref{fig:bm3d_stats_b}. 

Because T3D-CNN cannot process complete or noisy 2D inputs, and lacks a mechanism to exploit real occlusion annotations, it was excluded from the noise-level evaluation presented in Table~\ref{tab:noise_level}.

\subsection{Protocol 2: Per-Keypoint Occlusion Analysis}
Results in Table~\ref{tab:noise_level} and Figure~\ref{fig:normalized_mpjpe_sigmas} show that for \(\sigma=\{0.001, 0.005\}\), many models maintain or slightly improve their performance compared to using ground-truth 2D input. This suggests that such low levels of noise reflect the standard deviation typically present in training with detected 2D poses. At \(\sigma=0.01\), the noise magnitude begins to simulate mild occlusion behavior while still partially overlapping with the visible range. Once the noise increases to \(\sigma=0.03\), the performance of the models drops more noticeably, reflecting a realistic degree of occlusion-induced perturbation. Based on this, we select \(\sigma=0.03\) as a representative occlusion noise level, simulating moderate to severe occlusions. 

\begin{figure}
    \centering
    \includegraphics[width=0.45\linewidth]{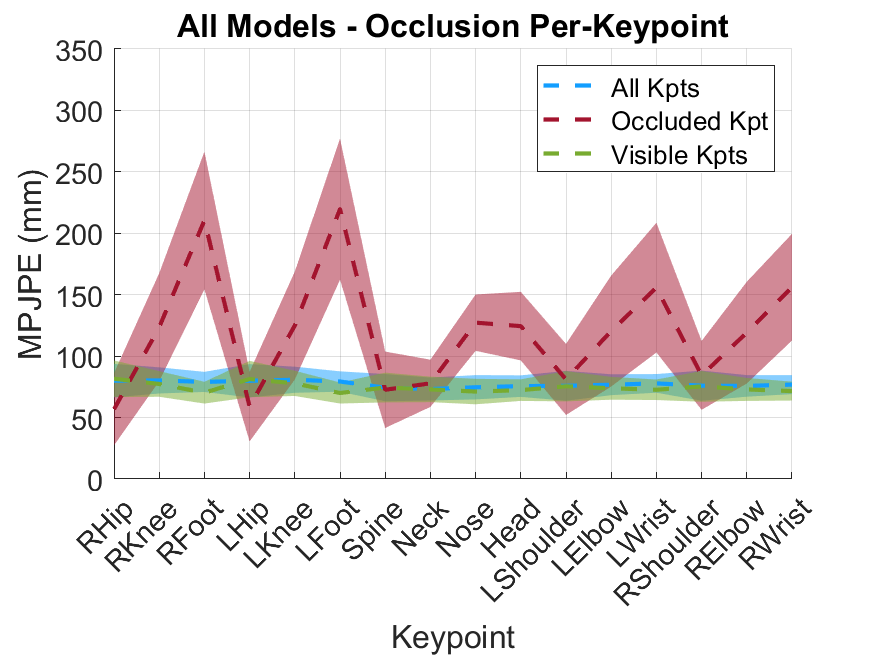}
    \caption{MPJPE per keypoint across occluded frames, where each point in the plot corresponds to the average (dashed line) and standard deviation of 10 runs where Gaussian noise (\(\sigma=0.03\)) is added solely to the corresponding keypoint.}
    \label{fig:all_mpjpe_per_occl_kpt}
\end{figure}

We now focus on identifying which keypoints are more vulnerable to occlusion and whether this behavior persists when the same keypoints are visible. So, to isolate the impact of each keypoint, in each experiment, we added zero-mean Gaussian noise (\(\sigma = 0.03\)) into one individual 2D keypoint, between frames 50 and 149 of the first 243-frame segment. This localized occlusion simulation allows us to evaluate per-keypoint error propagation in the 3D predictions. 

Figure~\ref{fig:all_mpjpe_per_occl_kpt} displays the aggregated MPJPE results across all models. We can see that a common pattern emerges: certain keypoints are inherently more difficult to estimate when occluded. Notably, the wrists (\textit{Lwrist}, \textit{Rwrist}), feet (\textit{RFoot}, \textit{LFoot}), \textit{Nose} and \textit{Head} show higher MPJPE under occlusion. These extremities are small, fast-moving, and often heavily occluded, making them particularly error-prone. In contrast, more central and structurally constrained keypoints, such as the hips, knees, shoulders, and elbows, exhibit lower errors, even under occlusion. 

\begin{figure}
    \centering
    \includegraphics[width=0.45\linewidth]{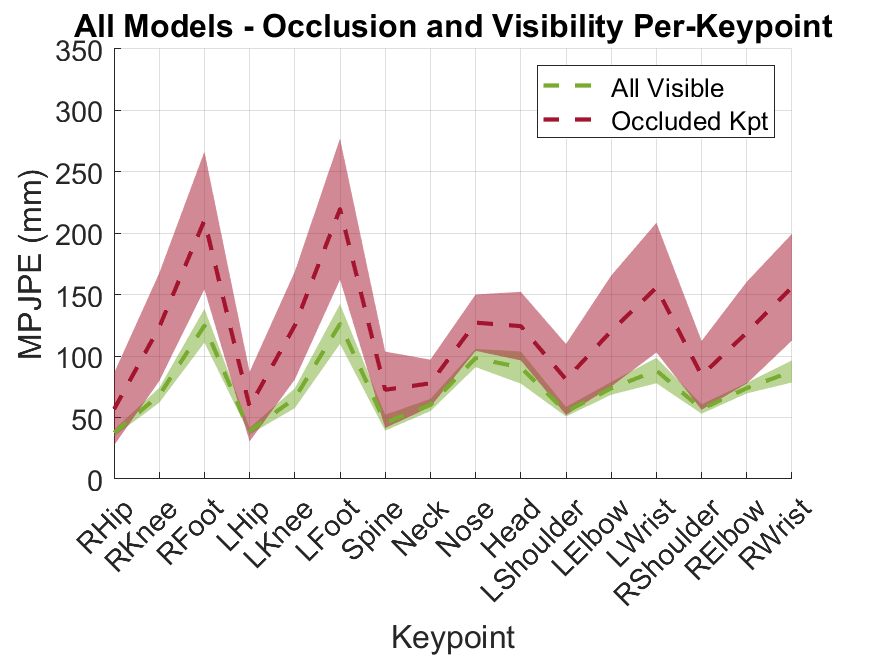}
    \caption{Comparison of MPJPE per keypoint when all keypoints are visible (2D ground truth) versus when each keypoint is individually occluded. Values are averaged across all models.}
    \label{fig:all_all_vis_one_occl}
\end{figure}

To understand whether this error pattern is purely due to occlusion or an intrinsic modeling difficulty, we compare occluded keypoint errors with their errors under fully visible conditions. Figure~\ref{fig:all_all_vis_one_occl} shows that even when all keypoints are visible, these same keypoints (e.g., wrists, feet) still exhibit higher MPJPE, indicating an inherent estimation difficulty.

\begin{figure}
    \centering
    \includegraphics[width=0.45\linewidth]{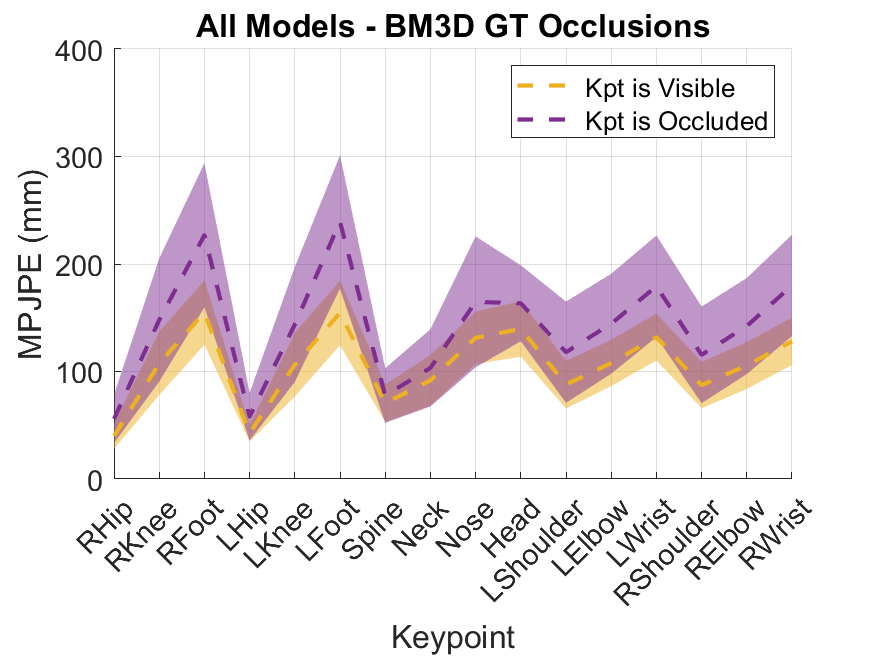}
    \caption{ MPJPE for each keypoint under realistic occlusion conditions in BlendMimic3D (BM3D). We compare errors when the keypoint is visible versus occluded across poses. These reflect the natural distribution of occlusions and keypoint visibility. Values are averaged across all models.}
    \label{fig:all_vis_vs_real_occl}
\end{figure}

We extended this analysis to realistic occlusion scenarios in BlendMimic3D using the provided labels and added Gaussian noise (\(\sigma=0.03\)). Figure~\ref{fig:all_vis_vs_real_occl} shows the MPJPE of each keypoint when visible vs. occluded. The same pattern persists: error increases with occlusion, and the most affected keypoints remain the same. This supports our conclusion that keypoints with higher degrees of freedom (knees, feet, wrists, and elbows) are consistently harder to estimate. 

A detailed per-model breakdown of all three analyses, per-keypoint occlusion (Figure~\ref{fig:all_mpjpe_per_occl_kpt}), comparison with fully visible pose (Figure~\ref{fig:all_all_vis_one_occl}), and real occlusion analysis (Figure~\ref{fig:all_vis_vs_real_occl}), is provided in the supplementary material. These results include model-specific plots of occluded vs. visible keypoint errors, enabling a deeper understanding of how each architecture responds to occlusion. Importantly, all models exhibit the same qualitative pattern described here.


\section{Beyond Occlusions: Evaluating Additional Factors} \label{sec:beyond_occl}

While our primary analysis focused on model robustness under varying levels of occlusions, it is important to consider other factors that might also contribute to performance degradation. In this section, we explore how dataset distribution shifts, particularly related to camera-subject configuration, can affect model generalization. All evaluated models were trained on Human3.6M~\cite{h36m}, a highly controlled indoor dataset, whereas our benchmark uses BlendMimic3D~\cite{bm3d}, a synthetic dataset designed to simulate more realistic and diverse occlusion scenarios. We also briefly mention other possible factors, such as velocity and action diversity, which were found to have little impact (see supplementary material).

\begin{figure}
    \centering
    \begin{minipage}{0.45\linewidth}
        \centering
        \includegraphics[width=\linewidth]{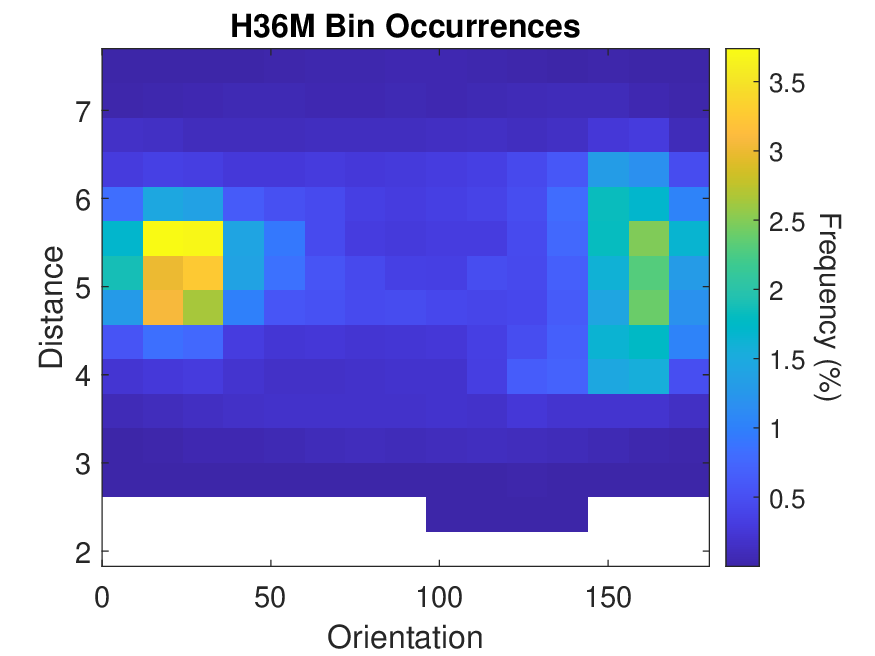}
        \subcaption{}
    \end{minipage}
    \hfill
    \begin{minipage}{0.45\linewidth}
        \centering
        \includegraphics[width=\linewidth]{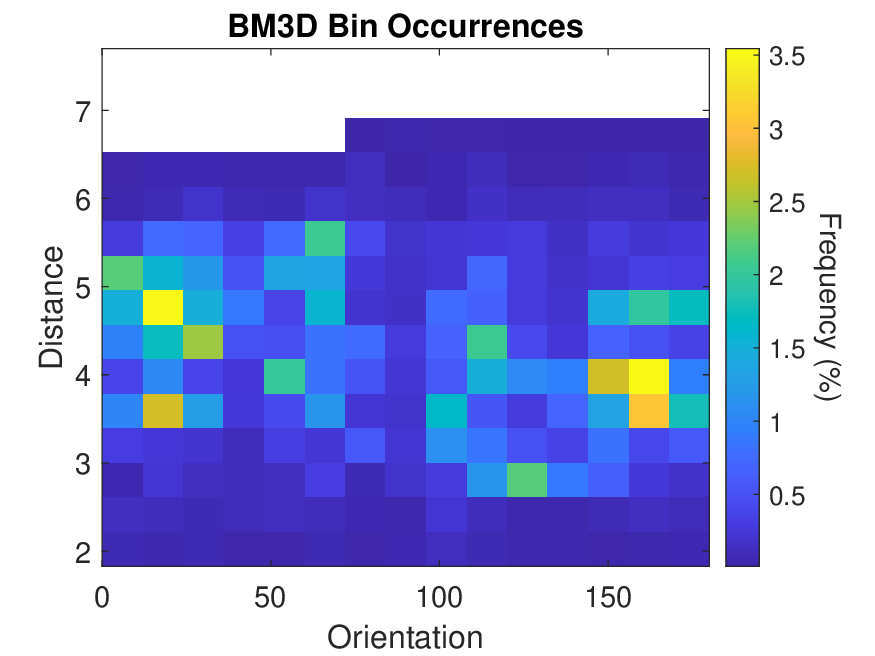}
        \subcaption{}
        \label{fig:bm3d_bins}
    \end{minipage}
    \caption{ Data distribution of subject orientation and distance to the camera in Human3.6M (a) and BlendMimic3D (b). White pixels indicate no samples.}
    \label{fig:bm3d_h36m_bins}
\end{figure}

\subsection{Domain Shift: Human3.6M vs. BlendMimic3D}

To better understand how domain shift affects model generalization, we compare the distributions of Human3.6M and BlendMimic3D in terms of the subject's orientation and distance relative to the camera. Using the 3D poses and camera parameters, we estimate the Euclidean distance between the subject's center (mean of all 3D keypoints) and the camera position. The orientation is measured as the angle between the camera's viewing direction (Z-axis) and the normal vector of the subject's shoulder-hip plane, defined using the hip and both shoulder keypoints. An orientation of \(\SI{0}{\degree}\) indicates the subject is facing away from the camera, while \(\SI{180}{\degree}\) corresponds to facing directly toward the camera.

Figure~\ref{fig:bm3d_h36m_bins} presents the data distribution of orientation and distance for training data from Human3.6M (subject S1, S5, S6, S7, and S8) and the test data from BlendMimic3D (subject S2). As shown, Human3.6M primarily captures subjects facing the camera, turning away, or viewed from the side, between 4 and 6 meters, with two dominant peaks around \(\SI{20}{\degree}\) and \(\SI{160}{\degree}\), highlighting limited diversity in viewpoint. In contrast, BlendMimic3D exhibit a broader and more uniform distribution across both orientation and distance. This makes it a comprehensive benchmark for evaluating model robustness. Notably, BlendMimic3D includes poses at closer distances, scenarios that were unseen during training.

\subsection{Camera-Related Sensitivity}

\begin{figure}[t]
    \centering
    \includegraphics[width=0.45\linewidth]{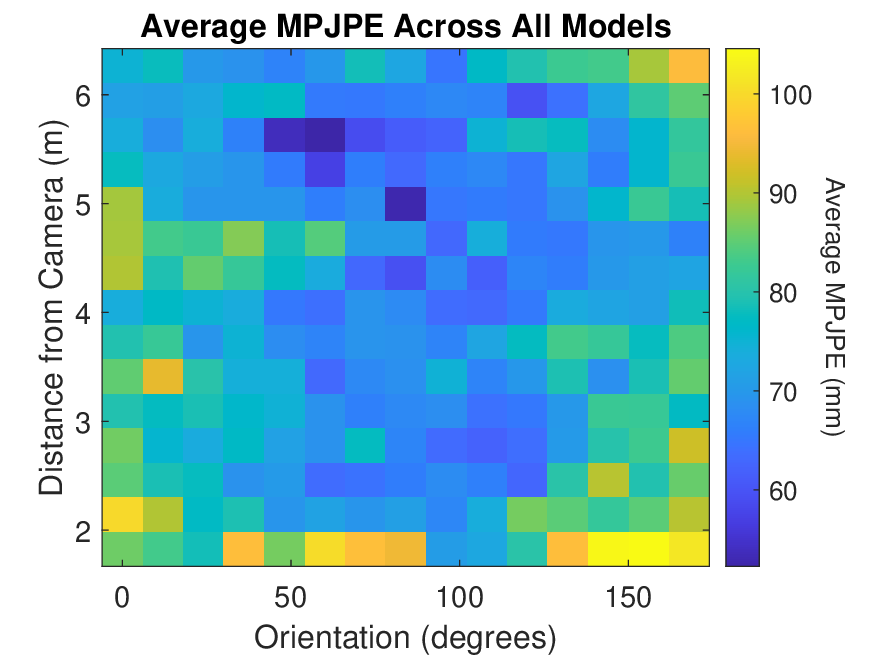}
    \caption{Average MPJPE across all models as a function of subject orientation and distance to the camera. Evaluated on BlendMimic3D with ground-truth 2D inputs.}
    \label{fig:heatmap_all_models}
\end{figure}

To investigate whether model errors correlate with subject orientation and distance to the camera, we analyzed performance heatmaps on BlendMimic3D using ground-truth 2D poses (i.e., without occlusions or detector noise). Figure~\ref{fig:heatmap_all_models} shows the average MPJPE across all models, aggregated over bins of orientation and distance.

We observe a consistent trend: performance degrades at shorter distances (below ~2.5 meters), a range underrepresented in Human3.6M (see Figure~\ref{fig:bm3d_h36m_bins}). This highlights a generalization challenge due to distribution shift in camera-subject configurations. Beyond this, orientation and distance do not exhibit a clear correlation with MPJPE when results are averaged across models. For a more detailed understanding of model-specific sensitivity patterns, including how individual architectures react to different orientations and distances, per-model heatmaps can be found in supplementary material. 

We also analyzed velocity and action variability, and concluded that they do not influence our results. This analysis is provided in the supplementary material.

\section{Conclusion}
\label{sec:conclusion}
In this study, we benchmarked SOTA 2D-to-3D HPE models under realistic occlusion scenarios using the BlendMimic3D dataset. Our focus was to evaluate model robustness to noisy 2D inputs, a common real-world challenge.

To this end, we introduced two controlled evaluation protocols. The first simulated increasing occlusion severity by adding Gaussian noise into all occluded keypoints based on ground-truth occlusion labels. The second evaluated the impact of occluding each keypoint individually, allowing us to isolate per-keypoint vulnerabilities. 

Results showed that all models degrade under occlusion. Diffusion-based models proved sensitive to noisy inputs since their denoising processes rely heavily on the 2D input pose. This highlights the need for uncertainty-aware or visibility-conditioned denoising mechanisms, as well as sampling strategies that reduce dependence on potentially corrupted inputs. Occlusion-aware models such as DTF showed improved robustness under severe occlusion but underperformed on clean inputs. Overall, no model was able to maintain consistent performance across the full occlusion spectrum, however PoseFormerV2 and DTF\(^\dagger\) exhibited the strongest resilience under moderate and severe corruption, respectively. This suggests that hybrid architectures combining frequency-domain smoothing with confidence-aware fusion should be explored. 

Protocol 2 revealed that keypoints with higher degrees of freedom, such as wrists and feet, are consistently more error-prone, even when not occluded. Therefore, future work should incorporate explicit per-joint modeling, such as local corrective branches or learned keypoint-reliability modules.

Beyond occlusions, we observed that camera-related distribution shifts between training (Human3.6M) and evaluation (BlendMimic3D) affect generalization, especially at closer distances and unseen viewpoints. We also identified and addressed evaluation inconsistencies, such as hip keypoint normalization and skeleton format mismatches, to ensure fairness.

Despite advances in 3D HPE, our results show that current models are not yet robust enough for real-world environments. Future research should focus on occlusion-aware models that generalize across diverse viewing conditions and skeleton formats, while maintaining strong performance on clean inputs.

\section{Acknowledgments}
\label{sec:Acknowledgments}
This work was supported by research grant 10.54499/2022.07849.CEECIND/CP1713/CT0001 and LARSyS funding (DOI: 10.54499/LA/P/0083/2020, 10.54499/UIDP/50009/2020 and 10.54499/UIDB/50009/2020), through \textit{Fundação para a Ciência e a Tecnologia} (FCT), and by the SmartRetail project [PRR- C645440011-00000062], through IAPMEI- Agência para a Competitividade e Inovação. Filipa Lino is supported by the FCT doctoral grant [2025.03757.BD].
\bibliographystyle{unsrt}  
\bibliography{references}  

 \newpage
\clearpage
\setcounter{page}{1}
\label{sec:supplementary}
\appendix 
\section{Per-Keypoint Occlusion Analysis (Per Model)}

\begin{figure}[t]
    \centering
    \begin{minipage}{0.245\linewidth}
        \centering
        \includegraphics[width=\linewidth]{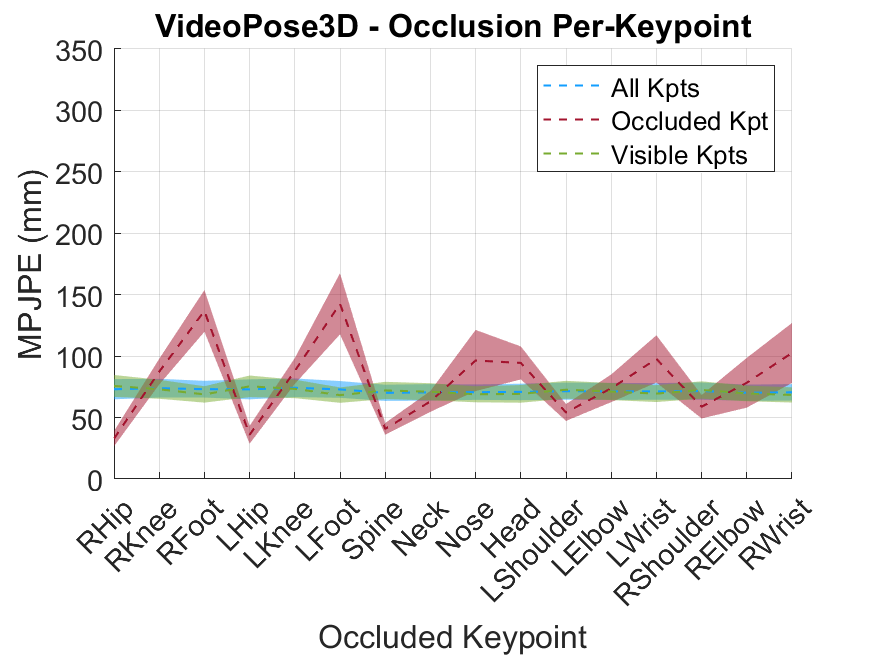}
        \subcaption{}
    \end{minipage}
    \hfill
    \begin{minipage}{0.245\linewidth}
        \centering
        \includegraphics[width=\linewidth]{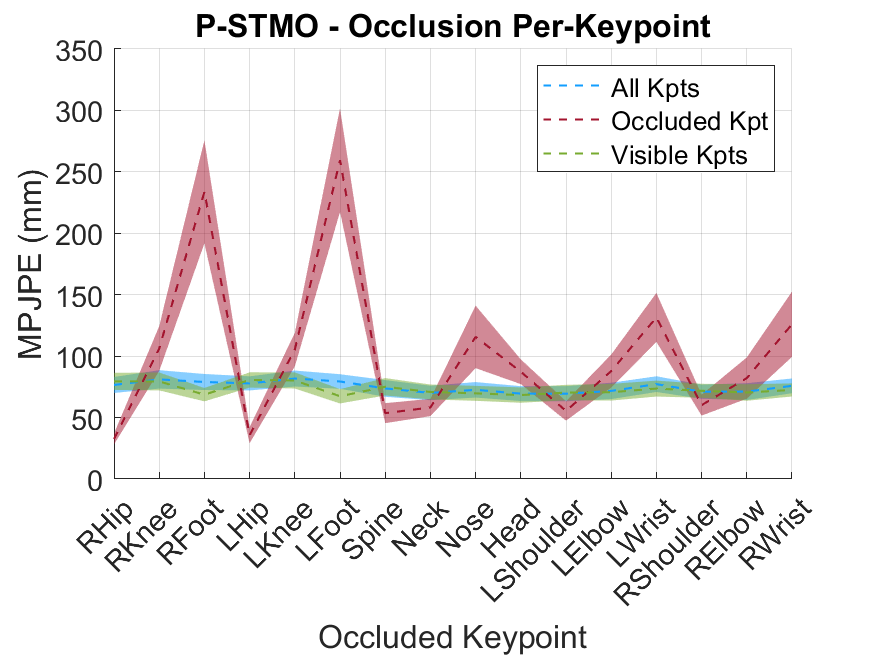}
        \subcaption{}
    \end{minipage}
    \hfill
    \begin{minipage}{0.245\linewidth}
        \centering
        \includegraphics[width=\linewidth]{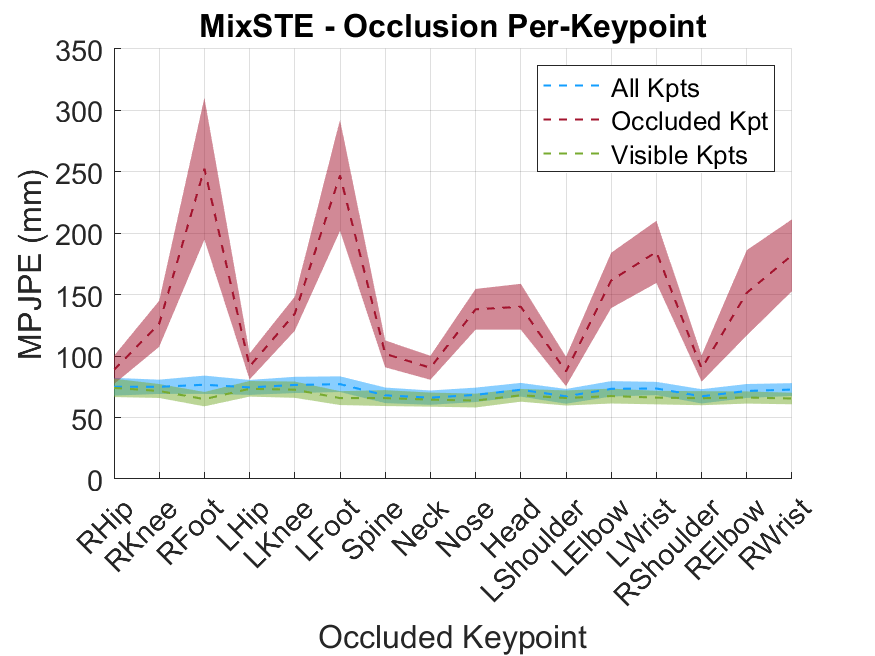}
        \subcaption{}
    \end{minipage}
     \hfill
    \begin{minipage}{0.245\linewidth}
        \centering
        \includegraphics[width=\linewidth]{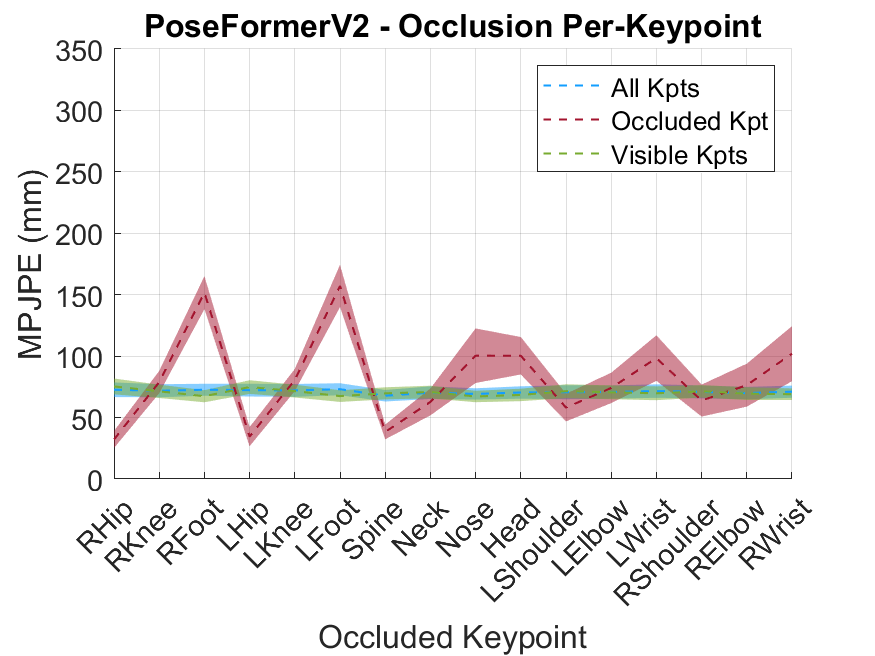}
        \subcaption{}
    \end{minipage}
     \hfill
    \begin{minipage}{0.245\linewidth}
        \centering
        \includegraphics[width=\linewidth]{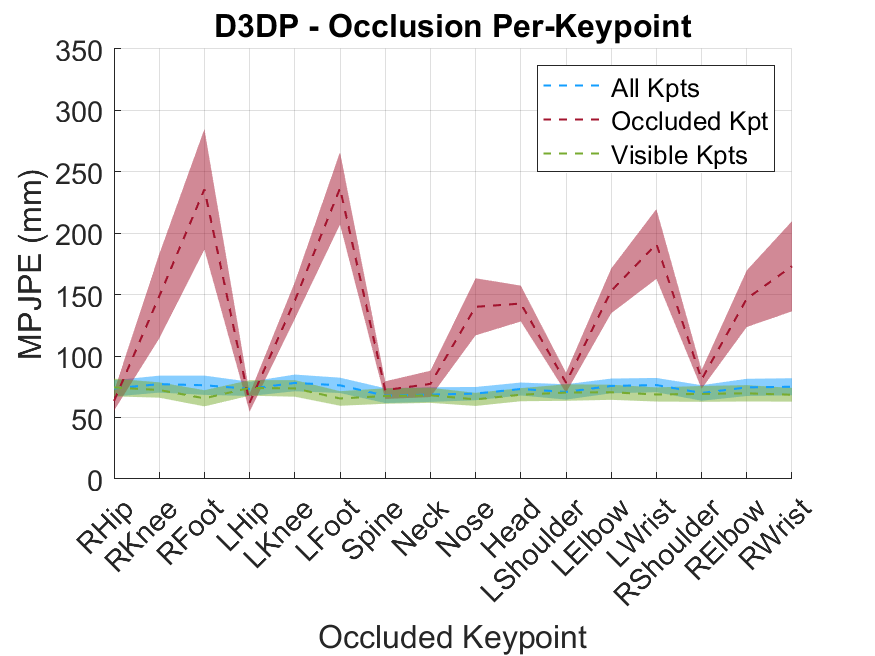}
        \subcaption{}
    \end{minipage}
     \hfill
    \begin{minipage}{0.245\linewidth}
        \centering
        \includegraphics[width=\linewidth]{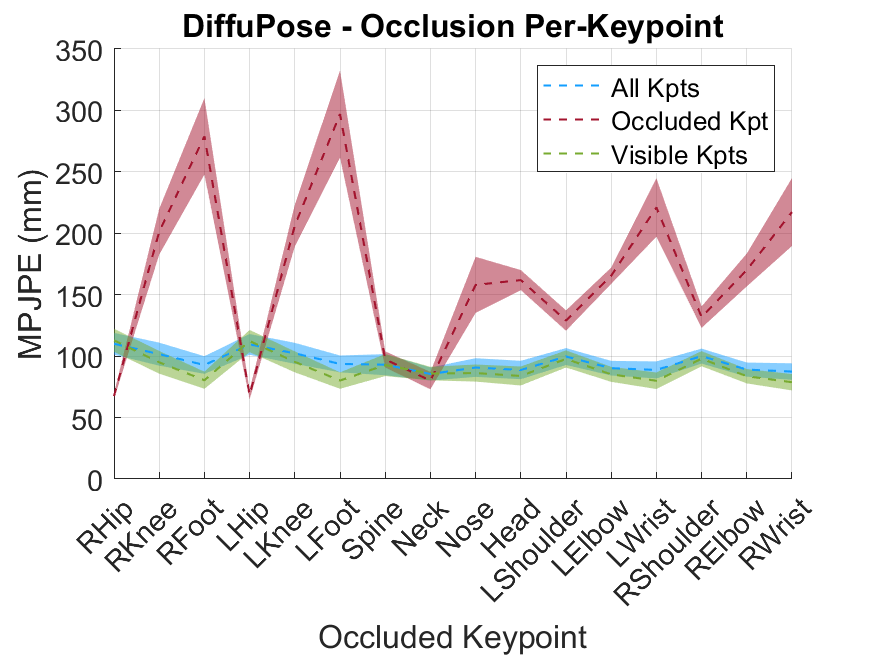}
        \subcaption{}
    \end{minipage}
     \hfill
    \begin{minipage}{0.245\linewidth}
        \centering
        \includegraphics[width=\linewidth]{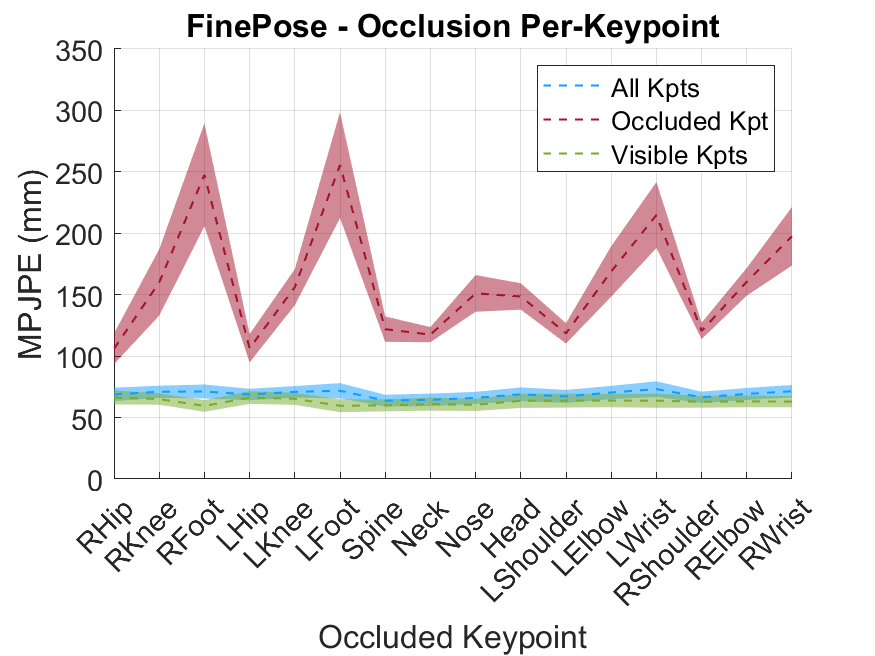}
        \subcaption{}
    \end{minipage}
     \hfill
    \begin{minipage}{0.245\linewidth}
        \centering
        \includegraphics[width=\linewidth]{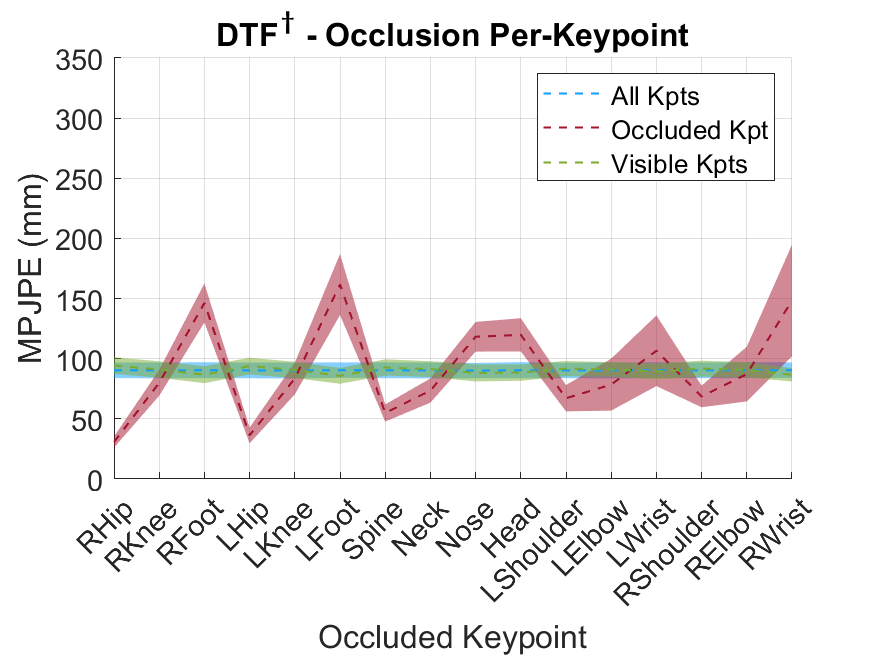}
        \subcaption{}
    \end{minipage}
    \caption{MPJPE per keypoint across occluded frames, where each keypoint is occluded independently using Gaussian noise (\(\sigma=0.03\)). We report (i) MPJPE of the occluded keypoint, (ii) MPJPE of visible keypoints, and (iii) average pose error. Models: (a) VideoPose3D, (b) P-STMO, (c) MixSTE, (d) PoseFormerv2, (e) D3DP, (f) DiffuPose, (g) FinePose, and (h) DTF\(^\dagger\).}
    \label{fig:mpjpe_per_occl_kpt}
\end{figure}

\begin{figure}[t]
    \centering
    \begin{minipage}{0.245\linewidth}
        \centering
        \includegraphics[width=\linewidth]{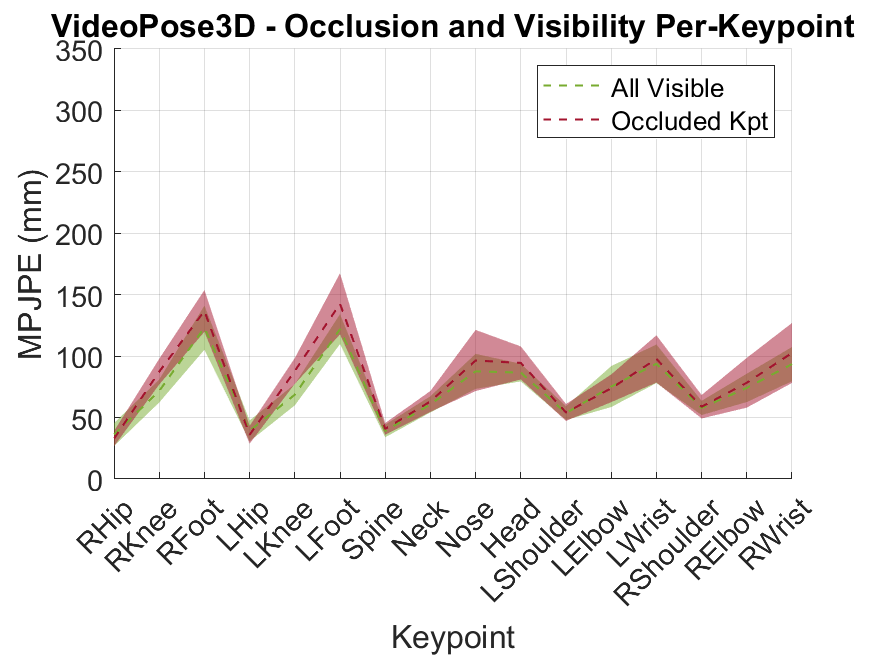}
        \subcaption{}
    \end{minipage}
    \hfill
    \begin{minipage}{0.245\linewidth}
        \centering
        \includegraphics[width=\linewidth]{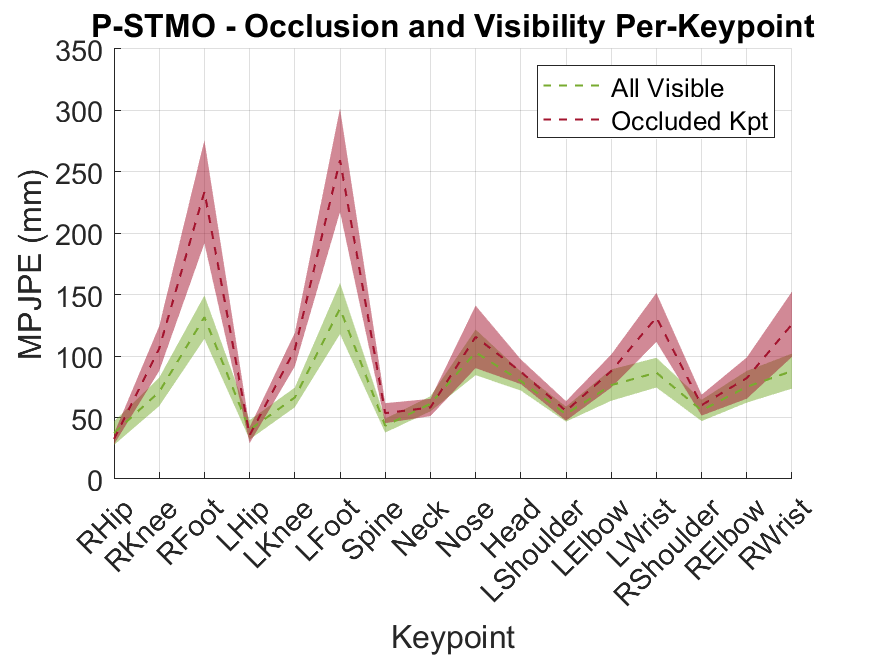}
        \subcaption{}
    \end{minipage}
    \hfill
    \begin{minipage}{0.245\linewidth}
        \centering
        \includegraphics[width=\linewidth]{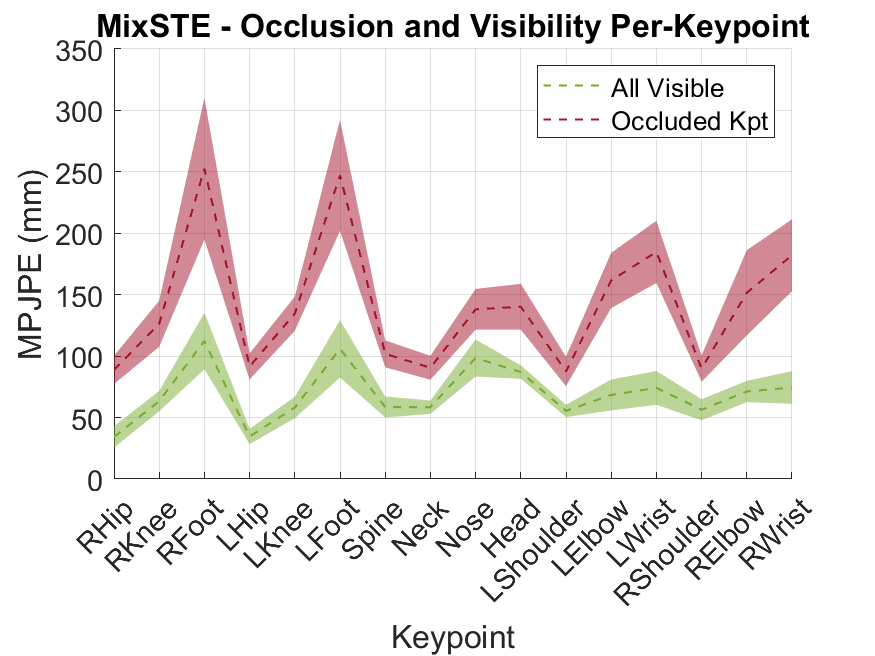}
        \subcaption{}
    \end{minipage}
     \hfill
    \begin{minipage}{0.245\linewidth}
        \centering
        \includegraphics[width=\linewidth]{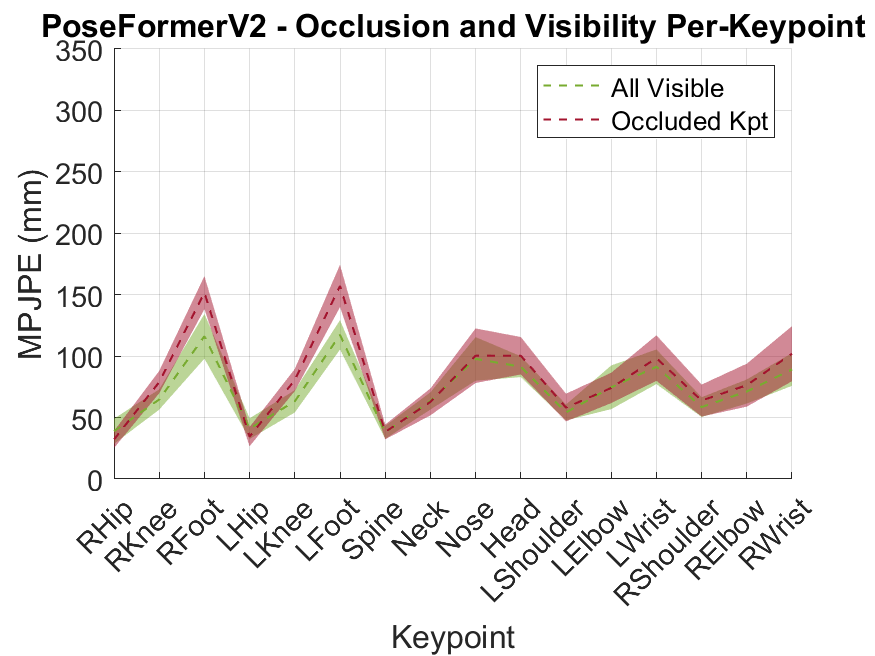}
        \subcaption{}
    \end{minipage}
     \hfill
    \begin{minipage}{0.245\linewidth}
        \centering
        \includegraphics[width=\linewidth]{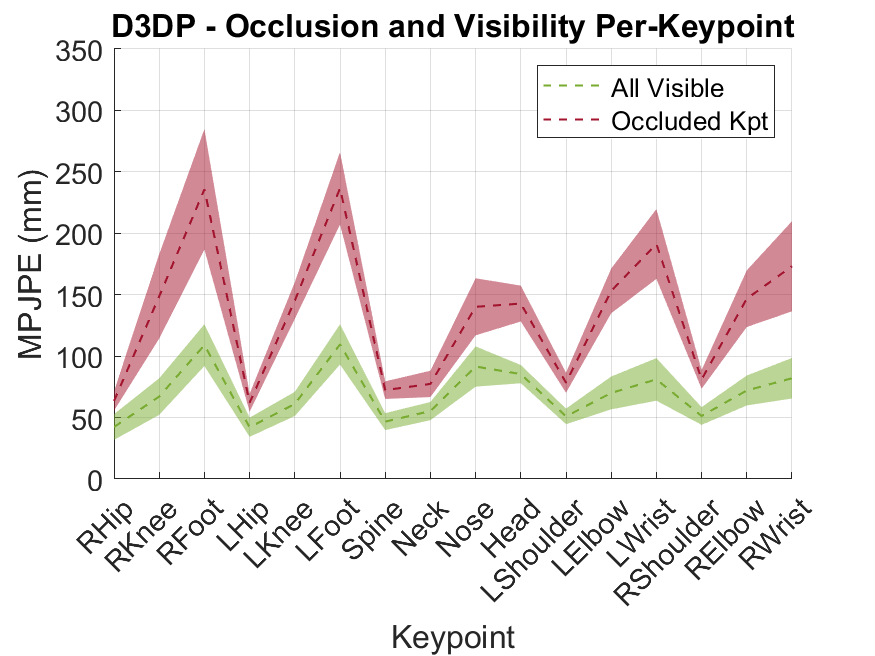}
        \subcaption{}
    \end{minipage}
     \hfill
    \begin{minipage}{0.245\linewidth}
        \centering
        \includegraphics[width=\linewidth]{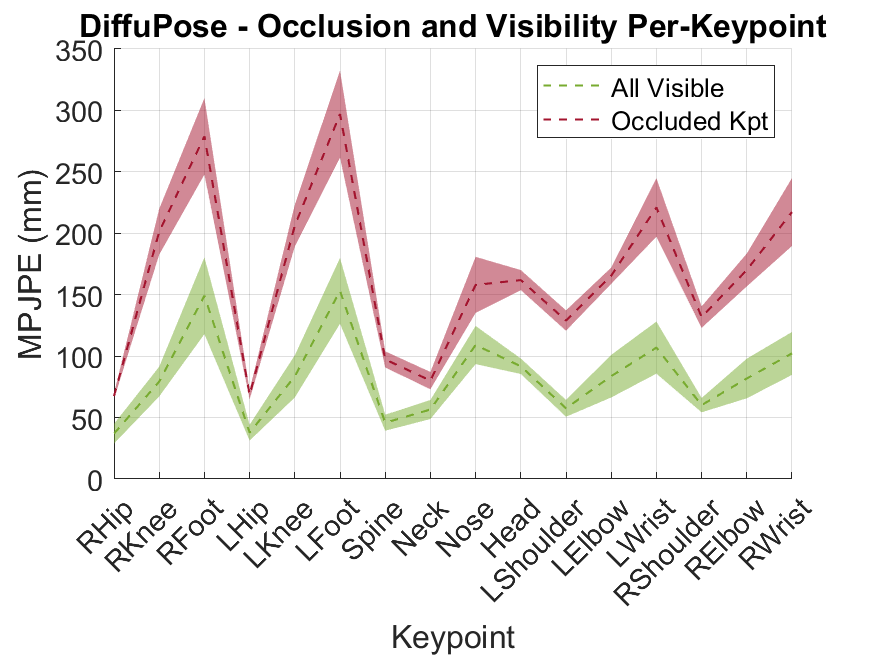}
        \subcaption{}
    \end{minipage}
     \hfill
    \begin{minipage}{0.245\linewidth}
        \centering
        \includegraphics[width=\linewidth]{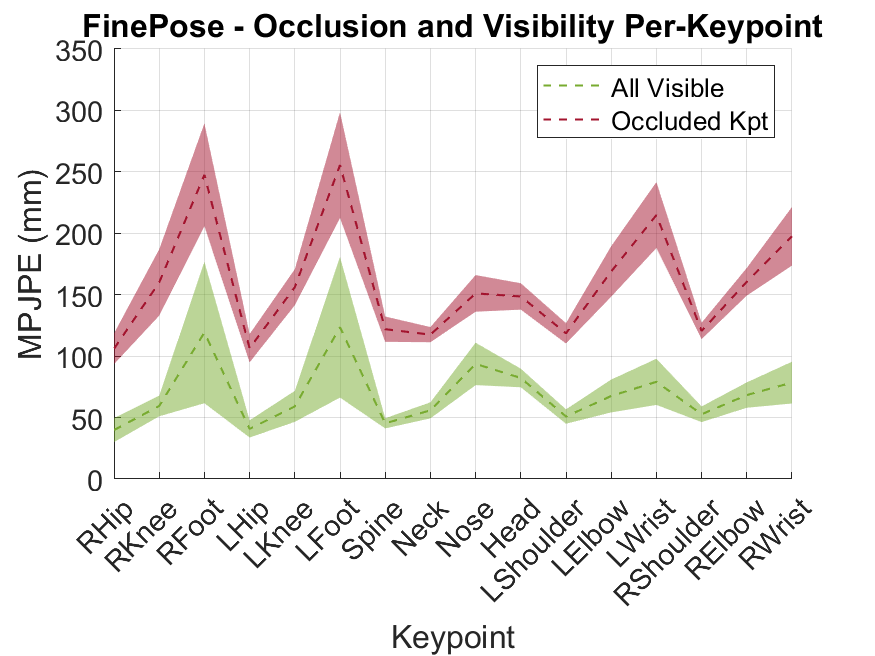}
        \subcaption{}
    \end{minipage}
     \hfill
    \begin{minipage}{0.245\linewidth}
        \centering
        \includegraphics[width=\linewidth]{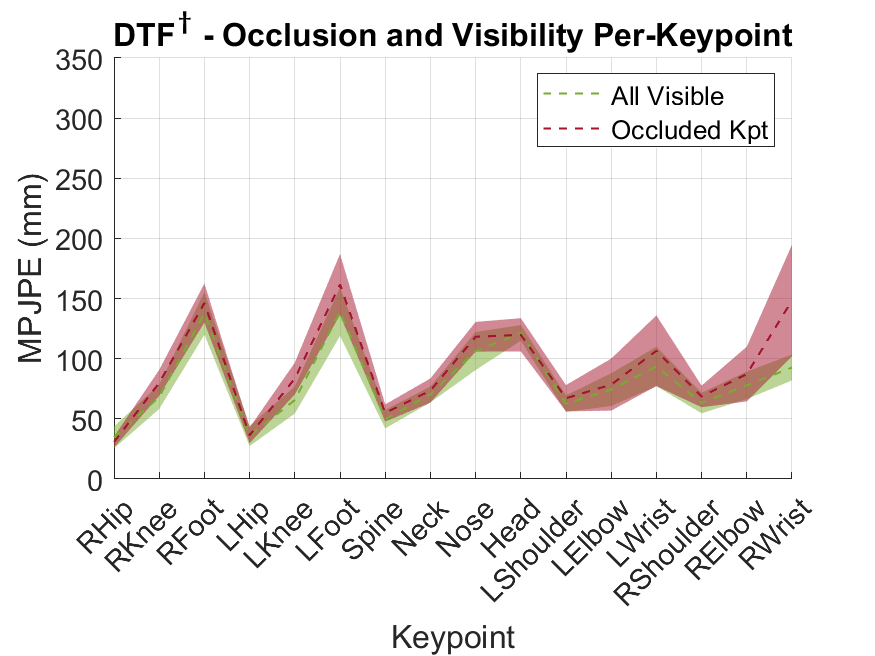}
        \subcaption{}
    \end{minipage}
    \caption{ Comparison of MPJPE per keypoint when all keypoints are visible (GT 2D) vs. when each keypoint is individually occluded with Gaussian noise. Models: (a) VideoPose3D, (b) P-STMO, (c) MixSTE, (d) PoseFormerv2, (e) D3DP, (f) DiffuPose, (g) FinePose, and (h) DTF\(^\dagger\).}
    \label{fig:all_vis_one_occl}
\end{figure}

\begin{figure}[t]
    \centering
    \begin{minipage}{0.245\linewidth}
        \centering
        \includegraphics[width=\linewidth]{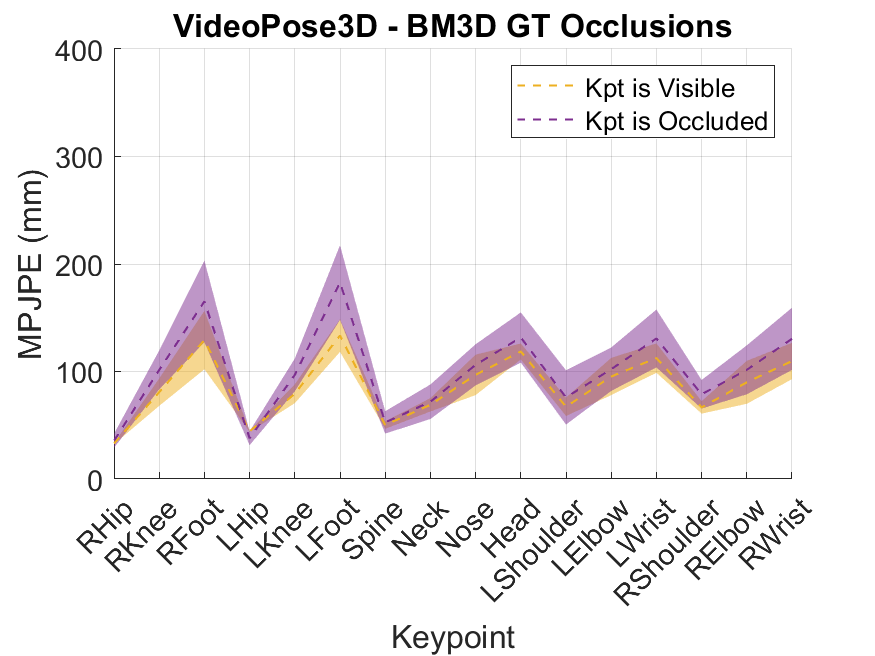}
        \subcaption{}
    \end{minipage}
    \hfill
    \begin{minipage}{0.245\linewidth}
        \centering
        \includegraphics[width=\linewidth]{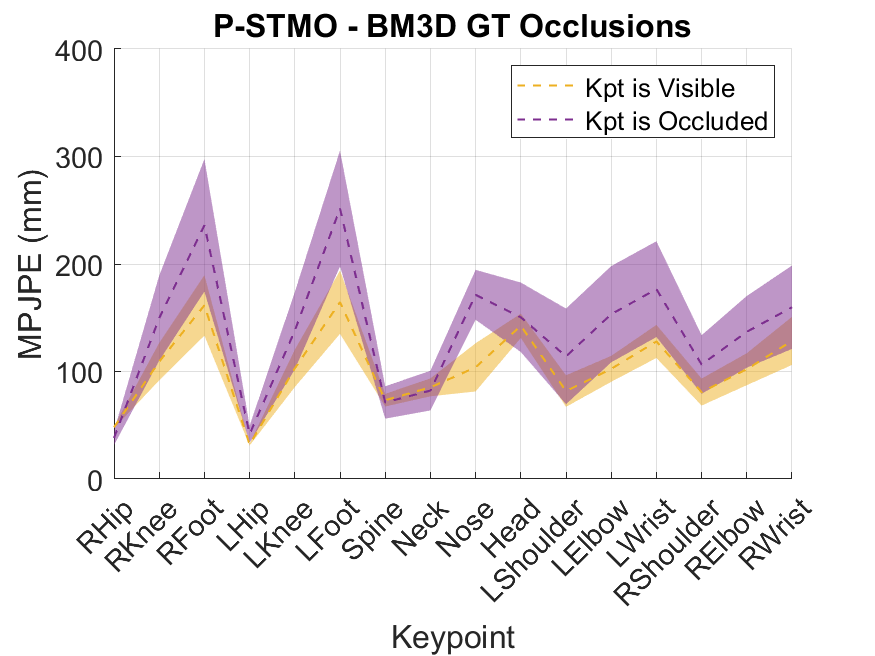}
        \subcaption{}
    \end{minipage}
    \hfill
    \begin{minipage}{0.245\linewidth}
        \centering
        \includegraphics[width=\linewidth]{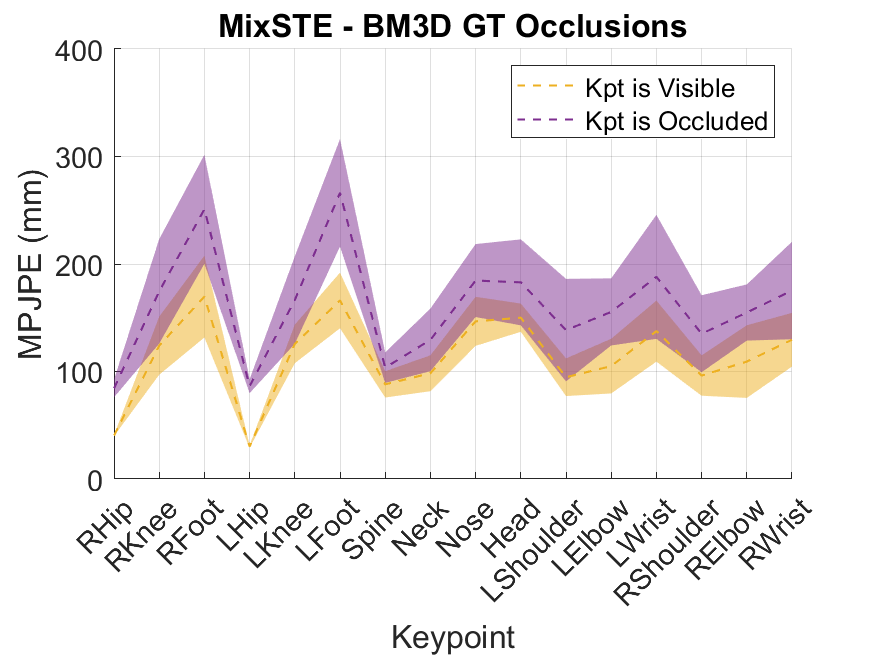}
        \subcaption{}
    \end{minipage}
     \hfill
    \begin{minipage}{0.245\linewidth}
        \centering
        \includegraphics[width=\linewidth]{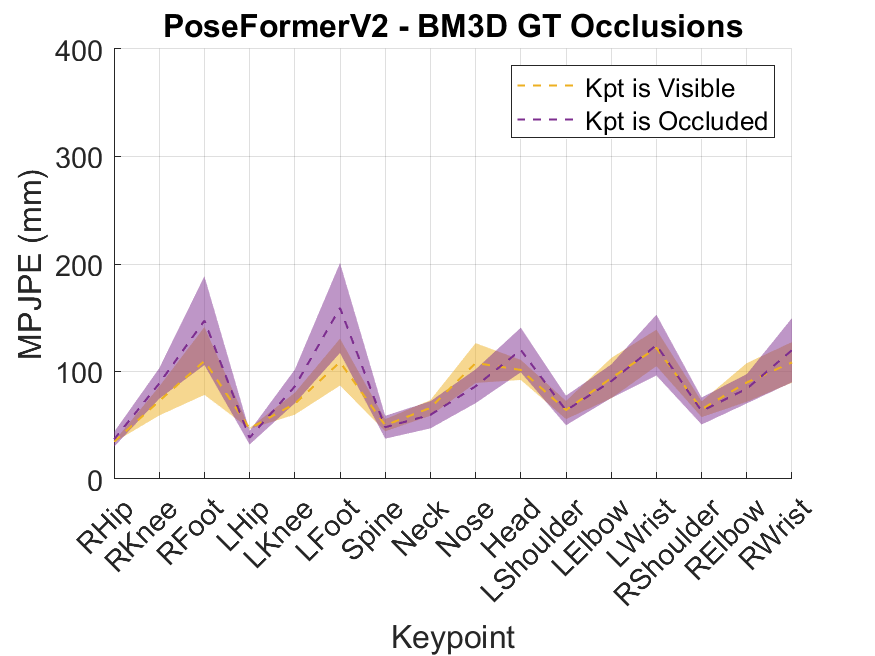}
        \subcaption{}
    \end{minipage}
     \hfill
    \begin{minipage}{0.245\linewidth}
        \centering
        \includegraphics[width=\linewidth]{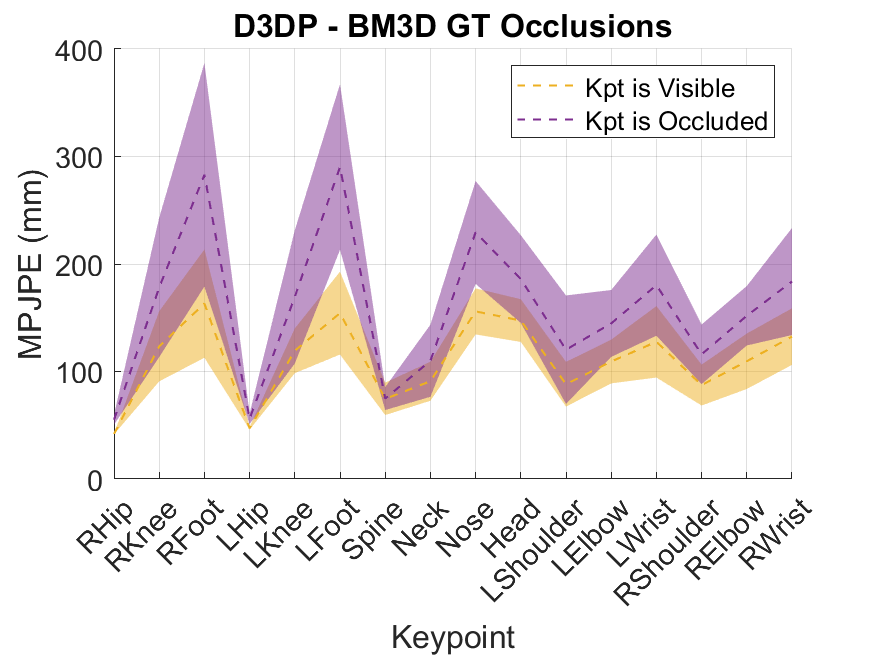}
        \subcaption{}
    \end{minipage}
     \hfill
    \begin{minipage}{0.245\linewidth}
        \centering
        \includegraphics[width=\linewidth]{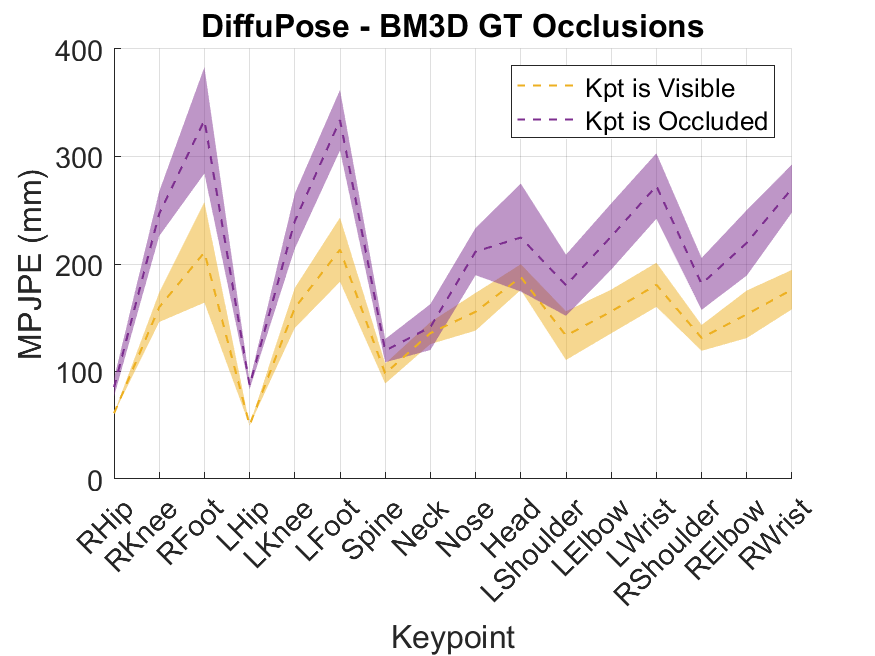}
        \subcaption{}
    \end{minipage}
     \hfill
    \begin{minipage}{0.245\linewidth}
        \centering
        \includegraphics[width=\linewidth]{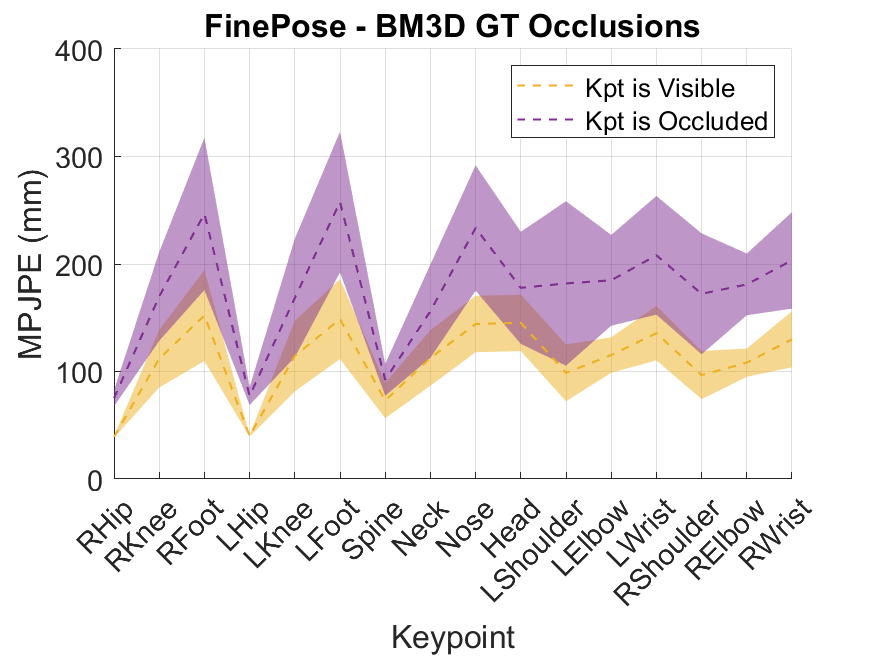}
        \subcaption{}
    \end{minipage}
     \hfill
    \begin{minipage}{0.245\linewidth}
        \centering
        \includegraphics[width=\linewidth]{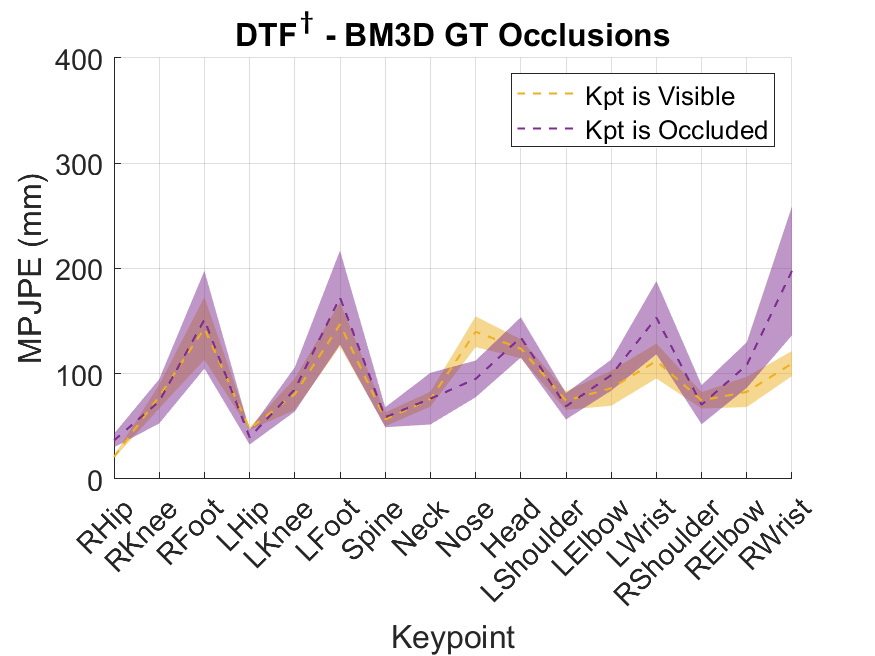}
        \subcaption{}
    \end{minipage}
    \caption{ MPJPE for each keypoint under real occlusion conditions in BlendMimic3D (BM3D). We compare errors when the keypoint is visible versus occluded across poses. These reflect the natural distribution of occlusions and keypoint visibility. Models: (a) VideoPose3D, (b) P-STMO, (c) MixSTE, (d) PoseFormerv2, (e) D3DP, (f) DiffuPose, (g) FinePose, and (h) DTF\(^\dagger\).}
    \label{fig:vis_vs_real_occl}
\end{figure}

This appendix complements the main per-keypoint occlusion analysis by providing detailed, model-specific results. While Figures~\ref{fig:all_mpjpe_per_occl_kpt}, \ref{fig:all_all_vis_one_occl}, and \ref{fig:all_vis_vs_real_occl} in the main paper show average performance across all models, here we present the corresponding individual results for each method under three complementary conditions:

\begin{itemize}
    \item Synthetic occlusion of one keypoint at a time, with analysis of the MPJPE for the occluded keypoint, the remaining visible keypoints, and the full-pose average (Figure~\ref{fig:mpjpe_per_occl_kpt}).
    \item Comparison of per-keypoint MPJPE when all keypoints are visible versus when one is occluded (Figure~\ref{fig:all_vis_one_occl}).
    \item Realistic occlusion scenarios using ground-truth occlusion labels from BlendMimic3D (Figure~\ref{fig:vis_vs_real_occl}).
\end{itemize}

Across all figures, MPJPE is computed for each keypoint and reported separately for: (i) occluded keypoints, (ii) visible keypoints, and (iii) average pose error. The shaded regions represent standard deviations across sequences.

From Figure~\ref{fig:mpjpe_per_occl_kpt}, we observe that MixSTE and FinePose are the most sensitive models to occlusions, followed by P-STMO, DiffuPose and D3DP. This is reflected in the higher MPJPE of occluded keypoints compared to visible ones. A similar behavior is evident in Figure~\ref{fig:all_vis_one_occl}, where the error difference between occluded and visible settings is more pronounced for the same group of models. In contrast, models such as VideoPose3D, PoseFormerV2, DTF\(^\dagger\) show minimal differences between occluded and visible errors per keypoint (Figure~\ref{fig:all_vis_one_occl}), suggesting greater robustness to localized perturbations.

Figure~\ref{fig:vis_vs_real_occl} shows that, under real occlusions most models reproduce the patterns seen in the per-keypoint case, confirming that keypoints like wrists, feet, and head remain the most error-prone. D3DP and DiffuPose, although diffusion-based, degrade significantly under real occlusions, revealing vulnerabilities in their conditioning mechanisms.

Together, these results support the aggregated trends discussed in the main paper and reinforce the importance of occlusion-aware training strategies.

\section{Beyond Occlusions: Camera-Related Sensitivity (Per Model)}

\begin{figure}[t]
    \centering
      \begin{minipage}{0.245\linewidth}
        \centering
        \includegraphics[width=\linewidth]{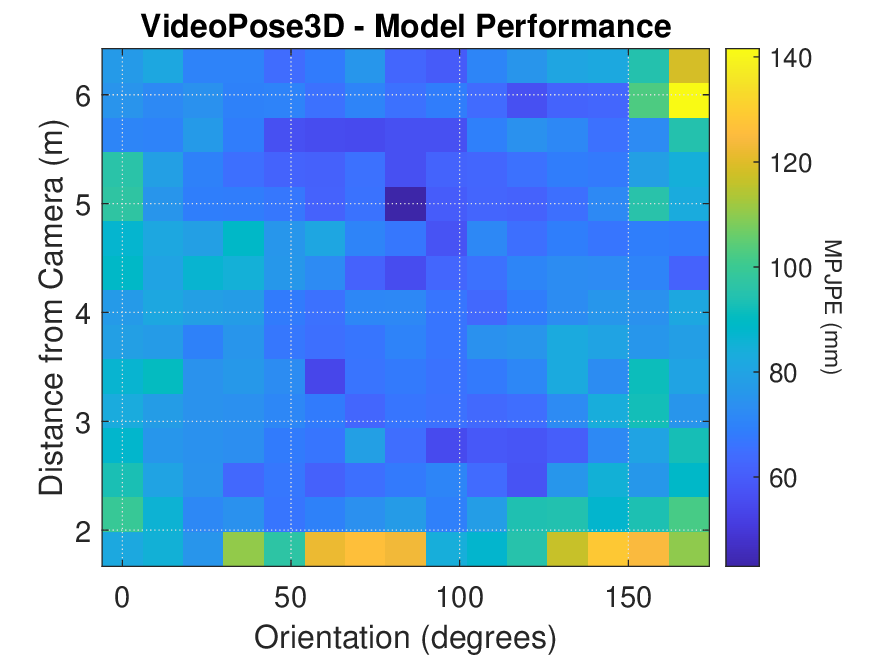}
        \subcaption{}
    \end{minipage}
    \hfill
    \begin{minipage}{0.245\linewidth}
        \centering
        \includegraphics[width=\linewidth]{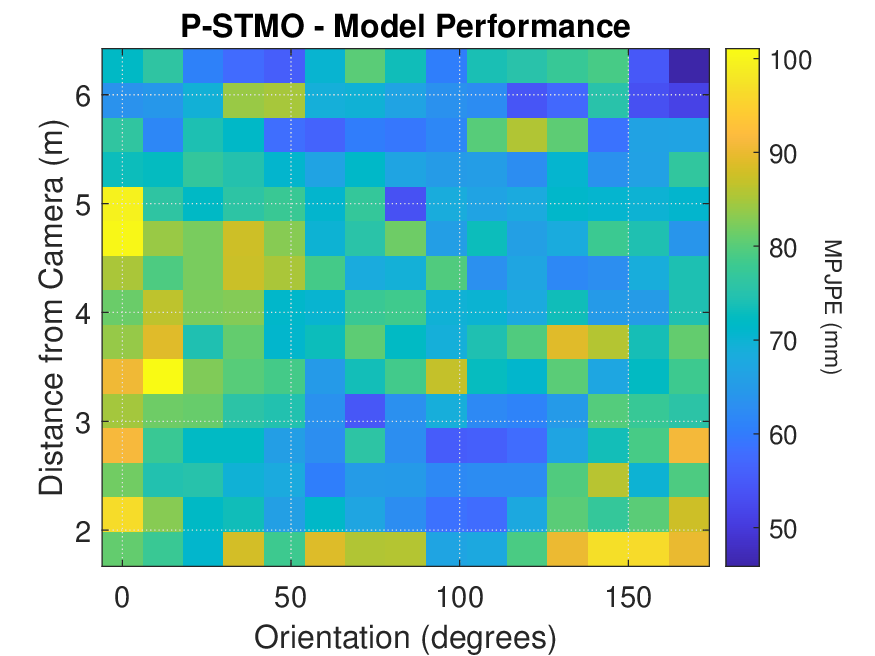}
        \subcaption{}
    \end{minipage} 
    \hfill
    \begin{minipage}{0.245\linewidth}
        \centering
        \includegraphics[width=\linewidth]{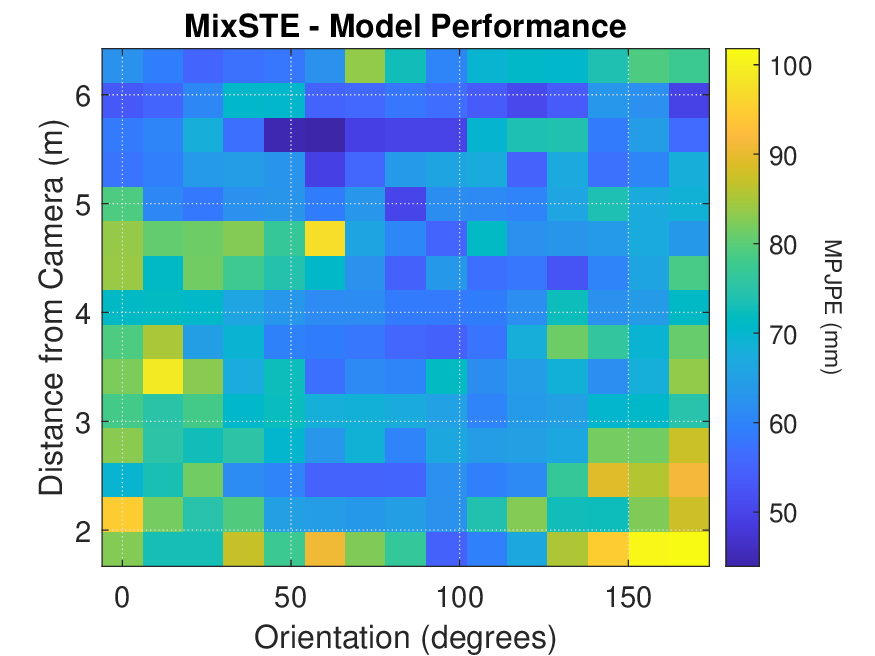}
        \subcaption{}
    \end{minipage}
     \hfill
    \begin{minipage}{0.245\linewidth}
        \centering
        \includegraphics[width=\linewidth]{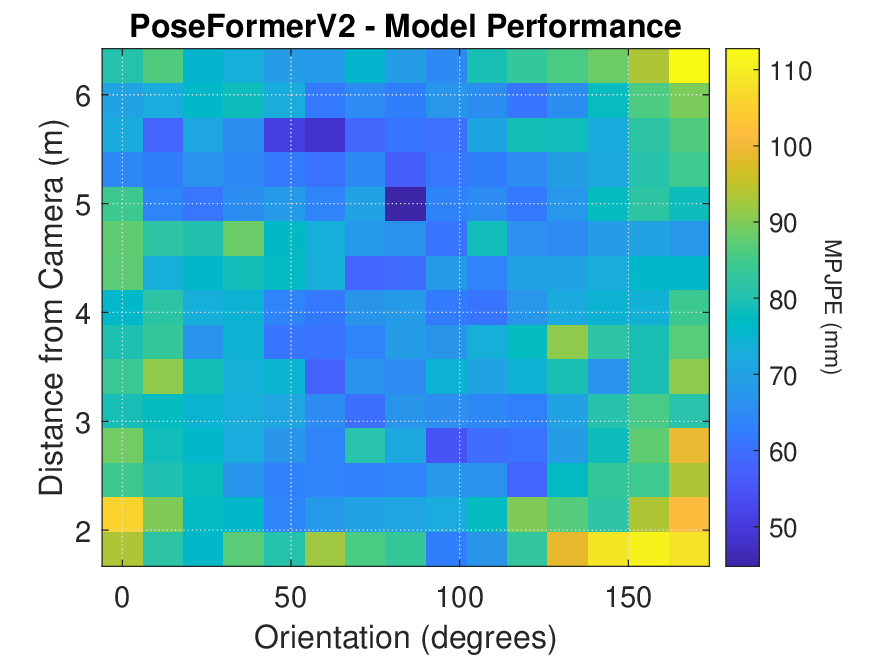}
        \subcaption{}
    \end{minipage}
     \hfill
    \begin{minipage}{0.245\linewidth}
        \centering
        \includegraphics[width=\linewidth]{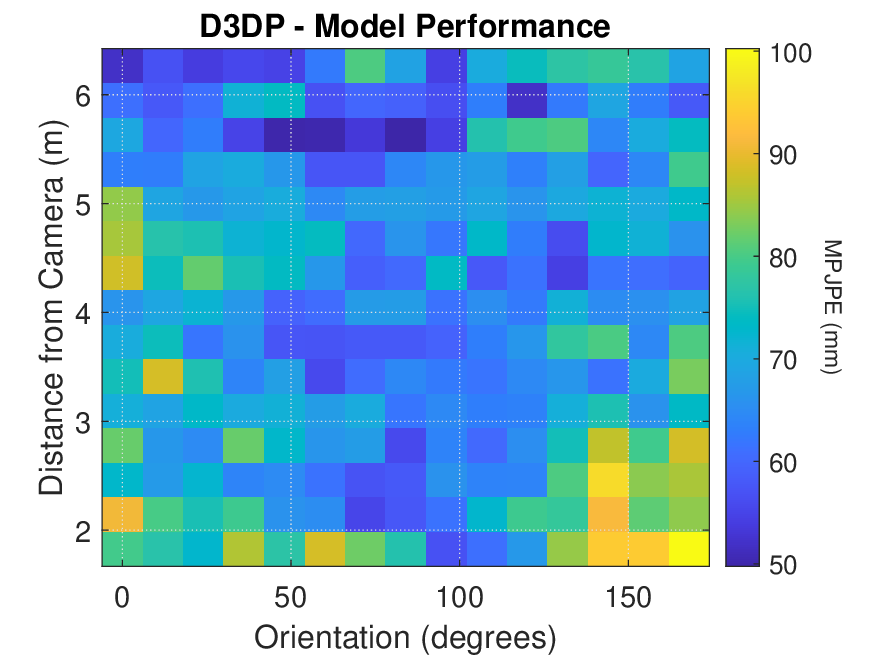}
        \subcaption{}
    \end{minipage}
     \hfill
    \begin{minipage}{0.245\linewidth}
        \centering
        \includegraphics[width=\linewidth]{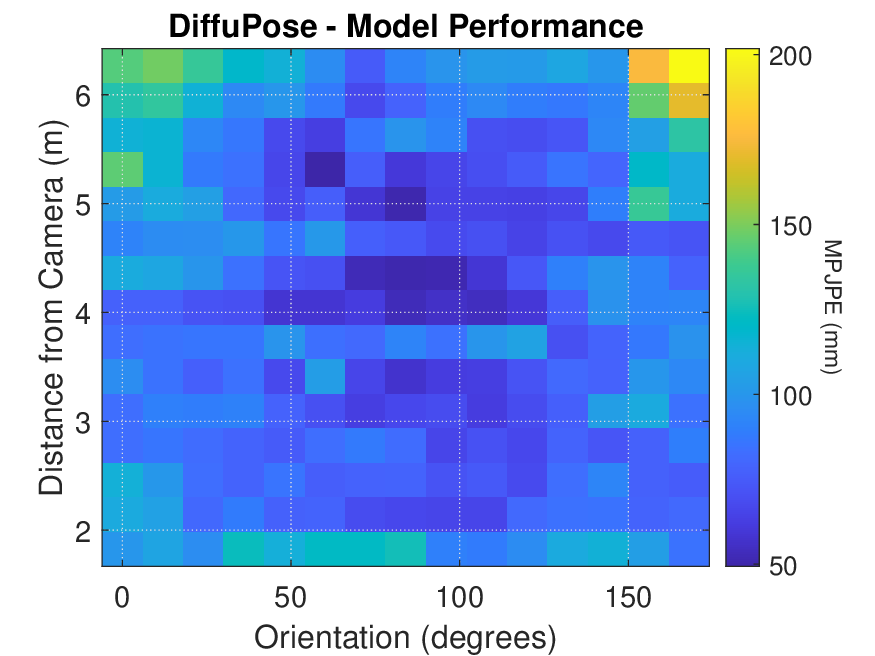}
        \subcaption{}
    \end{minipage}
     \hfill
    \begin{minipage}{0.245\linewidth}
        \centering
        \includegraphics[width=\linewidth]{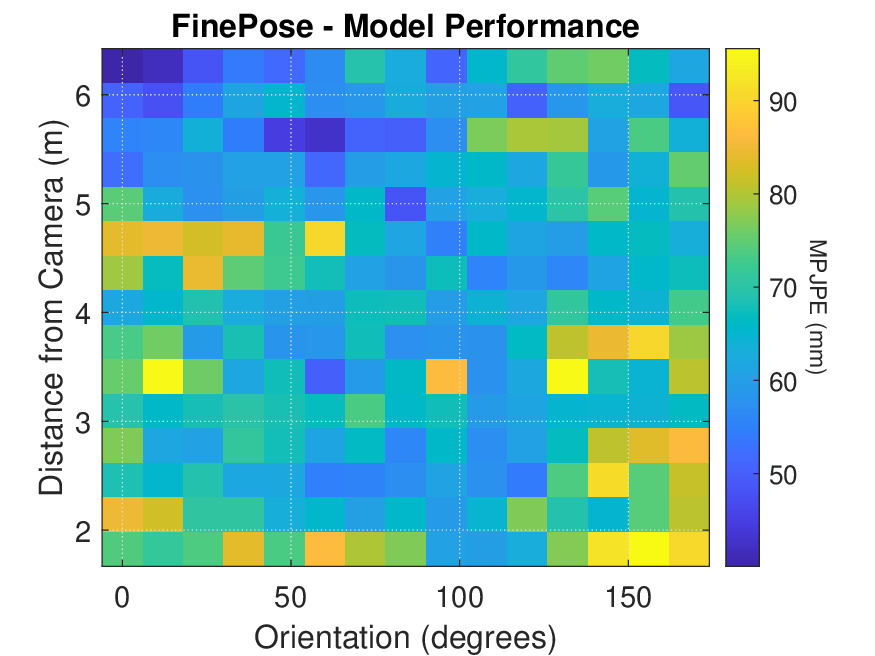}
        \subcaption{}
    \end{minipage}
     \hfill
    \begin{minipage}{0.245\linewidth}
        \centering
        \includegraphics[width=\linewidth]{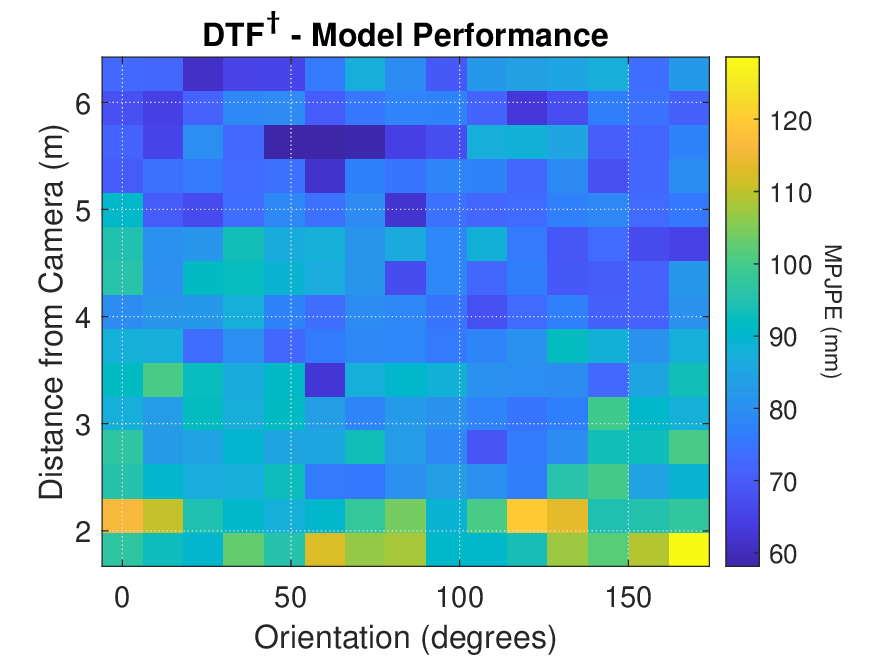}
        \subcaption{}
    \end{minipage}
    \caption{ Model performance as a function of subject orientation and distance from the camera, evaluated on BlendMimic3D under ground-truth 2D inputs. Models: (a) VideoPose3D, (b) P-STMO, (c) MixSTE, (d) PoseFormerv2, (e) D3DP, (f) DiffuPose, (g) FinePose, and (h) DTF\(^\dagger\).
    }
    \label{fig:models_heatmaps}
\end{figure}

While our primary analysis focused on robustness to occlusions, it is important to consider other contextual factors that may influence model performance, such as camera-related aspects. This appendix complements the camera-related sensitivity analysis presented in the main paper by providing detailed, model-specific heatmaps. While Figure~\ref{fig:heatmap_all_models} in the main paper reports average MPJPE across all models, here we examine individual model performance as a function of subject orientation and distance to the camera (see Figure~\ref{fig:models_heatmaps}).

Across all models, we observe a consistent trend: performance degrades at closer distances (below approximately 2.5 meters). As discussed in the main paper, this is likely due to the lack of close-range examples in the training data (see Figure~\ref{fig:bm3d_h36m_bins}).

Beyond this consistent degradation at shorter distances, no clear or systematic trend emerges with respect to subject orientation or longer-range distance. Each model exhibits distinct and often inconsistent performance patterns across orientation angles. While some models, such as VideoPose3D, DiffuPose, and DTF, maintain relatively stable errors across orientations, others, including P-STMO, MixSTE, PoseFormerV2, D3DP, and FinePose display scattered performance spikes concentrated in different regions. This variability suggests orientation and distance do not affect model performance in a predictable way. 

\section{Beyond Occlusions: Action Variability (Per Model)}

\begin{figure}[t]
    \centering
      \begin{minipage}{0.245\linewidth}
        \centering
        \includegraphics[width=\linewidth]{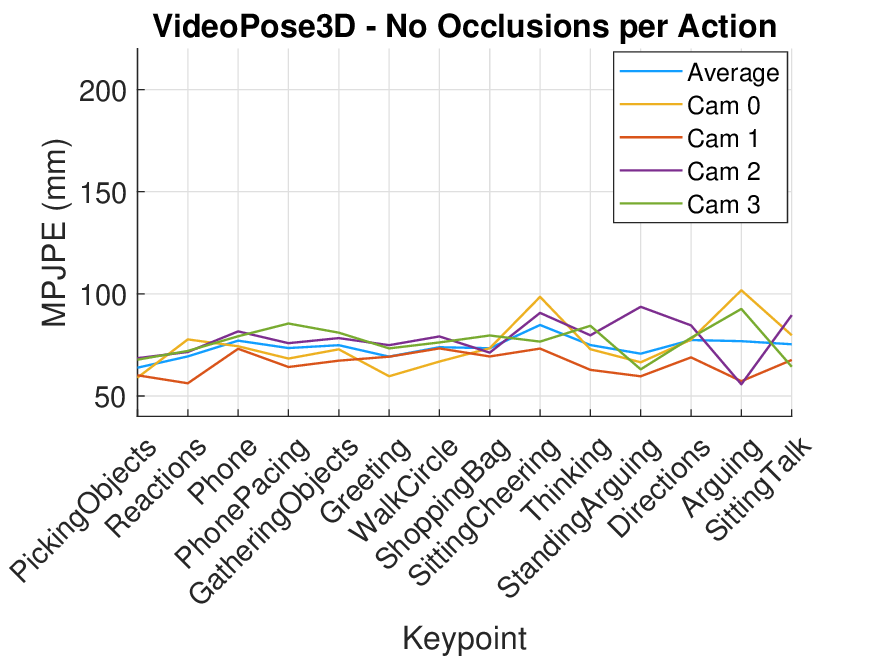}
        \subcaption{}
    \end{minipage}
    \hfill
    \begin{minipage}{0.245\linewidth}
        \centering
        \includegraphics[width=\linewidth]{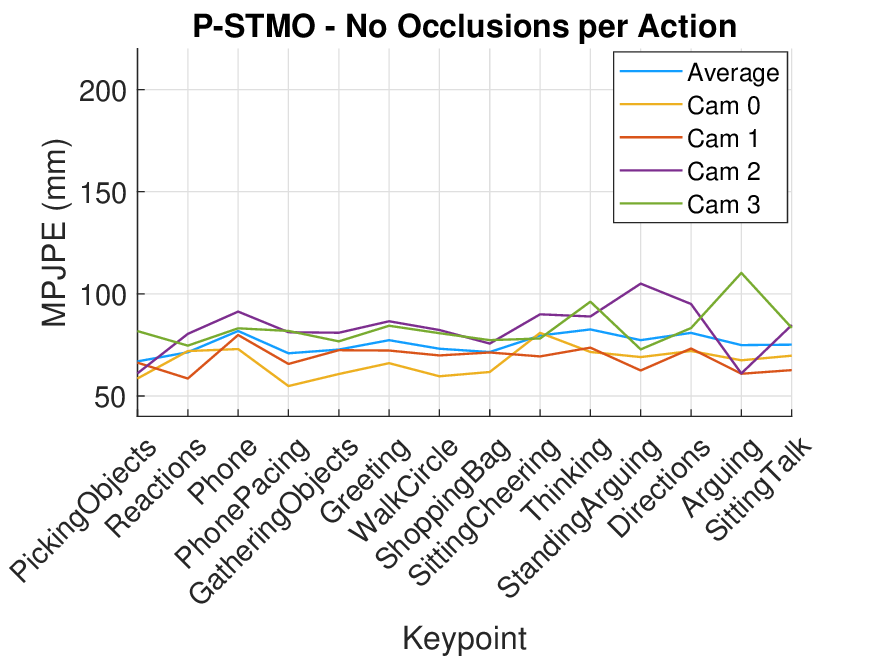}
        \subcaption{}
    \end{minipage} 
    \hfill
    \begin{minipage}{0.245\linewidth}
        \centering
        \includegraphics[width=\linewidth]{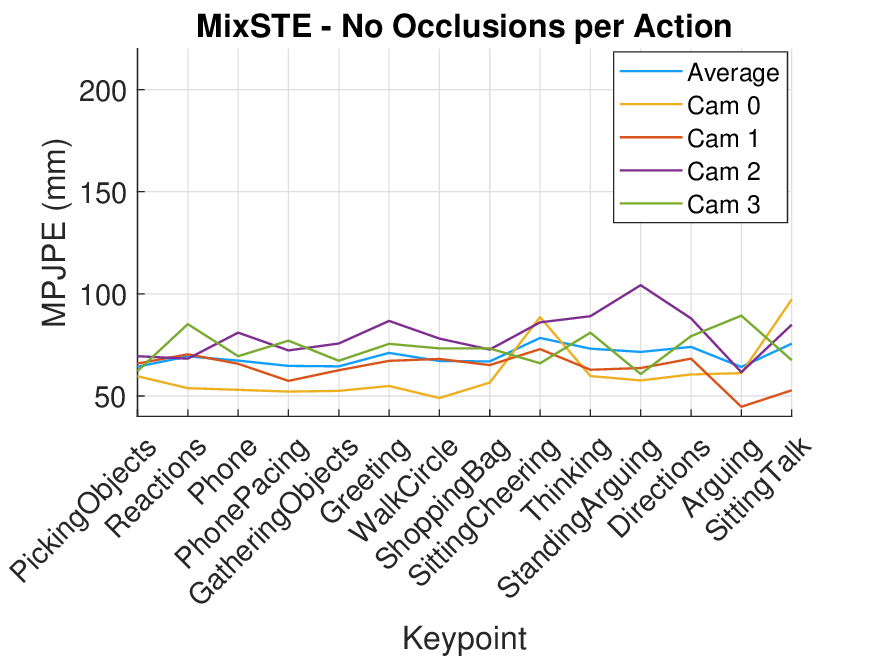}
        \subcaption{}
    \end{minipage}
     \hfill
    \begin{minipage}{0.245\linewidth}
        \centering
        \includegraphics[width=\linewidth]{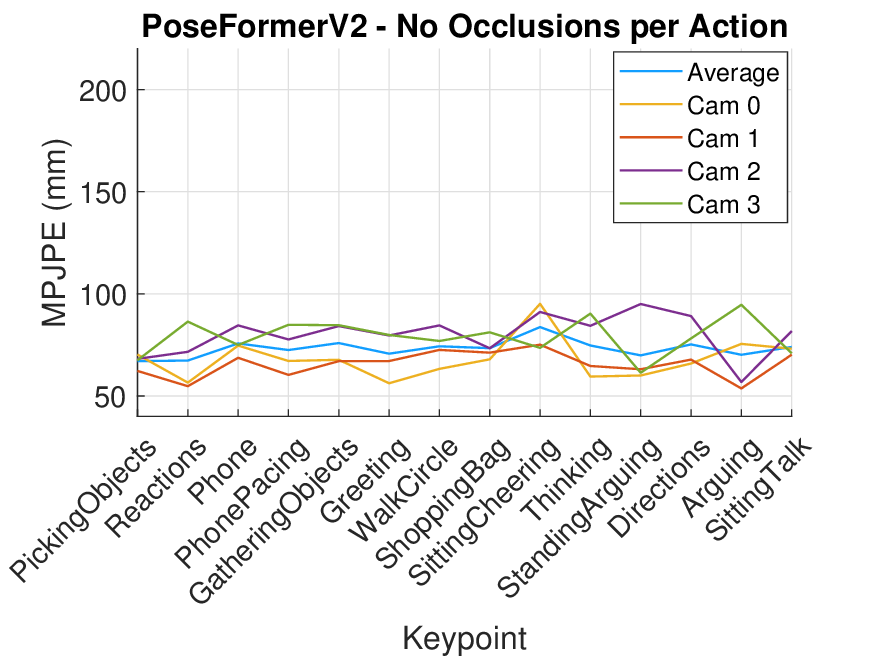}
        \subcaption{}
    \end{minipage}
     \hfill
    \begin{minipage}{0.245\linewidth}
        \centering
        \includegraphics[width=\linewidth]{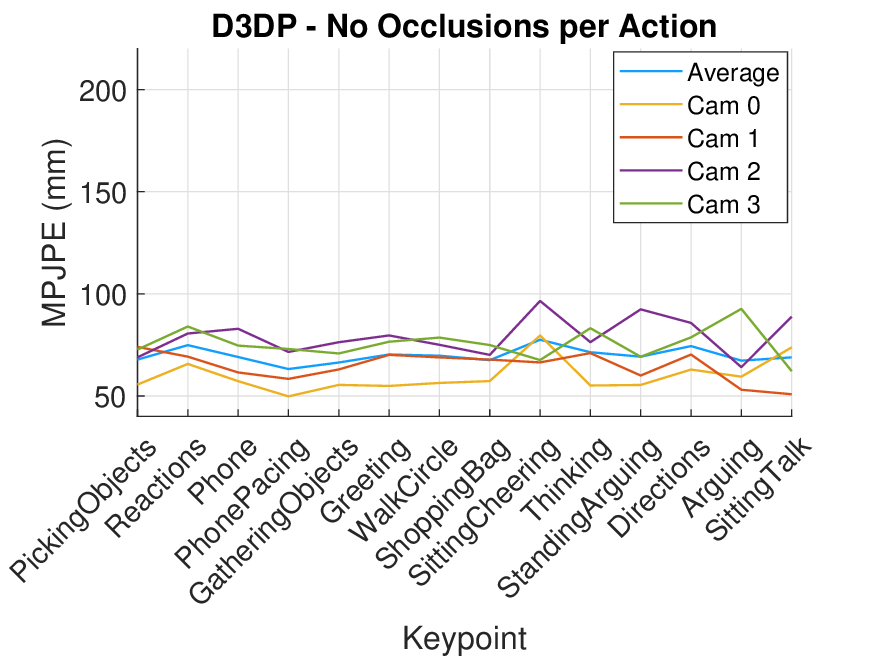}
        \subcaption{}
    \end{minipage}
     \hfill
    \begin{minipage}{0.245\linewidth}
        \centering
        \includegraphics[width=\linewidth]{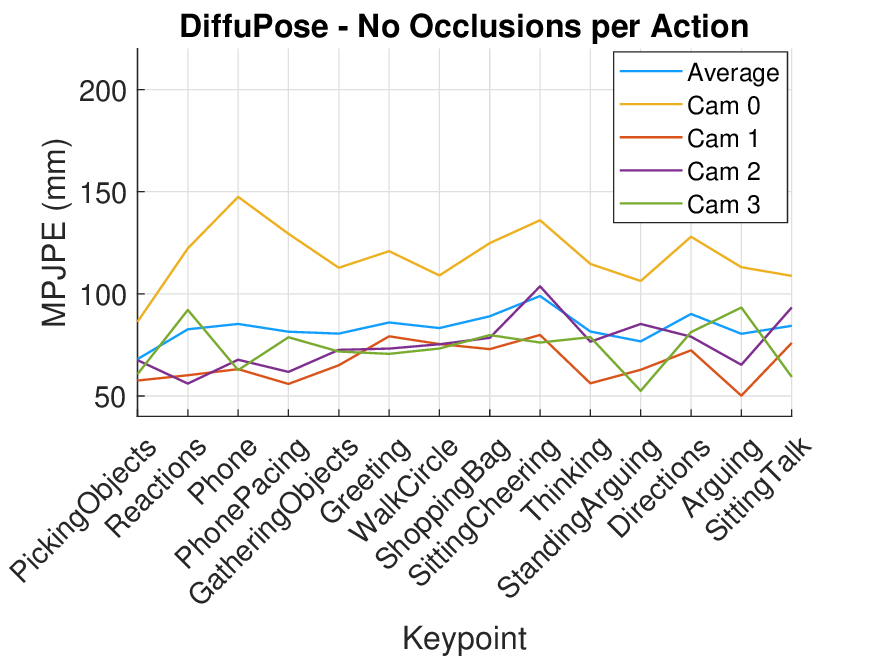}
        \subcaption{}
    \end{minipage}
     \hfill
    \begin{minipage}{0.245\linewidth}
        \centering
        \includegraphics[width=\linewidth]{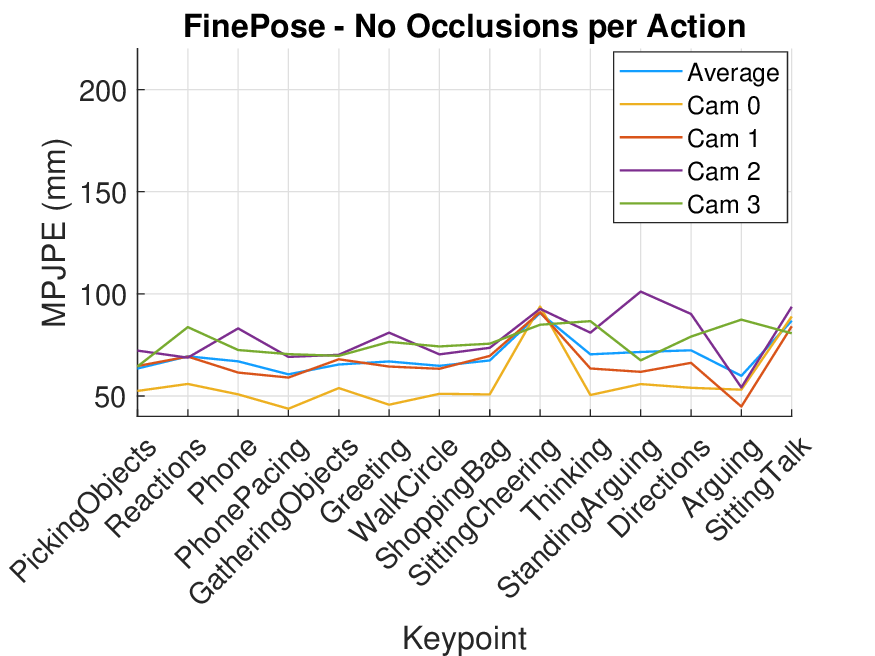}
        \subcaption{}
    \end{minipage}
     \hfill
    \begin{minipage}{0.245\linewidth}
        \centering
        \includegraphics[width=\linewidth]{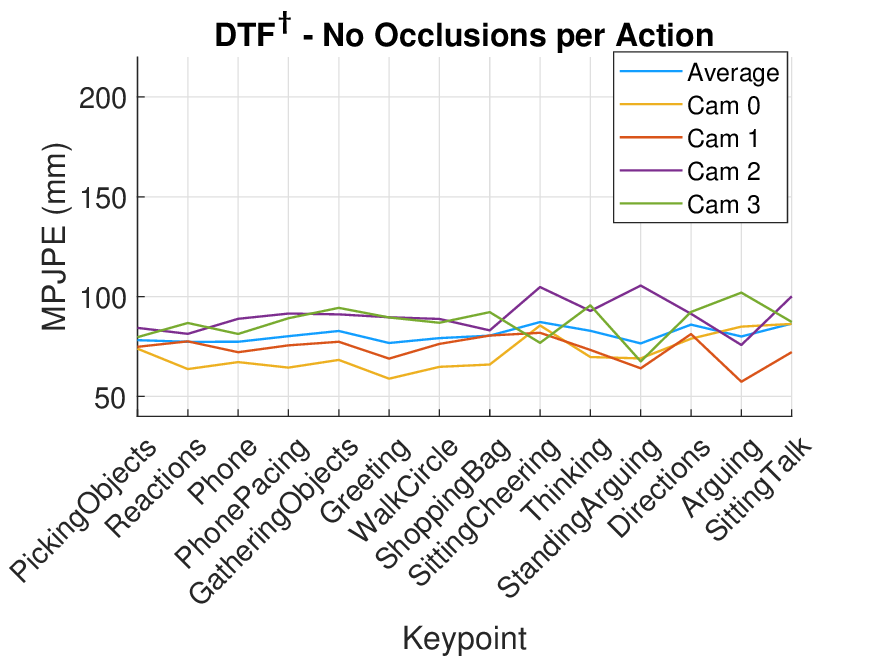}
        \subcaption{}
    \end{minipage}
    \caption{Per-model MPJPE for each action under no occlusions. Results are shown for four camera viewpoints and the average across views. Models: (a) VideoPose3D, (b) P-STMO, (c) MixSTE, (d) PoseFormerv2, (e) D3DP, (f) DiffuPose, (g) FinePose, and (h) DTF\(^\dagger\).}
    \label{fig:actions_per_model}
\end{figure}

\begin{figure}[t]
    \centering
    \includegraphics[width=0.45\linewidth]{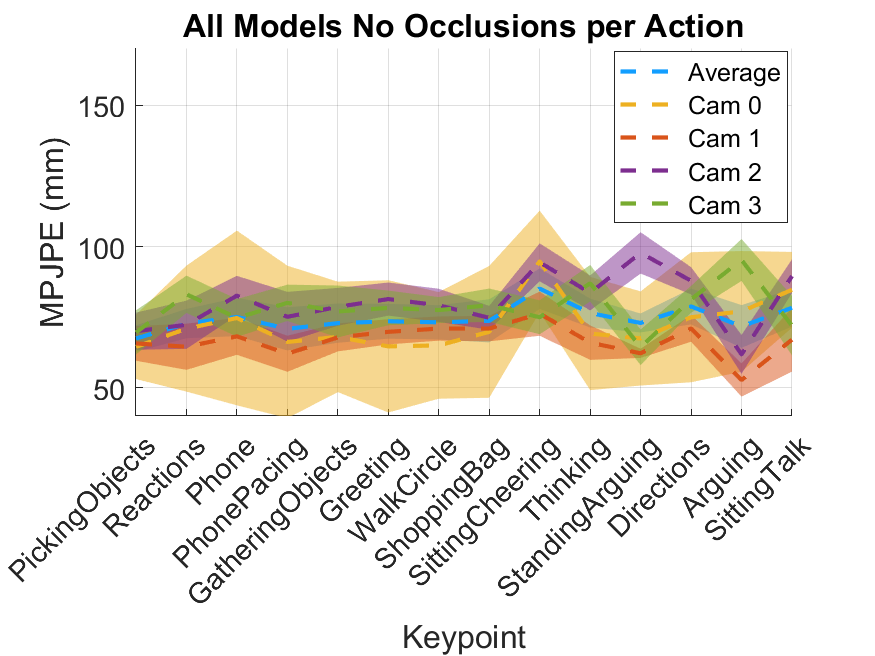}
    \caption{Average MPJPE across all models for each action and camera, under no occlusions.}
    \label{fig:all_models_actions}
\end{figure}

In addition to camera-related analysis, we also investigate the impact of action variability. Different actions include varying poses and spatial configurations, which may affect prediction accuracy. Figure~\ref{fig:actions_per_model} presents the per-model performance across actions, highlighting how each architecture responds to this variability. Complementarily, Figure~\ref{fig:all_models_actions} reports the MPJPE for each action averaged across all models and camera viewpoints. Together, these results indicate that the test set covers a broad and realistic distribution of actions and viewpoints (see also Figure~\ref{fig:bm3d_bins}), supporting the diversity required to evaluate generalization. 

To further assess action-related variability, we evaluate the per-model curves in Figure~\ref{fig:actions_per_model}. No consistent performance trend is observed across actions. Each model displays a unique performance profile depending on the camera and action. However, one exception is the \textit{SittingCheering} action, which frequently appears among the worst-performing across models, likely due to its complex, asymmetric movements. These findings suggest that action variability alone does not strongly influence model performance in the same way occlusions do.

From Figure~\ref{fig:all_models_actions} we can conclude that there is no single action that consistently degrades performance across all cameras or models. While the performance of Camera 0 shows slightly more variability across models, overall performance remains relatively stable across actions in the absence of occlusions.

\section{Beyond Occlusions: Pose Velocity Analysis (Per Model)}

\begin{figure}[t]
    \centering
    \includegraphics[width=0.45\linewidth]{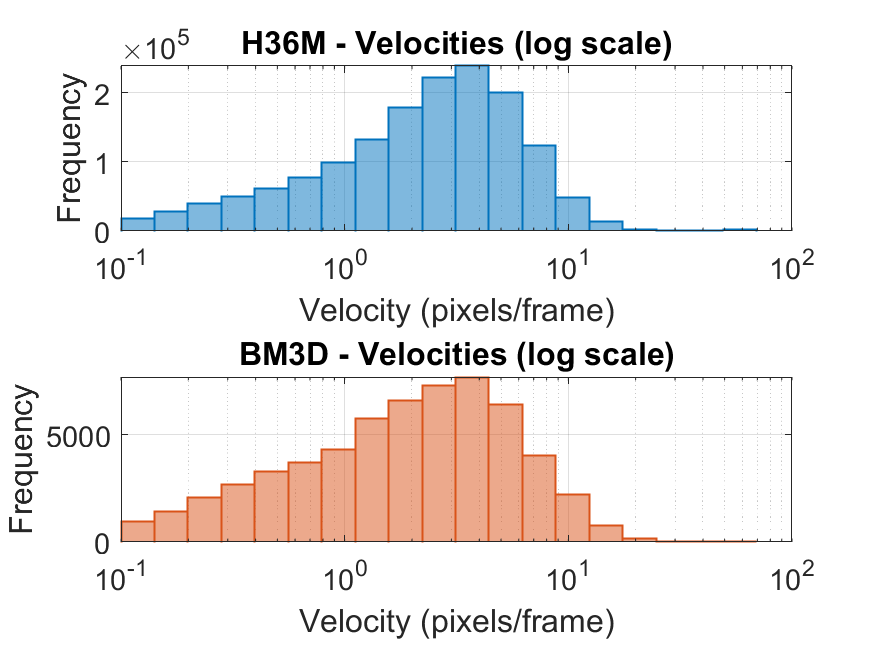}
    \caption{Velocity distribution (in log scale) of keypoint motion in Human3.6M (top) and BlendMimic3D (bottom).}
    \label{fig:velocity_distributions}
\end{figure}

Pose velocity, how fast keypoints move between frames, can also challenge temporal models, especially if velocity distributions in the test set differ from those seen during training. Figure~\ref{fig:velocity_distributions} compares the distribution of 2D pose velocities (in pixels per frame) between Human3.6M (training set) and BlendMimic3D (test set), both on a logarithmic scale. While the two datasets differ slightly in distribution, the overlap suggests that test-time velocities are within the training domain.

\begin{figure}[!t]
    \centering
      \begin{minipage}{0.245\linewidth}
        \centering
        \includegraphics[width=\linewidth]{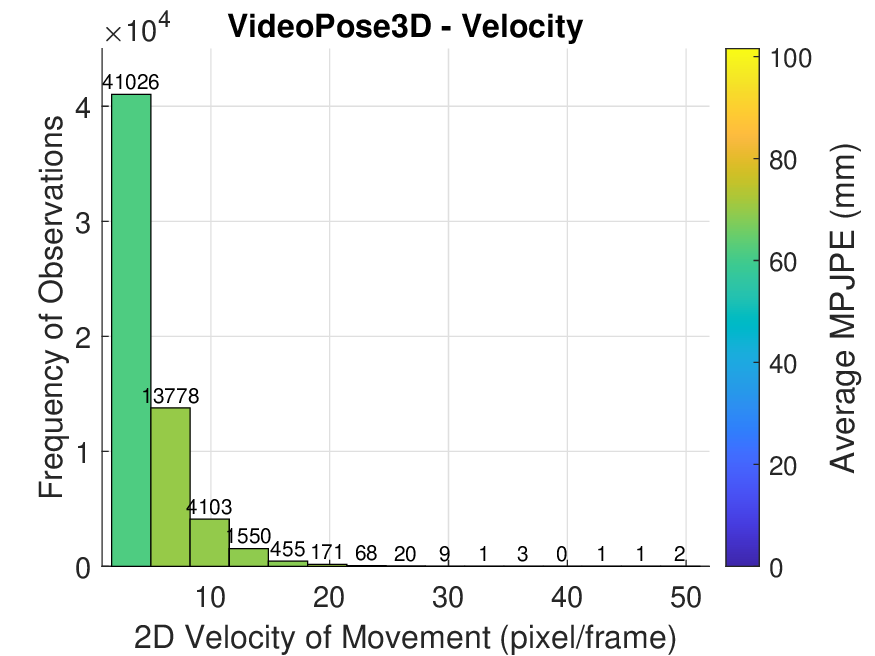}
        \subcaption{}
    \end{minipage}
    \hfill
    \begin{minipage}{0.245\linewidth}
        \centering
        \includegraphics[width=\linewidth]{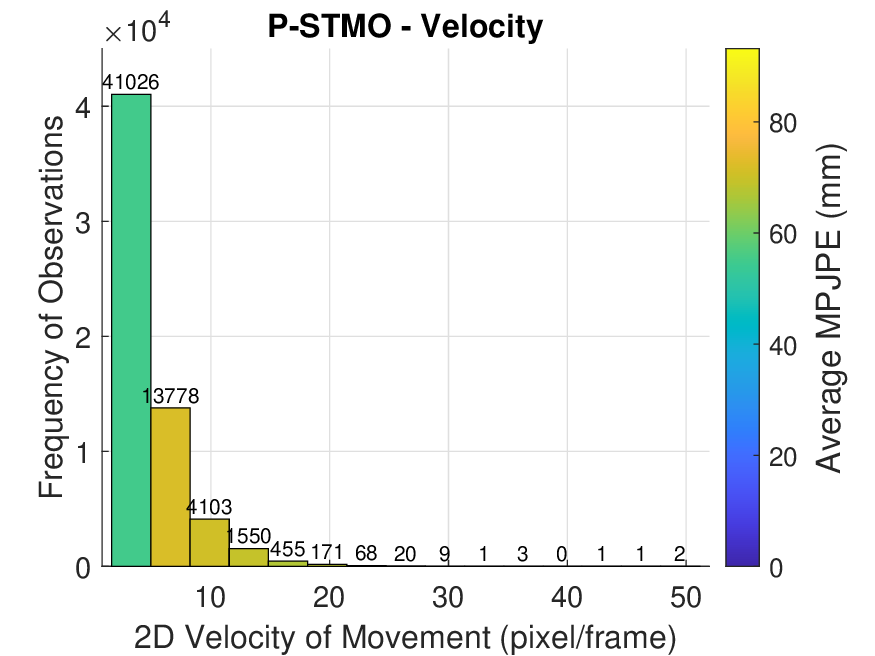}
        \subcaption{}
    \end{minipage} 
    \hfill
    \begin{minipage}{0.245\linewidth}
        \centering
        \includegraphics[width=\linewidth]{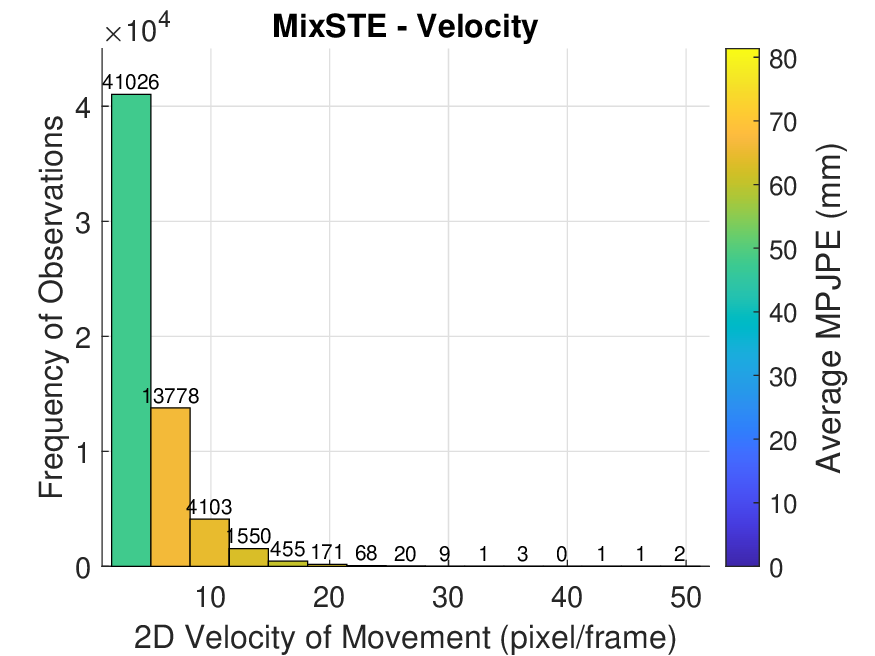}
        \subcaption{}
    \end{minipage}
     \hfill
    \begin{minipage}{0.245\linewidth}
        \centering
        \includegraphics[width=\linewidth]{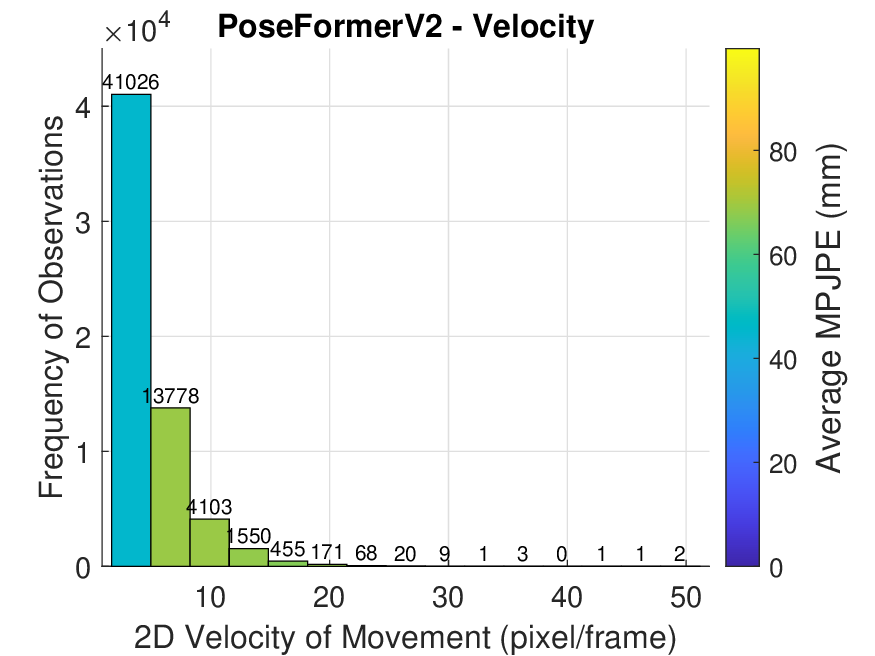}
        \subcaption{}
    \end{minipage}
     \hfill
    \begin{minipage}{0.245\linewidth}
        \centering
        \includegraphics[width=\linewidth]{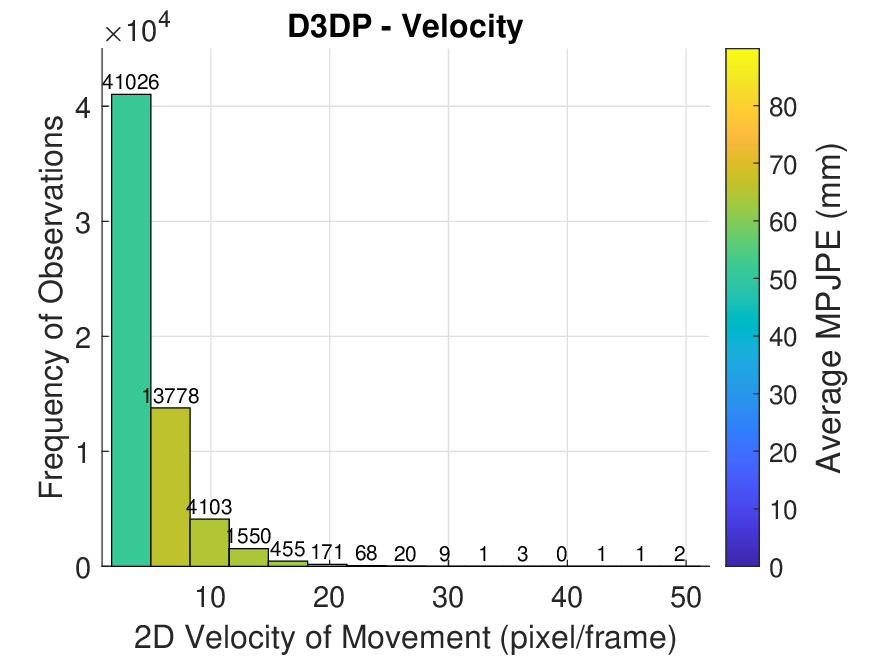}
        \subcaption{}
    \end{minipage}
     \hfill
    \begin{minipage}{0.245\linewidth}
        \centering
        \includegraphics[width=\linewidth]{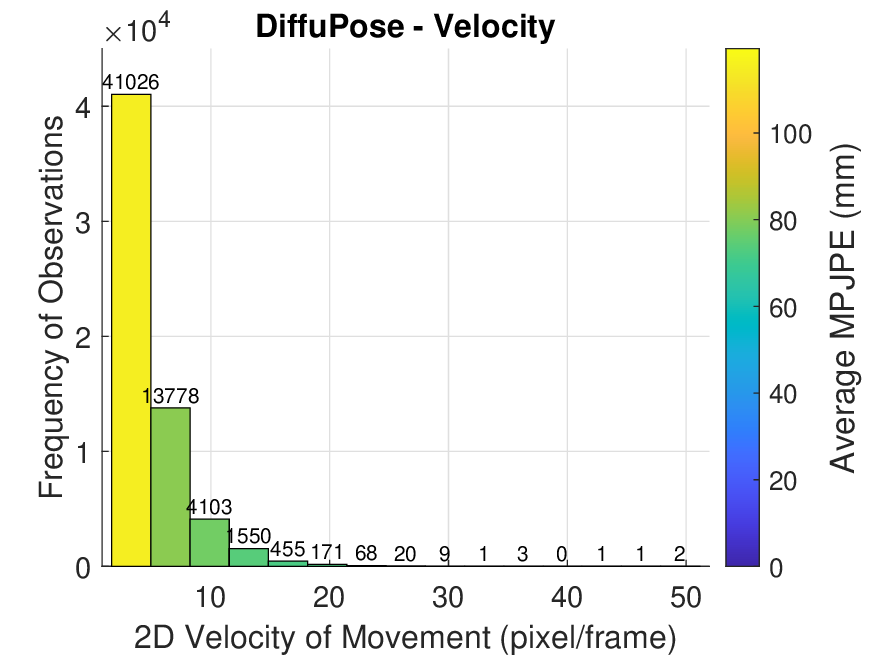}
        \subcaption{}
    \end{minipage}
     \hfill
    \begin{minipage}{0.245\linewidth}
        \centering
        \includegraphics[width=\linewidth]{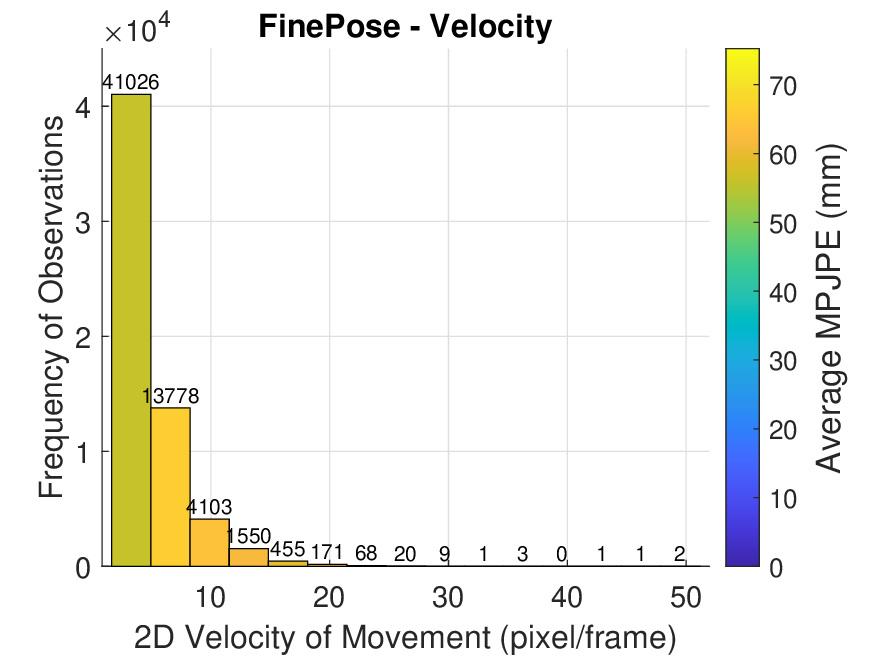}
        \subcaption{}
    \end{minipage}
     \hfill
    \begin{minipage}{0.245\linewidth}
        \centering
        \includegraphics[width=\linewidth]{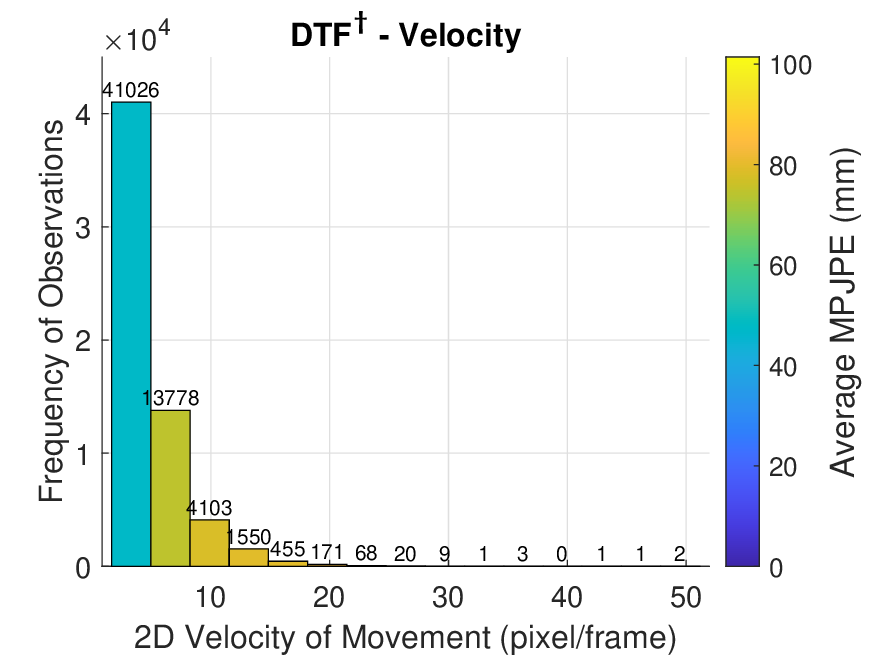}
        \subcaption{}
    \end{minipage}
    \caption{MPJPE as a function of keypoint velocity for each model. Frequency of keypoint occurrences is shown in the background.  Models: (a) VideoPose3D, (b) P-STMO, (c) MixSTE, (d) PoseFormerv2, (e) D3DP, (f) DiffuPose, (g) FinePose, and (h) DTF\(^\dagger\).}
    \label{fig:velocity_per_model}
\end{figure}

Figure~\ref{fig:velocity_per_model} provides a per-model breakdown of model performance as a function of keypoint velocity. Each plot shows the frequency of keypoint observations across velocity bins, along with the corresponding MPJPE. All models except DiffuPose show a small increase in error with velocity. Interestingly, DiffuPose shows higher errors at lower velocities, improving as velocity increases. This counterintuitive behavior may indicate that the diffusion-based model tends to over-process clean inputs, attempting to denoise even when unnecessary.

Overall, the slight initial error increase followed by near-stable performance across velocities suggests that velocity has minimal impact on MPJPE. These results confirm that velocity differences do not account for the performance degradation observed under occlusion, reinforcing the specificity of our occlusion sensitivity analysis.

\end{document}